%% file: Final_Version.tex
\documentclass[journal]{IEEEtran}
\usepackage{graphicx}
\usepackage{amsmath}
\usepackage[linesnumbered, ruled]{algorithm2e}
\usepackage{amsfonts}
\usepackage{amssymb}
\usepackage{bm}
\usepackage{cite}
\usepackage{color}
\usepackage{multirow}
\usepackage{tabularx}
\usepackage{booktabs} 
\usepackage{diagbox} 
\usepackage{multirow} 
\usepackage{makecell} 
\usepackage{longtable} 
\usepackage{makecell} 
\usepackage{colortbl}
\newtheorem{lemma}{Lemma}
\newtheorem{theorem}{\bf Theorem}
\newtheorem{prob}{Problem}
\newtheorem{rem}{\bf Remark}
\newtheorem{assumption}{\bf Assumption}
\newcolumntype{L}[1]{>{\raggedright\arraybackslash}p{#1}}
\newcolumntype{C}[1]{>{\centering\arraybackslash}p{#1}}
\newcolumntype{R}[1]{>{\raggedleft\arraybackslash}p{#1}}
\makeatletter

\newcommand{\Rmnum}[1]{\expandafter\@slowromancap\romannumeral #1@}
\makeatother

\usepackage{times, epsfig}
\usepackage{subfigure}
\usepackage[justification=centering]{caption}
\newlength{\figwidth}
\usepackage{booktabs}
\everymath{\displaystyle}

\begin{document}
\title{Federated Learning Resilient to Byzantine Attacks and Data Heterogeneity}

\author{
Shiyuan Zuo, 
Xingrun Yan,
Rongfei Fan,~\IEEEmembership{Member,~IEEE},
Han Hu,~\IEEEmembership{Member,~IEEE},
Hangguan Shan,~\IEEEmembership{Senior Member,~IEEE},
Tony Q. S. Quek, ~\IEEEmembership{Fellow,~IEEE},
and Puning~Zhao
\thanks{

Manuscript received December 17, 2024; revised April 17, 2025; accepted May 14, 2025.
This paper is supported by Joint Funds of the National Natural Science Foundation of China (NSFC) No. U2336211, NSFC under Grant No. 62171034, No. 92467206, and is also supported by National Key Research and Development Program of China.
(Corresponding Author: Rongfei Fan).}
\thanks{
S.~Zuo, X.~Yan, and R.~Fan are with the School of Cyberspace Science and Technology, Beijing Institute of Technology, Beijing 100081, China (e-mail: \{zuoshiyuan, 3220221473, fanrongfei\}@bit.edu.cn); 
H.~Hu is with the School of Information and Electronics, Beijing Institute of Technology, Beijing 100081, China (e-mail: hhu@bit.edu.cn); 
H.~Shan is with the College of Information Science and Electronic Engineering, Zhejiang University, Hangzhou 310027, China, and also with Zhejiang Provincial Key Laboratory of Information Processing, Communication and Networking, Hangzhou, China 
(e-mail: hshan@zju.edu.cn);
T. Q. S. Quek is with the Singapore University of Technology and Design, Singapore 487372, and also with the Yonsei Frontier lab, Yonsei University, South Korea (e-mail: tonyquek@sutd.edu.sg); 
P. Zhao is with the School of Cyber Science and Technology, Sun Yat-Sen University, Shenzhen 518107, China. (e-mail: zhaopn@mail.sysu.edu.cn).
}
}

\maketitle

\input{abstract}
\input{introduciton}
\input{probelem_statement}

\input{raga}
\input{convergence_analysid}
\input{experiments}

\input{conclusion}

\bibliography{references}
\bibliographystyle{IEEEtran}

\input{biography}

\clearpage

\input{appendix}


\end{document}

%% file: abstract.tex
\begin{abstract}

This paper addresses federated learning (FL) in the context of malicious Byzantine attacks and data heterogeneity. We introduce a novel Robust Average Gradient Algorithm (RAGA), which uses the geometric median for aggregation and {allows flexible round number for local updates.} Unlike most existing resilient approaches, which base their convergence analysis on strongly-convex loss functions or homogeneously distributed datasets, this work conducts convergence analysis for both strongly-convex and non-convex loss functions over heterogeneous datasets.
The theoretical analysis indicates that as long as the fraction of the {data} from malicious users is less than half, RAGA can achieve convergence at a rate of $\mathcal{O}({1}/{T^{2/3- \delta}})$ for non-convex loss functions, where $T$ is the iteration number and $\delta \in (0, 2/3)$. For strongly-convex loss functions, the convergence rate is linear. Furthermore, the stationary point or global optimal solution is shown to be attainable as data heterogeneity diminishes.
Experimental results validate the robustness of RAGA against Byzantine attacks and demonstrate its superior convergence performance compared to baselines under varying intensities of Byzantine attacks on heterogeneous datasets.
\end{abstract}


%% file: introduciton.tex
\section{Introduction} \label{sec:intro}

\IEEEPARstart{T}{he} rapid growth of intelligent applications in recent decade promotes a broad adoption of them on various devices or equipments, such as mobile phones, wearable devices, and autonomous vehicles, etc \cite{zhou2018security, mcmahan2016federated, li2020federated}.
To enable one intelligent application, raw data from various subscribed users has to be trained together to generate a uniform model. 
From the perspective of subscribed users, who may distribute broadly, it has a risk of privacy leakage if they offload their raw data to a central server for training \cite{agrawal2000privacy, duchi2013local}. 
Federated learning (FL) is a distributed training framework and has emerged as a promising solution for this dilemma \cite{konevcny2016federated}.
In FL, model training is completed by exchanging something about model parameters, such as the model parameter itself 
{or the gradient of the loss function with respect to model parameters},
between each involving user and one central server iteratively. In each round of iteration, the central server aggregates the model parameters from every user and subsequently broadcasts the aggregated one to all the users. After receiving the broadcasted model parameter, each user performs the role of local updating based on its own data. In such a procedure, there is no exchange of raw data between any user and the central server \cite{wang2019adaptive, chen2021distributed}.

{However, distributed training paradigms such as FL inherently encounter robustness challenges stemming from the participation of multiple heterogeneous users.}
To be specific, due to data corruption, device malfunctioning, or malicious attacks at some users, the information about the model parameter to be uploaded to the central server by these abnormal users may deviate from the expected one \cite{vempaty2013distributed, yang2020adversary}. In such a case, the group of abnormal users are called as {\it Byzantine users}, and the action of Byzantine users is referred to as {\it Byzantine attacks} \cite{so2020byzantine, cao2019distributed}. As a comparison, the group of normal users are called as {\it Honest users}. For a Byzantine attack, it is general to assume the uncertainty of the identity and the population of Byzantine users. What is more, the attack initiated by Byzantine users could be arbitrarily malicious \cite{chen2017distributed, cao2020distributed}. With such a setup, the training performance of FL will be surely degraded and aggregation strategies resistant to Byzantine attacks have been concerned in literature \cite{xie2018generalized}.

{On the other hand, the training process of FL faces challenges of data heterogeneity among users. This heterogeneity comes from the diverse distribution of local data, as the local data may be collected by each FL user in their specific environmental conditions, habits, or preferences \cite{zhao2018federated}.}
In such a case, data distributions over multiple users are always assumed to be non independently and identically distributed (non-IID).
The presence of data heterogeneity has already been demonstrated to cause local drift of user model, and thus, degradation of the global model's performance.
How to be adaptive to data heterogeneity is also an important issue and has been studied extensively in the area of FL \cite{xie2023fedkl}.

\subsection{Related Works}
{In the field of FL, numerous studies have considered a wide range of factors. However, research on aggregation strategies that are robust to Byzantine attacks in association with rigorous convergence analysis remains limited, primarily due to the significant challenges such attack introduces into the aggregation process.}
The system assumptions, {which serve as an essential precondition for convergence analysis}, usually relate to the nature of the data (IID or non-IID), the nature of the loss function (strongly-convex or non-convex), and the way of local updating (one-step or multiple steps), etc. 
For these assumptions, the data heterogeneity and non-convex loss function make FL framework more general, and multiple steps of local updating waives each user from frequent communication with the central server. 
The Byzantine-resilient aggregation strategies in the literature of FL under diverse system assumptions have been summarized in Table \ref{tab:ref}.

\begin{table*}[!htbp] 
  \centering
  \caption{\centering List of references on the convergence of FL under Byzantine attacks.} \label{tab:ref}
  \resizebox{\textwidth}{!}{
  \begin{tabular}{cccccccc}
    \toprule
    \textbf{References} &\textbf{Algorithm} &\textbf{Function Type} &\textbf{Data Heterogeneity} &\textbf{Local Updating} &\textbf{Optimality Gap} \\
    \midrule
    \cite{xie2018generalized} &median &- &IID &one step &- \\
    \midrule
    \cite{yin2018byzantine} &trimmed mean &\makecell{non-convex \\ strongly-convex} &IID &one step &non-zero \\ 
    \midrule
    \cite{turan2022robust} &RANGE &\makecell{non-convex \\ strongly-convex} &IID &one step &non-zero \\
    \midrule
    \cite{zhao2024huber} & - &\makecell{non-convex \\ strongly-convex} &little data heterogeneity &one step &non-zero \\
    \midrule
    \cite{10.1145/3322205.3311083} &iterative filtering &strongly-convex &IID & one step &non-zero \\ 
    \midrule
    \cite{zhao2023high} &- &strongly-convex &IID &one step &non-zero \\
    \midrule
    \cite{blanchard2017machine} &Krum &non-convex &IID &one step &non-zero \\ 
    \midrule
    \cite{fan2022bev} &BEV-SGD &non-convex &IID &one step &zero \\  
    \midrule
    \cite{li2019rsa} &RSA &strongly-convex &non-IID &one step &non-zero \\ 
    \midrule
    \cite{luan2024robust} &MCA &strongly-convex &non-IID &one step &non-zero \\ 
    \midrule
    \cite{pillutla2022robust} &RFA &strongly-convex &non-IID &multiple steps &non-zero \\ 
    \midrule
    \cite{wu2020federated} &Byrd-SAGA &strongly-convex &IID &one step &non-zero \\ 
    \midrule
    \cite{zhu2023byzantine} &BROADCAST &strongly-convex &IID &one step &non-zero \\ 
    \midrule
    {\cite{huang2024self}} &{SDEA} &{-} &{non-IID} &{multiple steps} &{-} \\ 
    \midrule
    {\cite{data2021byzantine}} &{RAGE} &{\makecell{non-convex \\ strongly-convex}} &{non-IID} &{multiple steps} &{non-zero} \\ 
    \midrule
    \textit{This work} &RAGA &\makecell{non-convex \\ strongly-convex} &non-IID &multiple steps &non-zero \\
    \bottomrule
  \end{tabular}
  }
\end{table*}
As of the aggregation strategies in related literature, by leveraging median \cite{xie2018generalized} or iterative filtering \cite{10.1145/3322205.3311083}, the associated FL system can tolerate a small fraction of Byzantine users. To be more robust to Byzantine attacks, some methods such as trimmed mean \cite{yin2018byzantine}, Robust Aggregating Normalized Gradient Method (RANGE) \cite{turan2022robust}, a Huber Loss Minimization \cite{zhao2024huber}, semi verified mean estimation \cite{zhao2023high}, Krum \cite{blanchard2017machine}, Best Effort Voting SGD (BEV-SGD) \cite{fan2022bev}, Robust Stochastic Aggregation (RSA) \cite{li2019rsa}, Maximum Correntropy Aggregation (MCA) \cite{luan2024robust}, Self-Driven Entropy Aggregation (SDEA) \cite{huang2024self}, and Robust Accumulated Gradient Estimation (RAGE) \cite{data2021byzantine} have been proposed.
Specifically, 
\cite{yin2018byzantine} aggregates the upload vectors by calculating their coordinate-wise median or coordinate-wise trimmed mean based on \cite{xie2018generalized}. RANGE takes the median of uploaded vectors from each client as the aggregated one, and each uploaded vector is supposed to be the median of the previously transmitted gradients. \cite{zhao2024huber} improve the trimmed mean on non-IID datasets by utilizing a multi-dimensional Huber loss function to aggregate uploaded vectors. \cite{zhao2023high} utilizes semi-verified mean estimation to filter out vectors from Byzantine attackers before aggregation, building on the iterative filtering method. Krum \cite{blanchard2017machine} selects a stochastic gradient as the global one, knowing the number of Byzantine attackers, by choosing the gradient with the shortest cumulative distance from a group of the most closely distributed stochastic gradients, BEV-SGD \cite{fan2022bev} requires each user to normalize the associated local gradient to be a vector with zero mean and unit variance before offloading it to the central server, RSA \cite{li2019rsa} penalizes the difference between local model parameters and global model parameters to isolate Byzantine users, while MCA \cite{luan2024robust} utilizes its the maximum correntropy of the model parameters' distribution characteristics and compute the central value to update the global model parameter.
{
Moreover, the SDEA method \cite{huang2024self} partially leverages a public dataset to enhance global model training and mitigate Byzantine attacks, with its effectiveness validated through empirical experiments.
In contrast, the RAGE approach \cite{data2021byzantine} enables multiple local gradient updates under heterogeneous data distributions. While theoretically well-founded, this method assumes that the Byzantine ratio, i.e., the proportion of malicious participants, remains below one-third.
}

Although the above aggregation strategies can be somewhat resistant to Byzantine attacks at some extent, some of them may still fail to work in front of heavy Byzantine attacks \cite{baruch2019little, xie2020fall}.
Recently, the geometric median has emerged as a promising candidate for aggregation due to its ability to naturally tolerate a larger fraction of strong Byzantine attacks without any additional conditions\cite{minsker2015geometric}.
Based on geometric median, some algorithms such as Robust Federated Aggregation (RFA) \cite{pillutla2022robust}, Byzantine attack resilient distributed Stochastic Average Gradient Algorithm (Byrd-SAGA) \cite{wu2020federated}, and Byzantine-RObust Aggregation with gradient Difference Compression And STochastic variance reduction (BROADCAST) \cite{zhu2023byzantine}, have been proposed. 
To be specific, 
geometric median is leveraged to aggregate the uploaded vectors for RFA, Byrd-SAGA, and BROADCAST. Differently, each user in RFA selects the trimmed mean of the model parameters over multiple local updating as the uploaded vector, while Byrd-SAGA and BROADCAST utilize the SAGA \cite{defazio2014saga} method to generate the vectors for aggregation first and then upload them in a lossless or compressive way.

{In the aforementioned geometric median aggregation strategies, there has been little work claiming to be adaptive to data heterogeneity, though this remains a non-negligible issue.}
RFA in \cite{pillutla2022robust} is an exception.
However, RFA only investigates strongly-convex loss function.
With such a convexity setup, RFA shows that it is can converge to a neighborhood of the optimal solution only if the number of local updates for each involving user is larger than then a threshold so as to fulfill a necessary inequality. However, this restricts the flexibility of selecting the number of local updates and maybe computationally intensive when these users have limited computation capability.


\subsection{Motivations and Contributions}
In this paper we investigate Byzantine-resilient FL with adaptivity to data heterogeneity, {which was only studied in \cite{li2019rsa}, \cite{luan2024robust}, \cite{pillutla2022robust}, \cite{huang2024self}, and \cite{data2021byzantine}, referred to as RSA, MCA, RFA, SDEA, and RAGE, respectively.
Different from RSA, MCA, and RFA, which merely considers strongly-convex loss function,
} 
we investigate not only strongly-convex but also non-convex loss function.
{Specifically, in departure from RFA, which also uses geometric median for aggregation but has a convergence condition that necessitates local update steps exceeding a theoretically established threshold, our architecture introduces a flexible round number of local updates scheme, thereby eliminating this fundamental constraint while maintaining convergence guarantees.}

{While SDEA is effective in practice, it lacks a formal theoretical convergence analysis and depends on the availability of public data, which may not always be practical or feasible.
In contrast to SDEA, our approach does not assume access to any public data and provides rigorous theoretical guarantees.
Moreover, unlike RAGE, which requires the Byzantine ratio to be no greater than one-third, our method can tolerate a higher Byzantine ratio, potentially approaching one-half.
}
With our featured system setup, we propose a novel FL structure, RAGA.
Specifically, the RAGA leverages the geometric median to aggregate the local gradients offloaded by the users after multiple rounds of local updating.
Associated analysis of convergence and robustness are then followed for both non-convex and strongly-convex loss functions.
Our main contributions are summarized as follows:
\begin{itemize}
  \item {Algorithmically, we propose a new FL aggregation strategy, named RAGA, which has the flexibility of arbitrary selection of local update's step number, and is robust to both Byzantine attack and data heterogeneity, and does not rely on the use of a public dataset.}
  \item {Theoretically, we {establish} a convergence guarantee for our proposed RAGA under not only non-convex but also strongly-convex loss function. Through rigorous proof, we show that the established convergence can be promised as long as the fraction of dataset from Byzantine users is less than half.} The achievable convergence rate can be at $\mathcal{O}({1}/{T^{2/3- \delta}})$  where $T$ is the iteration number and $\delta \in (0, 2/3)$ for non-convex loss function, and at linear for strongly-convex loss function. 
  Moreover, stationary point and global optimal solution are shown to be obtainable as data heterogeneity, algorithmic error, and stochastic local gradient variance diminish, for non-convex and strongly-convex loss function, respectively.
  \item Empirically, we conduct extensive experiments to evaluate the performance of our proposed RAGA. {With non-IID dataset, our proposed RAGA is verified to be robust to various Byzantine attack types and ratios. Moreover, compared with baseline methods, our proposed RAGA can achieve higher test accuracy.}
\end{itemize}

\subsection{Organization and Notation}
The rest of this paper is organized as follows: Section \ref{sec:model} introduces the system model and formulates the problem to be solved in this paper, Section \ref{sec:RAGA} presents our proposed RAGA. In Section \ref{sec:conver}, the analysis of convergence and robustness for our proposed RAGA under non-convex and strongly-convex loss function are disclosed. Experimental results are illustrated in Section \ref{sec:num}, followed by conclusion remarks in Section \ref{sec:conclu}. 

{\it Notation:} We use plain and bold letters to represent scalars and vectors respectively, e.g., $a$ and $b$ are scalars, $\bm{x}$ and $\bm{y}$ are vectors. $\left\langle \bm{x}, \bm{y} \right\rangle$ denotes the inner product of the vectors $\bm{x}$ and $\bm{y}$, $\nabla F(\bm{x}_0)$ represents the gradient vector of function $F( \bm{x} )$ at $\bm{x}=\bm{x}_0$, $\left\lVert \bm{x} \right\rVert$ stands for the Euclidean norm $\left\lVert \bm{x} \right\rVert _2$, $\mathbb{E}\left\{ a \right\}$ and $\mathbb{E}\left\{ \bm{x} \right\}$ implies the expectation of random variable $a$ and random vector $\bm{x}$ respectively, and $a \propto b$ means that $a$ is proportional to $b$.

%% file: probelem_statement.tex
\section{Problem Statement}
\label{sec:model}



Consider a FL system with one central server and $M$ users, which composite the set $\mathcal{M} \triangleq \{1, 2, ..., M\}$. At the side of $m$th user, it has a dataset $\mathcal{S}_m$, which has $S_m$ ground-true labels. In the FL system, the learning task is to train a model parameter $\bm{w} \in \mathbb{R}^p$ to minimize the global loss function, denoted as $F(\bm{w})$, for approximating the data labels of all the users. Specifically, we need to solve the following problem
\begin{prob} \label{pro:min_F}
  \begin{equation*}
    \mathop{\min}_{\bm{w} \in \mathbb{R}^p } F(\bm{w}).
  \end{equation*}
\end{prob}
In Problem \ref{pro:min_F}, the global loss function $F(\bm{w})$ is defined as
{
\begin{equation}
  F(\bm{w}) \triangleq \frac{1}{\sum_{m \in \mathcal{M} }  S_m } \sum_{m \in \mathcal{M} }  \sum_{\bm{s} \in \mathcal{S}_m}  f(\bm{w}, \bm{s}), 
\end{equation}}
where $f(\bm{w}, \bm{s})$ denotes the loss function for data sample $\bm{s}$. For convenience, we define the local loss function of $m$th user as
\begin{equation}
  F_m(\bm{w}) \triangleq \frac{1}{S_m } \sum_{\bm{s} \in \mathcal{S}_m}  f(\bm{w}, \bm{s}), 
\end{equation}
then the global loss function $F(\bm{w})$ can be rewritten as
\begin{equation}
  F(\bm{w}) = \sum_{m \in \mathcal{M} } \frac{S_m}{\sum_{i \in \mathcal{M} } S_i} F_m(\bm{w}).
\end{equation}

In traditional FL, iterative interactions between a group of $M$ users and a central server are performed to update the gradient parameter $\nabla F(\bm{w})$ until convergence. For $t$th round of iteration, a conventional operation procedure can be given as the following steps:
  
  {\bf Step 1 (Local Update)}: For $m$th user, it calculates the local gradient parameter $\nabla F_m(\bm{w}^t)$ from its local dataset $\mathcal{S}_m$ and global model parameter $\bm{w}^t$.
  
  {\bf Step 2 (Aggregation and Broadcasting)}: Each user sends its local gradient parameter $\nabla F_m(\bm{w}^t)$ to the central server. The central server then aggregates all the local gradient vectors into a common one, denoted as $\bm{w}^{t+1}$, as follows, 
  \begin{equation}
    \bm{w}^{t+1} = \bm{w}^t - \eta \cdot \sum_{m = 1}^{M} \nabla F_m(\bm{w}^t),
  \end{equation}
then the central server broadcasts $\bm{w}^{t+1}$ to all the users. 

With some combination of previously claimed assumptions, the above iterative operation may help to reach the optimal or convergent solution of Problem \ref{pro:min_F} if each user sends trustworthy message to the central server. However, under Byzantine attacks, the main challenge of solving Problem \ref{pro:min_F} comes from the fact that the Byzantine attackers can collude and send arbitrary malicious messages to the central server so as to bias the optimization process. We aspire to develop a robust FL algorithm to address this issue in next section.

%% file: raga.tex
\section{Proposed Aggregation Strategy: Robustness Average Gradient Algorithm (RAGA)} \label{sec:RAGA}

Before the introduction of the RAGA, we first explain the scenario of Byzantine attack. 
Assume there are $B$ Byzantine users out of $M$ users, which compose the set of $\mathcal{B}$.   
Any Byzantine user can send an arbitrary vector $\star , \star \in \mathbb{R}^p$ to the central server. Suppose $\bm{z}_m^t$ is the real vector uploaded by $m$th user to the central server in $t$th round of iteration, then there is
{
\begin{equation}
  \bm{z}_m^t = \left\{
  \begin{aligned}
    &{\bm{g}_m^t, }  &       &{m \in \mathcal{M} \setminus \mathcal{B}}\\
    &{\bm{\star}, }       &        &{m \in \mathcal{B} }
  \end{aligned}  
  \right. , 
\end{equation}}
where $g_m^t$ denotes the vector to be uploaded by honest user ($m \in \mathcal{M} \setminus \mathcal{B}$). 
Additionally, we also assume $\sum_{m \in \mathcal{B}} S_m < \frac{1}{2} \sum_{m \in \mathcal{M}} S_m$. 
In related works {\cite{wu2020federated}}, by implicitly supposing $S_m$ to be identical for every $m\in \mathcal{M}$, it is usually assumed $B < \frac{1}{2} M$, which is a special realization of our assumption.

To reduce communication burden due to frequent interactions between $M$ users and the central server and overcome the Byzantine attacks, we propose the RAGA, which runs multi-local updates and aggregates all the uploaded vectors in a robust way. Specifically, in $t$th round of iteration:
  
  {\bf Step 1 (Multi-Local Update)}: For any honest user $m$, $m \in \mathcal{M} \setminus \mathcal{B}$. Denote $\bm{w}_m^{t, k}$ as the local model parameter of $m$th user for $k$th update and $\eta_m^{t, k}$ as the associated learning rate. By implementing stochastic gradient descend (SGD) method, there is 
  {
  \begin{align}
    \bm{w}_m^{t, k} = \bm{w}_m^{t, k-1} - \eta_m^{t, k} \cdot \nabla F_m(\bm{w}_m^{t, k-1};\xi_m^{t, k}), \nonumber \\
      k = 1, 2, \dots, K^t, 
  \end{align}}
  where $\xi_m^{t, k}$ is a randomly selected subset of dataset $\mathcal{S}_m$ and $\bm{w}_m^{t, 0} = \bm{w}^t$. The $\bm{w}^t$ is received from $t-1$ round of iteration. 
 Specially, for the sake of robustness and convergence, we set $\bm{z}_m^t = \bm{g}_m^t = \frac{1}{K^t} \sum_{k=1}^{K^t} \nabla F_m(\bm{w}_m^{t, k-1};\xi_m^{t, k})$ for $m \in \mathcal{M} \setminus \mathcal{B}$. 
  
  {\bf Step 2 (Aggregation and Broadcasting)}: For the $m$th user, it uploads its own vector $\bm{z}_m^t$ to the central server for $m \in \mathcal{M}$. To ensure the robustness of the FL system, we use the numerical geometric median of all the uploaded vectors, which is defined as
  {
  \begin{align}
    \bm{w}^{t+1}
    &=\bm{w}^{t} -\eta^t \cdot \bm{z}^t,   \label{equ:geom}
  \end{align}}
where $\eta^t$ represents the global learning rate for $t$th round of iteration and $\bm{z}^t$ represents the numerical estimation of the geometric median of $\{\bm{z}_m^t| m\in \mathcal{M}\}$, i.e., 
${\rm geomed} \left(\{\bm{z}_m^t| m\in \mathcal{M}\}\right)$.
The ${\rm geomed} \left(\{\bm{z}_m^t| m\in \mathcal{M}\}\right)$ in this paper is defined as  
  \begin{align} \label{e:geomedian_def}
   \bm{z}^{t, *} 
   &\triangleq {\rm geomed} \left(\{\bm{z}_m^t| m\in \mathcal{M}\}\right) \nonumber \\ 
   &= \arg \min_{y} \sum_{i=1}^{M}\frac{S_i}{\sum_{j=1}^{M}S_j} \left\lVert y - \bm{z}_i^t \right\rVert. 
  \end{align}
The problem in (\ref{e:geomedian_def}) is actually a convex optimization problem and an $\epsilon$-optimal solution is obtainable by resorting to Weiszfeld algorithm \cite{aftab2014generalized}. Hence the $\bm{z}^t$ fulfills the following condition
  \begin{align}
    \sum_{i=1}^{M} \frac{S_i}{\sum_{j=1}^{M}S_j} \left\lVert \bm{z}^t - \bm{z}_i^t \right\rVert 
    \leqslant \sum_{i=1}^{M} \frac{S_i}{\sum_{j=1}^{M}S_j} \left\lVert \bm{z}^{t, *} - \bm{z}_i^t \right\rVert + \epsilon, 
  \end{align}
and is also denoted as ${\rm geomed} \left(\{\bm{z}_m^t| m\in \mathcal{M}\}, \epsilon \right)$ in the sequel. Then the central server broadcasts the global model parameter $\bm{w}^{t+1}$ to all the users for preparing the calculation in $t+1$ round of iteration. 

\begin{figure}[htbp]
  \centering
  \includegraphics[angle=0, width=0.8 \columnwidth]{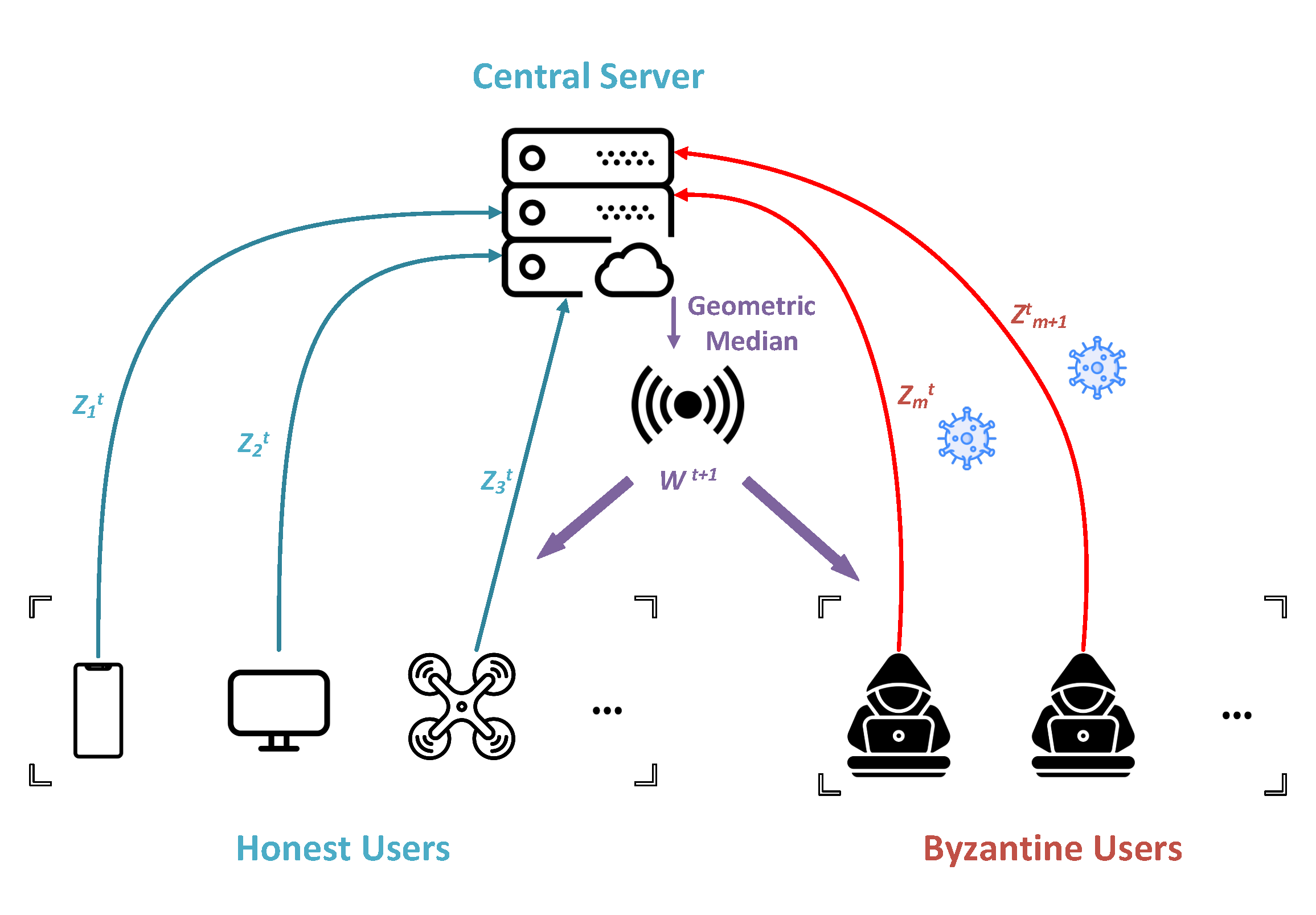}
  \caption{Illustration of Byzantine-resilient RAGA FL system.} 
  \label{fig:model}
\end{figure}

Based on the above description, the RAGA  is conceptually illustrated in Fig \ref{fig:model} and summarized in Algorithm \ref{alo:raga}.

\begin{algorithm}[!tbp]
  \caption{The procedure of RAGA.} \label{alo:raga}
The central server initializes the model parameter $\bm{w}^1$ and sends it to all the users. \\
  \For{$t = 1, 2, \ldots , T$}{
    \For{$m \in \mathcal{M} \setminus \mathcal{B}$}{
    Set $\bm{w}_m^{t, 0} = \bm{w}^t$.  \\
      \For{$k = 1, 2, \ldots, K^t$}{
        Randomly select local data subset $\xi_m^{t, k}$ from $\mathcal{S}_m$ and set $\bm{w}_m^{t, k} = \bm{w}_m^{t, k - 1} - \eta_m^{t, k} \nabla F_m(\bm{w}_m^{t, k - 1};\xi_m^{t, k})$.
      }
      Set $\bm{z}_m^t = \frac{1}{K^t} \sum_{k=1}^{K^t} \nabla F_m(\bm{w}_m^{t, k-1};\xi_m^{t, k}).$
    }
     \For{$m \in \mathcal{B}$}
     {
     Set $z_m^t = \bm{\star}$.
     }
{All the users send their own $\bm{z}_m^t$ to the central server.} The central server {obtains the geometric median $\bm{z}^t = {\rm geomed} \left(\{\bm{z}_m^t| m\in \mathcal{M}\}, \epsilon \right)$}, updates $\bm{w}^{t+1}$ by 
     $ \bm{w}^{t+1} 
      = \bm{w}^{t} - \eta^t \cdot \bm{z}^t$,  
and then broadcasts $\bm{w}^{t + 1}$ to all the users. 
  } 
\end{algorithm}

%% file: convergence_analysid.tex
\section{Analysis of Convergence and Robustness} \label{sec:conver}

In this section, we provide theoretical analysis of the robustness and convergence of the RAGA under Byzantine attacks under two cases of loss function's convexity. Below, we first present the necessary assumptions.

\subsection{Assumption}

For all honest users $m \in \mathcal{M} \setminus \mathcal{B}$, we have the following assumptions. 

\begin{assumption}[Lipschitz Continuity] \label{ass:lips}
  The loss function $f(\bm{w}, \bm{s})$ has $L$-Lipschitz continuity \cite{huang2023achieving, li2022analysis}, i.e., for $\forall \bm{w}_1, \bm{w}_2 \in \mathbb{R}^p$, there is
  \begin{align} \label{e:ass_lips_1}
    f(\bm{w}_1, \bm{s}) - f(\bm{w}_2, \bm{s}) 
    &\leqslant \left\langle \nabla f(\bm{w}_2, \bm{s}), \bm{w}_1 - \bm{w}_2 \right\rangle \nonumber \\ 
    &\qquad \qquad \quad + \frac{L}{2} \left\lVert \bm{w}_1 - \bm{w}_2 \right\rVert^2, 
  \end{align}
which is equivalent to the following inequality, 
  \begin{align} \label{e:ass_lips_2}
    \left\lVert \nabla f(\bm{w}_1, \bm{s}) - \nabla f(\bm{w}_2, \bm{s}) \right\rVert \leqslant L \left\lVert \bm{w}_1 - \bm{w}_2 \right\rVert .
  \end{align}
With (\ref{e:ass_lips_1}) or (\ref{e:ass_lips_2}), it can be derived that the global loss function $F(\bm{w})$ and local loss function $F_m(\bm{w})$ both have $L$-Lipschitz continuity. 
\end{assumption}

\begin{assumption}[{Strongly-Convex}] \label{ass:convex}
  The loss function $f(\bm{w}, \bm{s})$ is $\mu$ strongly-convex \cite{huang2023achieving}, i.e., for $\forall \bm{w}_1, \bm{w}_2 \in \mathbb{R}^p$, there is
  \begin{equation} \label{e:ass_convex}
    \left\lVert \nabla f(\bm{w}_1, \bm{s}) - \nabla f(\bm{w}_2, \bm{s}) \right\rVert \geqslant \mu \left\lVert \bm{w}_1 - \bm{w}_2 \right\rVert.
  \end{equation}
Similarly, with (\ref{e:ass_convex}), $F(\bm{w})$ and $F_m(\bm{w})$ can be derived to be both $\mu$ strongly-convex.
\end{assumption}

\begin{assumption}[Local Unbiased Gradient] \label{ass:unbias}
  Suppose a subset of dataset $\mathcal{S}_m$ is selected randomly, which is denote as $\xi_m$, define
  \begin{equation}
    F_m(\bm{w};\xi_m)  \triangleq \frac{1}{\left\lvert \xi_m \right\rvert} \sum_{\bm{s} \in \xi_m} f(\bm{w}, \bm{s}), 
  \end{equation}
  the unbiased gradient \cite{huang2023achieving, li2022analysis} implies that
  \begin{equation}
    \mathbb{E}\left\{ \nabla F_m(\bm{w};\xi_m) \right\} = \nabla F_m(\bm{w}).
  \end{equation}
\end{assumption}
\begin{assumption}[Bounded Inner Variance] \label{ass:var}
  For each honest user $m$ and $\bm{w} \in \mathbb{R}^p$, the variance of stochastic local gradient $\nabla F_m(\bm{w};\xi_m)$ is upper-bounded \cite{huang2023achieving, li2022analysis} by
  \begin{align}
    \text{Var} (\nabla F_m(\bm{w};\xi_m)) 
    \leqslant \sigma^2.
  \end{align}
\end{assumption}

\begin{assumption}[Data Heterogeneity] \label{ass:hete}
    For each honest user $m$ and $\bm{w} \in \mathbb{R}^p$, the data heterogeneity can be defined as
    \begin{align}
        \theta_m = \left\lVert \nabla F_m(\bm{w}) - \nabla F(\bm{w}) \right\rVert. 
    \end{align}
    We also assume the data heterogeneity is bounded \cite{huang2023achieving, li2022analysis}, which implies 
    \begin{align}
        \theta_m \leqslant \theta. 
    \end{align}
\end{assumption}

\begin{assumption}[Bounded Gradient] \label{ass:boundedg}
  For each honest user $m$ and $\bm{w} \in \mathbb{R}^p$, the ideal local gradient $\nabla F_m(\bm{w})$ is upper-bounded \cite{huang2023achieving, li2022analysis} by 
  \begin{align}
    \left\lVert \nabla F_m(\bm{w}) \right\rVert \leqslant G.
  \end{align}
\end{assumption}

\subsection{Convergence Analysis}

In this subsection, we present the analytical results of convergence on our proposed RAGA. All proofs are deferred to Appendices B and C. 

\subsubsection{Case \Rmnum{1}: With Lipschitz Continuity Only for Loss Function}

We first inspect the case with  Assumptions \ref{ass:lips}, \ref{ass:unbias}, \ref{ass:var}, \ref{ass:hete} and \ref{ass:boundedg} imposed, which allows the loss function to be non-convex. 
By defining  
 \begin{align} \label{def:rato}
    \alpha_m = \frac{S_m}{\sum_{j=1}^{M} S_j}, C_{\alpha} = \sum_{m \in \mathcal{M} \setminus \mathcal{B}} \alpha_m, 
  \end{align}
  and 
  \begin{equation} \label{e:p_t_def}
    p^t = \left\{
    \begin{aligned}
      &{0, }  &       &{\eta^t \leqslant \frac{1}{L}}\\
      &{1, }  &       &{\eta^t > \frac{1}{L}}
    \end{aligned} 
    \right. , 
  \end{equation}
the following results can be expected.
\begin{theorem} \label{theo:smooth}
  With {$C_{\alpha} > 0.5$}, there is 
  {
  \begin{align} \label{equa:smooth}
    &\quad \sum_{t=1}^{T} \frac{\eta^t}{\sum_{{t'}=1}^{T} \eta^{t'}} \mathbb{E}\left\{ \left\lVert \nabla F(\bm{w}^t) \right\rVert ^2 \right\} \nonumber \\ 
    &\leqslant \frac{2\mathbb{E} \left\{ F(\bm{w}^{1}) - F(\bm{w}^{T}) \right\}}{\sum_{{t'}=1}^{T} \eta^{t'}} + \frac{1}{\sum_{{t'}=1}^{T} \eta^{t'}} \sum_{t=1}^{T} \eta^t \Delta^t \nonumber \\
    &\quad + \sum_{t=1}^{T} \frac{2(\eta^t + p^t(\eta^t)^2 L - p^t\eta^t)\epsilon^2}{(2C_{\alpha}-1)^2\sum_{{t'}=1}^{T} \eta^{t'} } \nonumber \\
    &\quad + \sum_{t=1}^{T} \frac{8(C_{\alpha})^2(\eta^t\sigma^2+(p^t(\eta^t)^2 L - p^t\eta^t)(G^2+\sigma^2))}{(2C_{\alpha}-1)^2\sum_{{t'}=1}^{T} \eta^{t'}}, 
  \end{align}}
  where
  \begin{align} \label{equ:delta}
    \Delta^t = 
    &\frac{8C_{\alpha}}{(2C_{\alpha}-1)^2} \sum_{m \in \mathcal{M} \setminus \mathcal{B}} \Bigg( \frac{2\alpha_mL^2(G^2 + \sigma^2)}{K^t} \sum_{k=2}^{K^t} (k - 1) \nonumber \\ 
    &\quad \cdot \sum_{i=1}^{k-1} (\eta_m^{t, i})^2 + 2 \alpha_m \theta^2 \Bigg). 
  \end{align}  
\end{theorem}
\begin{IEEEproof}
  Please refer to Appendix B. 
\end{IEEEproof}

\begin{rem} \label{rem:non_convex}
  \begin{itemize}
   \item {In general case, {when $\lim_{T \rightarrow \infty} \sum_{t=1}^{T} \eta^t = \infty$, \\$\lim_{T \rightarrow \infty} \sum_{t=1}^{T} \eta^t \sum_{m \in \mathcal{M} \setminus \mathcal{B}} \sum_{i=1}^{K^t} (\eta_m^{t, i})^2 < \infty $, and there are only finitely many $t$ such that $p^t = 1$}, together with the prerequisite condition of Theorem \ref{theo:smooth}, i.e., $C_{\alpha}>0.5$, it can be derived that the term $\sum_{t=1}^{T} \frac{\eta^t}{\sum_{t'=1}^{T} \eta^t} \mathbb{E}\left\{ \left\lVert \nabla F(\bm{w}^t) \right\rVert ^2 \right\}$ will converge to $\mathcal{O} \left( \theta^2 + \epsilon^2 + \sigma^2 \right)$. 
    \item To be concrete, we set $\eta_m^{t, k} = \eta^t \leqslant \frac{1}{L}$ and $K^t = K$. Then inequality (\ref{equa:smooth}) dwells into 
    {
    \begin{small}
        \begin{align}
        &\quad \sum_{t=1}^{T} \frac{\eta^t}{\sum_{t'=1}^{T} \eta^t} \mathbb{E}\left\{ \left\lVert \nabla F(\bm{w}^t) \right\rVert ^2 \right\} \nonumber \\ 
        &\leqslant \frac{8(C_{\alpha})^2L^2(G^2 + \sigma^2)(K-1)(2K-1)\sum_{t=1}^{T} (\eta^t)^3}{3(2C_{\alpha}-1)^2\sum_{t=1}^{T} \eta^t} \nonumber \\
        &+ \frac{2\mathbb{E} \left\{ F(\bm{w}^{1}) - F(\bm{w}^{T}) \right\}}{\sum_{t=1}^{T} \eta^t}  + \frac{16(C_{\alpha})^2\theta^2 + 2\epsilon^2 + 8(C_{\alpha})^2\sigma^2}{(2C_{\alpha}-1)^2}, 
      \end{align}
    \end{small}}
  which will converge to 
  {
  \begin{equation} \label{e:converge_result_nonconvex}
  \frac{16(C_{\alpha})^2}{(2C_{\alpha}-1)^2}\theta^2 + \frac{2}{(2C_{\alpha}-1)^2}\epsilon^2 + \frac{8(C_{\alpha})^2}{(2C_{\alpha}-1)^2}\sigma^2, 
  \end{equation}}
  as long as $\lim_{T \rightarrow \infty} \sum_{t=1}^{T} (\eta^t)^3 < \infty $.
    \item 
    {
    To further concrete convergence rate, we set $\eta^t = \frac{\eta_0}{L}\frac{1}{(t+c)^{1/3 + \delta}}$, where $\delta$ is an arbitrary value lying between $(0, 2/3)$, $\eta_0<1$ and $c>0$, the right shift introduced by adding $c>0$ to the denominator is intended to mitigate fluctuations during the initial training iterations, then the right-hand side of (21) of the revised manuscript is upper bounded by
    \textcolor{black}{
        \begin{small}
           \begin{align}
              &\mathcal{O} \left( \frac{8\eta_0^2(C_{\alpha})^2(G^2 + \sigma^2)(K-1)(2K-1)(1+3\delta -3\delta T^{-3\delta})}{3(2C_{\alpha}-1)^2 T^{2/3- \delta}} \right) \nonumber \\
              &+ \mathcal{O} \left( \frac{2L\mathbb{E} \left\{ F(\bm{w}^{1}) - F(\bm{w}^{T}) \right\}}{\eta_0 T^{2/3- \delta}} \right) \nonumber \\
              & + \frac{16(C_{\alpha})^2\theta^2 + 2\epsilon^2 + 8(C_{\alpha})^2\sigma^2}{(2C_{\alpha}-1)^2}, 
            \end{align} 
        \end{small}}
    which still converges to the result in (24) at rate $\mathcal{O}\left( \frac{1}{T^{2/3- \delta}} \right)$. }} 
\item From previous inference from Theorem \ref{theo:smooth}, it can be found that the minimum of $\mathbb{E}\left\{ \left\lVert \nabla F(\bm{w}^t) \right\rVert ^2 \right\}$ over $t\in [1, T]$, which is upper bounded by the left-hand side of (\ref{equa:smooth}), and thus the expression in (\ref{e:converge_result_nonconvex}). The non-zero gap shown in (\ref{e:converge_result_nonconvex}) will vanish as $\theta$, $\epsilon$ and $\sigma$, which represent data heterogeneity, error tolerance to numerically work out the geometric median through Weiszfeld algorithm, and the variance of stochastic local gradient respectively,  go to zero. In other words, once the $\mathcal{S}_m$ among $m \in \mathcal{M}$ are IID, $\mathcal{S}_m$ for every $m\in \mathcal{M}$ are fully utilized to calculate local gradient, and the geometric median is calculated perfectly, there will be one $\nabla F(\bm{w}^t)$ for $t\in [1, T]$ being zero, which implies the reaching of stationary point for Problem \ref{pro:min_F}. 
\item Last but not least, the above convergence results are based on the assumption that $C_{\alpha}>0.5$, i.e., the fraction of the datasets from Byzantine attackers is less than half, which shows a strong robustness to Byzantine attacks.
  \end{itemize}
\end{rem}

\subsubsection{Case \Rmnum{2}: With Lipschitz Continuity and Strong Convexity for Loss Function}

For the case with {Assumptions \ref{ass:lips}, \ref{ass:convex}, \ref{ass:unbias}, \ref{ass:var}, \ref{ass:hete} and \ref{ass:boundedg}} imposed, which requires the loss function to be not only Lipschitz but also strongly-convex, the following results can be anticipated.
\begin{theorem} \label{theo:conv}
  With {$0 < \lambda^t <1$} and $C_{\alpha} > 0.5$, the optimality gap $\mathbb{E} \left\{ F(\bm{w}^{T}) - F(\bm{w}^*) \right\}$ is upper bounded in (\ref{equa:gene}) as follows, 
  {
  \begin{align} \label{equa:gene}
    &\quad \mathbb{E} \left\{ F(\bm{w}^{T}) - F(\bm{w}^*) \right\} \nonumber \\
    &\leqslant \frac{L}{2} \mathbb{E} \left\{ \left\lVert \bm{w}^1  - \bm{w}^{*} \right\rVert^2 \right\} \prod_{t = 1}^{T-1} \gamma^t + \frac{L}{2} \sum_{t=1}^{T-1} \frac{(\eta^t)^2}{\lambda^t} \Bigg( \Delta^t +  \nonumber \\ 
    &\quad \frac{8\sigma^2(C_{\alpha})^2}{(2C_{\alpha}-1)^2} + \frac{2\epsilon^2}{\left( 2C_{\alpha}-1 \right)^2} \Bigg) \cdot \prod_{i=t}^{T-1} (\gamma^{i+1})^{q^{i+1}}, 
  \end{align} }
  where 
  \begin{equation}
    q^t = \left\{
    \begin{aligned}
      &{0, }  &       &{t = T}\\
      &{1, }  &       &{t < T}
    \end{aligned}  \nonumber
    \right. , 
  \end{equation}
  and
  \begin{align}
    \gamma^t = \frac{1 -2\eta^t\mu+L^2(\eta^t)^2}{1-\lambda^t}. \nonumber
  \end{align}
\end{theorem}
\begin{IEEEproof}
  Please refer to Appendix C. 
\end{IEEEproof}
\begin{rem}
  \begin{itemize}
      \item For general case, {as long as $\eta_m^{t, k} \propto T^{-\beta}, \beta > 0$, $\lambda^t = \rho^t \eta^t$ with $\rho^t < 2\mu$ and $\eta^t < \frac{2\mu-\rho^t}{L^2}$, $0< \gamma^t <1$, and the preconditions listed in Theorem \ref{theo:conv}} hold, the convergence of our proposed RAGA  can be ensured and the optimality gap is at the scale $\mathcal{O} \left( \theta^2 + \epsilon^2 + \sigma^2 \right)$.
    \item To concrete the optimality gap, we suppose $\eta_m^{t, k} = \frac{1}{T}\eta$, $\eta^t = \eta$, $\gamma^t = \gamma$, $K^t = K$, $\lambda^t = \mu \eta^t = \mu \eta$, and $\eta^t < \frac{\mu}{L^2}$, then there is
    \begin{align} 
      &\quad \mathbb{E} \left\{ F(\bm{w}^{T}) - F(\bm{w}^*) \right\} \nonumber \\
      &\leqslant \frac{L (\gamma)^{T-1}}{2} \mathbb{E} \left\{ \left\lVert \bm{w}^1 - \bm{w}^{*} \right\rVert^2 \right\}  \nonumber \\
      & + \frac{4L^3(K-1)(2K-1)(C_{\alpha})^2(G^2+\sigma^2)}{3\mu(2C_{\alpha}-1)^2T^2} \frac{1-(\gamma)^T}{1-\gamma} (\eta)^3  \nonumber \\
      & + \frac{8L(C_{\alpha})^2\theta^2 + \epsilon^2L + 4L(C_{\alpha})^2\sigma^2}{\mu(2C_{\alpha}-1)^2} \frac{1-(\gamma)^T}{1-\gamma} \eta. 
    \end{align}
 When $T \rightarrow \infty$, $(\gamma)^{T-1} \rightarrow 0$ and $\frac{1 - (\gamma)^T}{1 - \gamma} \rightarrow \frac{1}{1 - \gamma} = \frac{1 - \mu\eta}{\mu\eta - L^2(\eta)^2}< \infty$. 
 Therefore, $\frac{1 - (\gamma)^T}{T^2 (1 - \gamma)} \rightarrow \frac{1 - \mu\eta}{T^2(\mu\eta - L^2(\eta)^2)} \rightarrow 0$ and $\frac{\eta(1 - (\gamma)^T)}{1 - \gamma} \rightarrow  \frac{1 - \mu \eta}{\mu - L^2\eta}$. 
 In this case, $\mathbb{E} \left\{ F(\bm{w}^{T}) - F(\bm{w}^*) \right\}$ will converge to 
 {
\begin{align}\label{e:convergence_convex}
    \frac{1 - \mu \eta}{\mu - L^2\eta}\left(\frac{8L(C_{\alpha})^2}{\mu(2C_{\alpha}-1)^2}\theta^2 + \frac{L}{\mu(2C_{\alpha}-1)^2}\epsilon^2 \right. \nonumber \\ 
    \left. + \frac{4L(C_{\alpha})^2}{\mu(2C_{\alpha}-1)^2}\sigma^2\right), 
\end{align}}

at a rate $\mathcal{O}\left((\gamma)^{T-1} + \frac{1}{T^2}\right)$.
    \item To further concrete convergence rate, we alternatively set $\eta_m^{t, k} = (TK)^{-\beta}\eta$ with $\beta > 0$, then the optimality gap of (\ref{equa:gene}) can be bounded by
    {
    \begin{align}
      &\left( \frac{4L^3(K-1)(2K-1)(C_{\alpha})^2(G^2+\sigma^2)}{3\mu(2C_{\alpha}-1)^2(TK)^{2\beta}} \frac{1-(\gamma)^T}{1-\gamma} (\eta)^3 \right) \nonumber \\ 
      &+ \left(\frac{L(\gamma)^{T-1}}{2} \mathbb{E} \left\{ \left\lVert \bm{w}^1 - \bm{w}^{*} \right\rVert^2 \right\}\right) \nonumber \\ 
      &+ \frac{8L(C_{\alpha})^2\theta^2 + \epsilon^2L + 4L(C_{\alpha})^2\sigma^2}{\mu(2C_{\alpha}-1)^2} \frac{1-(\gamma)^T}{1-\gamma} \eta, 
\end{align}}
which also convergences to (\ref{e:convergence_convex}) but at a rate {$\mathcal{O}\left((\gamma)^{T-1} + (TK)^{-2 \beta}\right)$}. This rate can be adjusted by $\beta$ and would be linear when $\beta \propto T$. 
    \item Compared with the discussion in Remark \ref{rem:non_convex} that works for non-convex loss function, we can obtain zero optimality gap as $\theta$, $\epsilon$, and $\sigma$ go to zero for strongly-convex loss function in this case, which implies the achievement of global optimal solution rather than stationary point as in Remark \ref{rem:non_convex}. Moreover, linear convergence rate is achievable in this case and is faster than the rate $\mathcal{O}\left(\frac{1}{T^{2/3- \delta}}\right)$ as shown in Remark \ref{rem:non_convex}. 
    \item It is also worthy to note that the above results still hold on the condition that $C_{\alpha}>0.5$, which promises a strong robustness to Byzantine attacks.
  \end{itemize}
\end{rem}

%% file: experiments.tex
\section{Experiments} \label{sec:num}

\begin{table*}[t]
\caption{\centering The maximum test accuracy(\%) of 500 iteration number of RAGA, FedAvg, Median, Byrd-SAGA, RFA and MCA with different types, ratios of Byzantine attack, and concentration parameter on MNIST dataset and LeNet model. } 
\label{tab:accml}
\resizebox{\textwidth}{!}{
\renewcommand{\arraystretch}{1.5}
\begin{tabular}{c|cc|c|cccc|cccc|cccc}
\multirow{3}{*}{Model}           & \multicolumn{2}{c|}{Attack Name}                       & \textbf{No Attack} & \multicolumn{4}{c|}{\textbf{Gaussian Attack}}                                             & \multicolumn{4}{c|}{\textbf{Sign-flip Attack}}                                            & \multicolumn{4}{c}{\textbf{LIE Attack}}                                                   \\ \cline{2-16} 
                                 & \multicolumn{2}{c|}{$\bar{C}_{\alpha}$}                & \multirow{2}{*}{0} & \multirow{2}{*}{0.1} & \multirow{2}{*}{0.2} & \multirow{2}{*}{0.3} & \multirow{2}{*}{0.4} & \multirow{2}{*}{0.1} & \multirow{2}{*}{0.2} & \multirow{2}{*}{0.3} & \multirow{2}{*}{0.4} & \multirow{2}{*}{0.1} & \multirow{2}{*}{0.2} & \multirow{2}{*}{0.3} & \multirow{2}{*}{0.4} \\ \cline{2-3}
                                 & \multicolumn{1}{c|}{$\phi$}               & Algorithms &                    &                      &                      &                      &                      &                      &                      &                      &                      &                      &                      &                      &                      \\ \hline
\multirow{12}{*}{\textbf{LeNet}} & \multicolumn{1}{c|}{\multirow{6}{*}{0.6}} & RAGA       & \textbf{97.89}     & \textbf{97.78}       & \textbf{97.79}       & \textbf{97.82}       & \textbf{97.75}       & \textbf{97.72}       & \textbf{97.76}       & \textbf{97.69}       & \textbf{97.25}       & \textbf{97.79}       & \textbf{97.82}       & \textbf{97.80}       & \textbf{97.71}       \\
                                 & \multicolumn{1}{c|}{}                     & FedAvg     & 95.54              & 10.28                & 11.35                & 11.35                & 11.35                & 94.03                & 9.80                 & 9.82                 & 9.80                 & 95.10                & 94.96                & 93.66                & 92.63                \\
                                 & \multicolumn{1}{c|}{}                     & Median     & 94.41              & 94.31                & 94.22                & 94.53                & 94.18                & 94.12                & 94.34                & 93.69                & 93.45                & 94.33                & 94.35                & 94.60                & 94.22                \\
                                 & \multicolumn{1}{c|}{}                     & Byrd-SAGA       & 91.70              & 91.66                & 91.61                & 91.77                & 91.26                & 91.23                & 91.54                & 91.13                & 89.95                & 91.64                & 91.66                & 91.87                & 91.30                \\
                                 & \multicolumn{1}{c|}{}                     & RFA        & 95.70              & 95.52                & 95.53                & 95.74                & 95.41                & 95.39                & 95.05                & 95.23                & 94.85                & 95.54                & 95.48                & 95.51                & 95.26                \\
                                 & \multicolumn{1}{c|}{}                     & MCA        & 91.82              & 91.89                & 91.88                & 92.16                & 91.77                & 90.77                & 19.02                & 83.10                & 18.75                & 91.86                & 91.94                & 92.14                & 91.76                \\ \cline{2-16} 
                                 & \multicolumn{1}{c|}{\multirow{6}{*}{0.2}} & RAGA       & \textbf{97.62}     & \textbf{97.47}       & \textbf{97.67}       & \textbf{97.50}       & \textbf{97.30}       & \textbf{97.42}       & \textbf{97.16}       & \textbf{97.28}       & \textbf{97.38}       & \textbf{97.75}       & \textbf{97.64}       & \textbf{97.60}       & \textbf{97.28}       \\
                                 & \multicolumn{1}{c|}{}                     & FedAvg     & 95.13              & 9.80                 & 11.35                & 10.32                & 9.80                 & 10.10                & 10.10                & 10.09                & 9.80                 & 94.81                & 94.38                & 93.74                & 92.09                \\
                                 & \multicolumn{1}{c|}{}                     & Median     & 90.81              & 91.88                & 91.43                & 91.03                & 91.91                & 89.85                & 89.05                & 90.28                & 88.28                & 91.68                & 91.20                & 90.90                & 92.00                \\
                                 & \multicolumn{1}{c|}{}                     & Byrd-SAGA       & 90.36              & 90.76                & 90.24                & 89.81                & 89.77                & 89.51                & 87.95                & 89.28                & 88.29                & 90.90                & 90.35                & 89.69                & 89.82                \\
                                 & \multicolumn{1}{c|}{}                     & RFA        & 95.09              & 95.23                & 95.06                & 94.99                & 94.61                & 93.67                & 93.67                & 93.87                & 94.32                & 95.15                & 95.07                & 94.74                & 94.28                \\
                                 & \multicolumn{1}{c|}{}                     & MCA        & 90.96              & 91.44                & 91.04                & 90.85                & 90.72                & 90.04                & 85.89                & 36.78                & 9.80                 & 91.50                & 90.99                & 90.67                & 90.79                    
\end{tabular}
}
\end{table*}

\begin{figure*}[h]
    \centering
    \subfigure[LeNet and $\bar{C}_{\alpha} = 0.2$.]{
    \includegraphics[width = 0.2\textwidth]{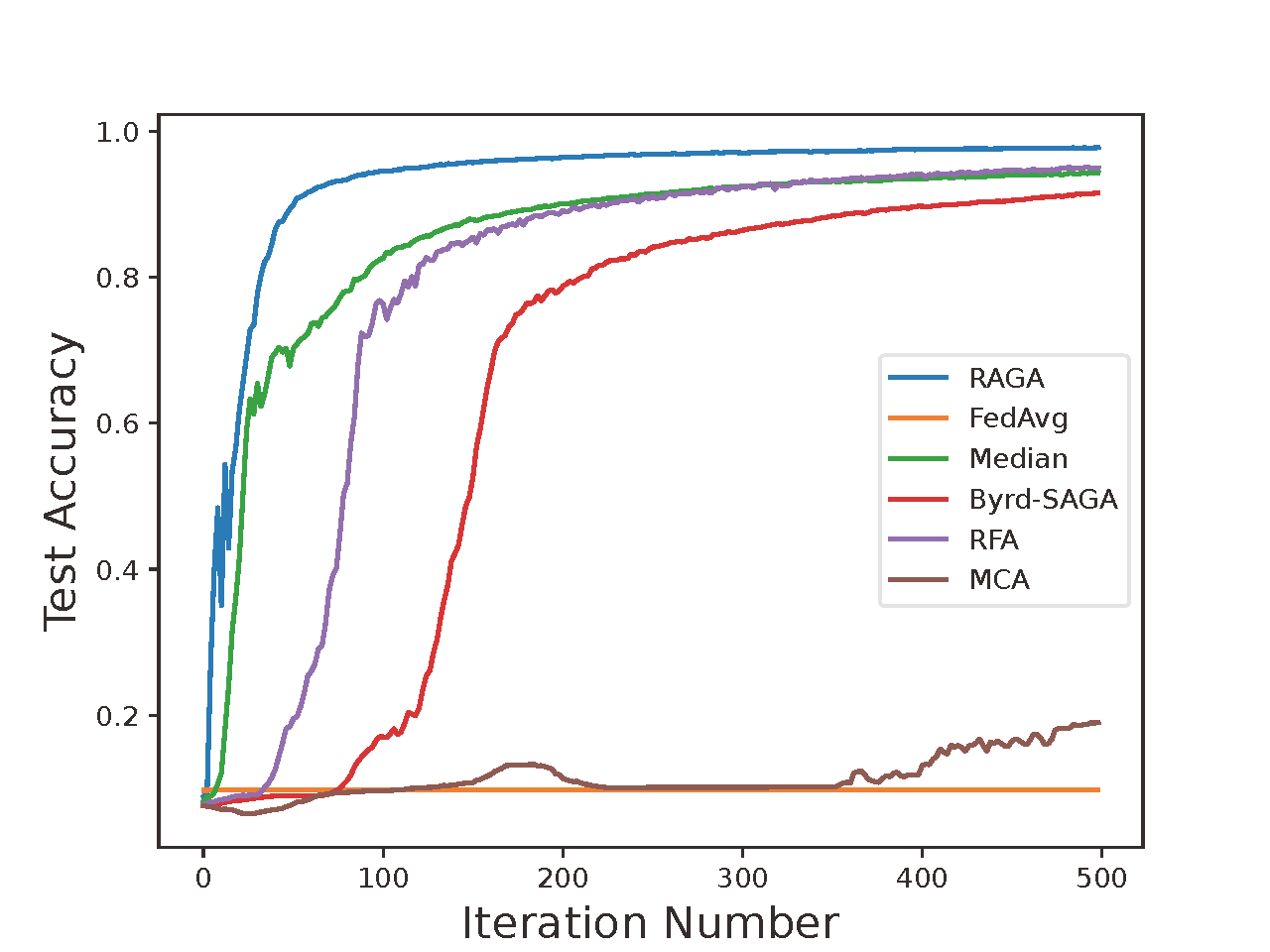}
    }
    \subfigure[LeNet and $\bar{C}_{\alpha} = 0.4$.]{
    \includegraphics[width = 0.2\textwidth]{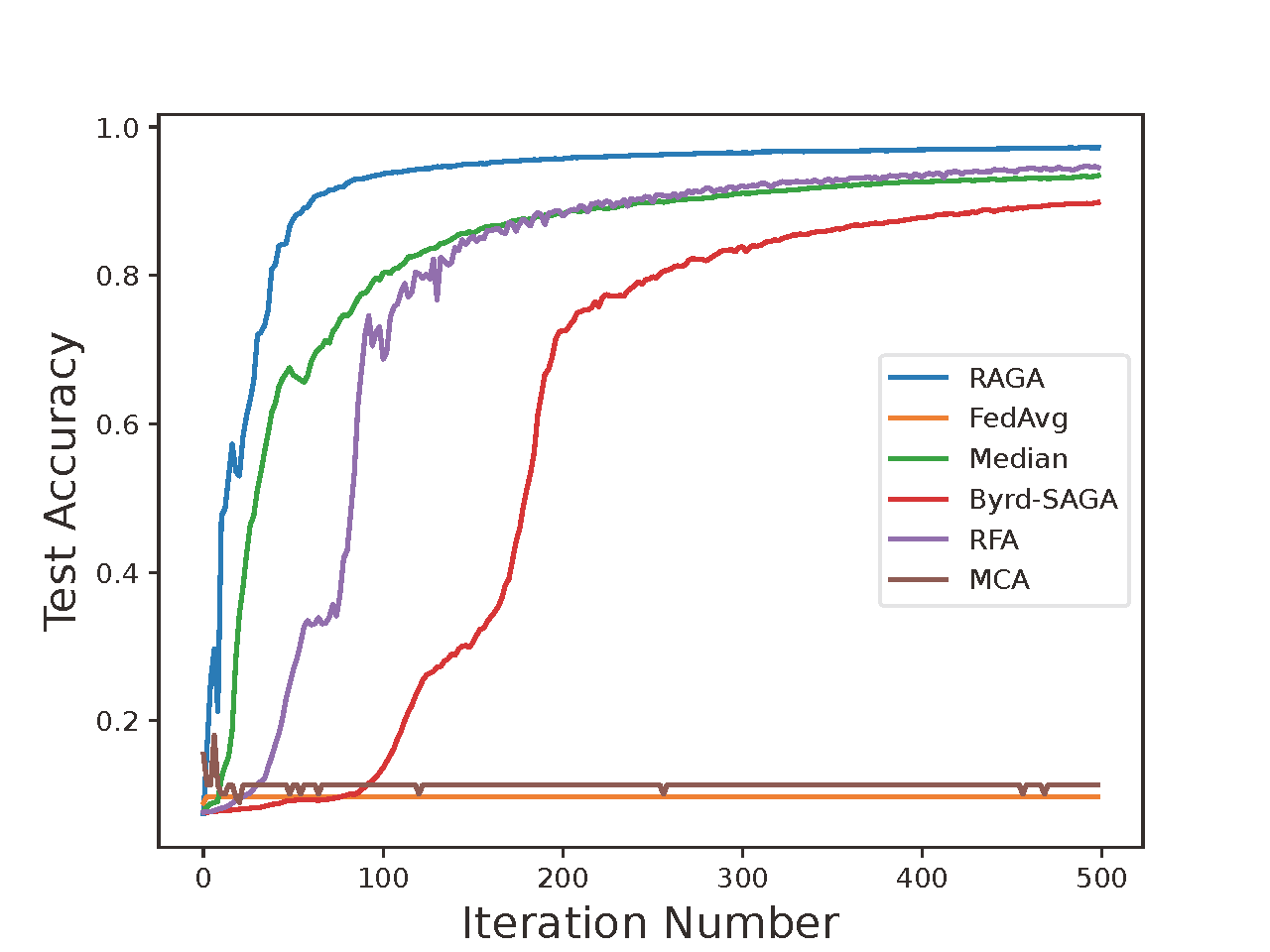}
    }
    \subfigure[MLP and $\bar{C}_{\alpha} = 0.2$.]{
    \includegraphics[width = 0.2\textwidth]{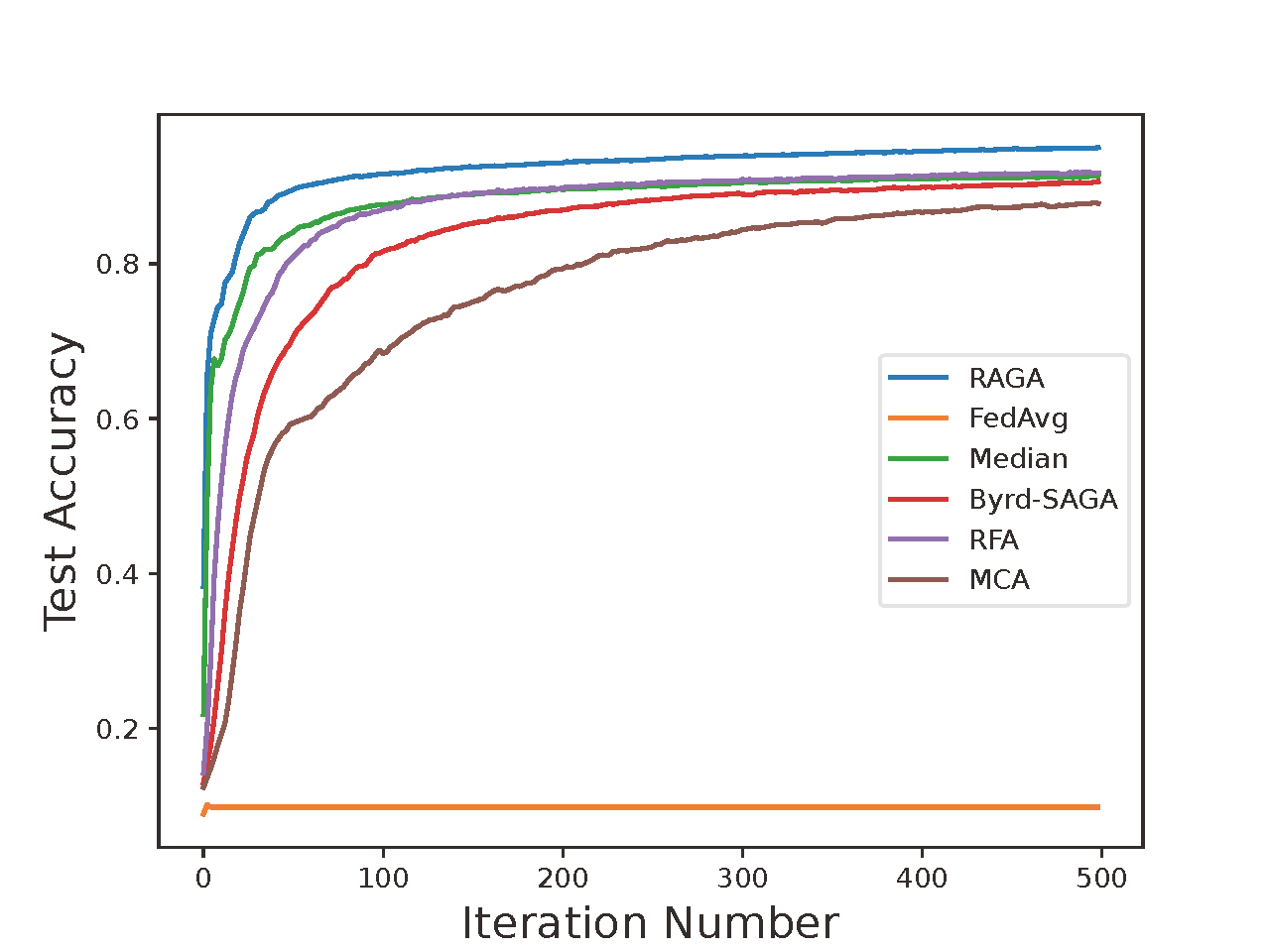}
    }
    \subfigure[MLP and $\bar{C}_{\alpha} = 0.4$.]{
    \includegraphics[width = 0.2\textwidth]{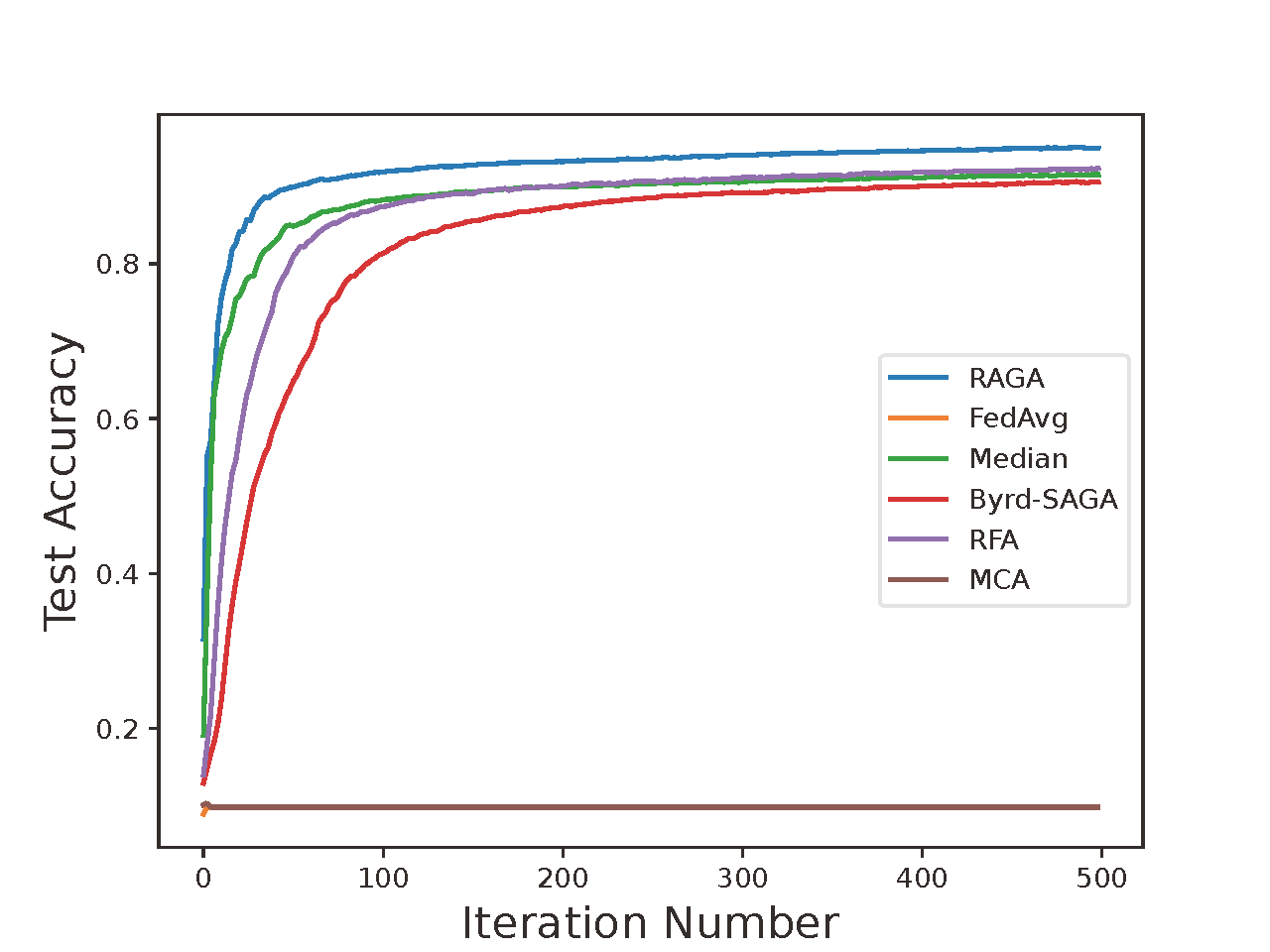}
    }
    \caption{Test accuracy of RAGA and baselines on $\phi = 0.6$, Sign-flip attack, and MNIST dataset. }\label{fig:mta6}
\end{figure*}

In this section, we conduct experiments to inspect the convergence performance of our proposed RAGA under various system setups together with baseline methods, so as to verify the advantage of RAGA over peers in being robust to Byzantine attacks in front of heterogeneous dataset.

\subsection{Setup}

To carry out experiments, we set up a machine learning environment in PyTorch 2.3.1 on Ubuntu 20.04, powered by two 3090 GPUs and two Intel Xeon Gold 6226R CPUs. 

\textbf{Datasets:} 
\begin{itemize}
\item \textbf{MNIST:} MNIST dataset includes a training set and a test set. The training set contains 60000 samples and the test set contains 10000 samples, each sample of which is a 28 × 28 pixel grayscale image. 
\item \textbf{CIFAR10:} The CIFAR10 dataset includes a training set and a test set. The training set contains 50,000 samples, and the test set contains 10,000 samples, each of which is a 32 × 32 pixel color image. 
\item {\textbf{CIFAR100:} The CIFAR100 dataset includes a training set and a test set. The training set contains 50,000 samples, and the test set contains 10,000 samples, each of which is a 32 × 32 pixel color image.} 
\end{itemize}

\begin{table*}[t]
\caption{\centering The maximum test accuracy(\%) of 500 iteration number of RAGA, FedAvg, Median, Byrd-SAGA, RFA and MCA with different types, ratios of Byzantine attack, and concentration parameter on MNIST dataset and MLP model. } 
\label{tab:accmm}
\resizebox{\textwidth}{!}{
\renewcommand{\arraystretch}{1.5}
\begin{tabular}{c|cc|c|cccc|cccc|cccc}
\multirow{3}{*}{Model}           & \multicolumn{2}{c|}{Attack Name}                       & \textbf{No Attack} & \multicolumn{4}{c|}{\textbf{Gaussian Attack}}                                             & \multicolumn{4}{c|}{\textbf{Sign-flip Attack}}                                            & \multicolumn{4}{c}{\textbf{LIE Attack}}                                                   \\ \cline{2-16} 
                                 & \multicolumn{2}{c|}{$\bar{C}_{\alpha}$}                & \multirow{2}{*}{0} & \multirow{2}{*}{0.1} & \multirow{2}{*}{0.2} & \multirow{2}{*}{0.3} & \multirow{2}{*}{0.4} & \multirow{2}{*}{0.1} & \multirow{2}{*}{0.2} & \multirow{2}{*}{0.3} & \multirow{2}{*}{0.4} & \multirow{2}{*}{0.1} & \multirow{2}{*}{0.2} & \multirow{2}{*}{0.3} & \multirow{2}{*}{0.4} \\ \cline{2-3}
                                 & \multicolumn{1}{c|}{$\phi$}               & Algorithms &                    &                      &                      &                      &                      &                      &                      &                      &                      &                      &                      &                      &                      \\ \hline
\multirow{12}{*}{\textbf{MLP}}   & \multicolumn{1}{c|}{\multirow{6}{*}{0.6}} & RAGA       & \textbf{95.23}     & \textbf{95.10}       & \textbf{95.06}       & \textbf{95.01}       & \textbf{94.98}       & \textbf{94.84}       & \textbf{95.00}       & \textbf{94.23}       & \textbf{95.01}       & \textbf{95.08}       & \textbf{95.16}       & \textbf{95.02}       & \textbf{94.92}       \\
                                 & \multicolumn{1}{c|}{}                     & FedAvg     & 92.09              & 23.10                & 24.69                & 25.17                & 9.84                 & 89.41                & 10.10                & 9.80                 & 10.32                & 91.92                & 90.44                & 87.98                & 85.85                \\
                                 & \multicolumn{1}{c|}{}                     & Median     & 91.48              & 91.38                & 91.63                & 91.29                & 91.24                & 91.23                & 91.32                & 90.94                & 91.55                & 91.39                & 91.46                & 91.26                & 91.27                \\
                                 & \multicolumn{1}{c|}{}                     & Byrd-SAGA       & 90.64              & 90.69                & 90.70                & 90.49                & 90.63                & 90.46                & 90.62                & 89.93                & 90.62                & 90.64                & 90.69                & 90.59                & 90.57                \\
                                 & \multicolumn{1}{c|}{}                     & RFA        & 92.38              & 92.31                & 92.41                & 92.04                & 92.27                & 92.10                & 91.86                & 90.81                & 92.33                & 92.36                & 92.46                & 92.05                & 92.18                \\
                                 & \multicolumn{1}{c|}{}                     & MCA        & 90.62              & 90.66                & 90.77                & 90.52                & 90.63                & 90.36                & 87.86                & 60.77                & 10.32                & 90.66                & 90.69                & 90.56                & 90.70                \\ \cline{2-16} 
                                 & \multicolumn{1}{c|}{\multirow{6}{*}{0.2}} & RAGA       & \textbf{94.53}     & \textbf{94.45}       & \textbf{94.69}       & \textbf{94.16}       & \textbf{94.32}       & \textbf{94.09}       & \textbf{94.43}       & 93.84       & \textbf{94.72}       & \textbf{94.35}       & \textbf{94.70}       & \textbf{94.16}       & \textbf{94.38}       \\
                                 & \multicolumn{1}{c|}{}                     & FedAvg     & 91.84              & 21.48                & 19.66                & 30.05                & 12.34                & 10.10                & 11.98                & 9.80                 & 10.09                & 91.48                & 90.27                & 87.50                & 85.41                \\
                                 & \multicolumn{1}{c|}{}                     & Median     & 89.54              & 89.16                & 89.74                & 89.04                & 89.37                & 88.65                & 89.70                & 87.14                & 89.83                & 89.22                & 89.70                & 89.18                & 89.51                \\
                                 & \multicolumn{1}{c|}{}                     & Byrd-SAGA       & 90.24              & 90.12                & 90.08                & 89.96                & 89.76                & 89.86                & 90.08                & 88.40                & 90.06                & 90.04                & 90.08                & 89.92                & 89.85                \\
                                 & \multicolumn{1}{c|}{}                     & RFA        & 92.01              & 91.88                & 91.92                & 91.65                & 91.66                & 91.72                & 91.91                & 91.23                & 92.04                & 91.89                & 91.95                & 91.62                & 91.49                \\
                                 & \multicolumn{1}{c|}{}                     & MCA        & 90.32              & 90.21                & 90.40                & 90.23                & 89.99                & 89.63                & 81.01                & \textbf{94.81}                & 10.28                & 90.29                & 90.32                & 90.21                & 89.91               
\end{tabular}
}
\end{table*}

We split the above three datasets into $M$ non-IID training sets, which is realized by letting the label of data samples to conform to Dirichlet distribution. The extent of non-IID can be adjusted by tuning the concentration parameter $\phi$ of Dirichlet distribution. 

\textbf{Models:} We adopt LeNet, Multilayer Perceptron (MLP), and VGG16 model. The introduction of these three models are as follows: 

\begin{itemize}
  \item \textbf{LeNet:} The LeNet model is one of the earliest published convolutional neural networks. For the experiments, like \cite{lecun1998gradient}, we are going to train a LeNet model with two convolutional layers (both kernel size are 5 and the out channel of first one is 6 and second one is 16), two pooling layers (both kernel size are 2 and stride are 2), and three fully connected layer (the first one is (16 × 4 × 4, 120), the second is (120, 60) and the last is (60, 10)) for MNIST dataset. And for CIFAR10 dataset, we are going to train a LeNet model with two convolutional layers (both kernel size are 5 and the out channel are 64), two pooling layers (both kernel size are 2 and stride are 2), and three fully connected layer (the first one is (64 × 5 × 5, 384), the second is (384, 192) and the last is (192, 10)). Cross-entropy function is taken as the training loss. 
  \item \textbf{MLP:} The MLP model is a machine learning model based on Feedforward Neural Network that can achieve high-level feature extraction and classification. 
  \textcolor{black}{We configure the MLP model to be with three connected layers (the first one is (28 × 28, 200), the second is (200, 100) and the last is (100, 10)) like \cite{yue2022neural}}. And for CIFAR10 dataset, we configure the MLP model to be with three connected layers (the first one is (3× 32 × 32, 200), the second is (200, 200) and the last is (200, 10)). And also cross-entropy function is taken as the training loss.
  \item {\textbf{VGG16:} The VGG16 model, originally developed by Simonyan and Zisserman from the Visual Geometry Group, represents a seminal 16-layer convolutional neural network architecture comprising 13 convolutional layers and 3 fully-connected (FC) layers. Each convolutional stage utilizes cascaded 3×3 kernels with stride 1 and ReLU activation, interspersed with 2×2 max-pooling operations that halve spatial resolution while preserving depth. The fully-connected hierarchy consists of two 4,096-unit hidden layers (FC1-2) followed by a 1,000-class output layer (FC3), totaling 138M trainable parameters. For CIFAR100 adaptation, we implemented fine-tuning to adapt to this dataset.}
\end{itemize}

\begin{figure*}[h]
    \centering
    \subfigure[LeNet and $\bar{C}_{\alpha} = 0.2$.]{
    \includegraphics[width = 0.2\textwidth]{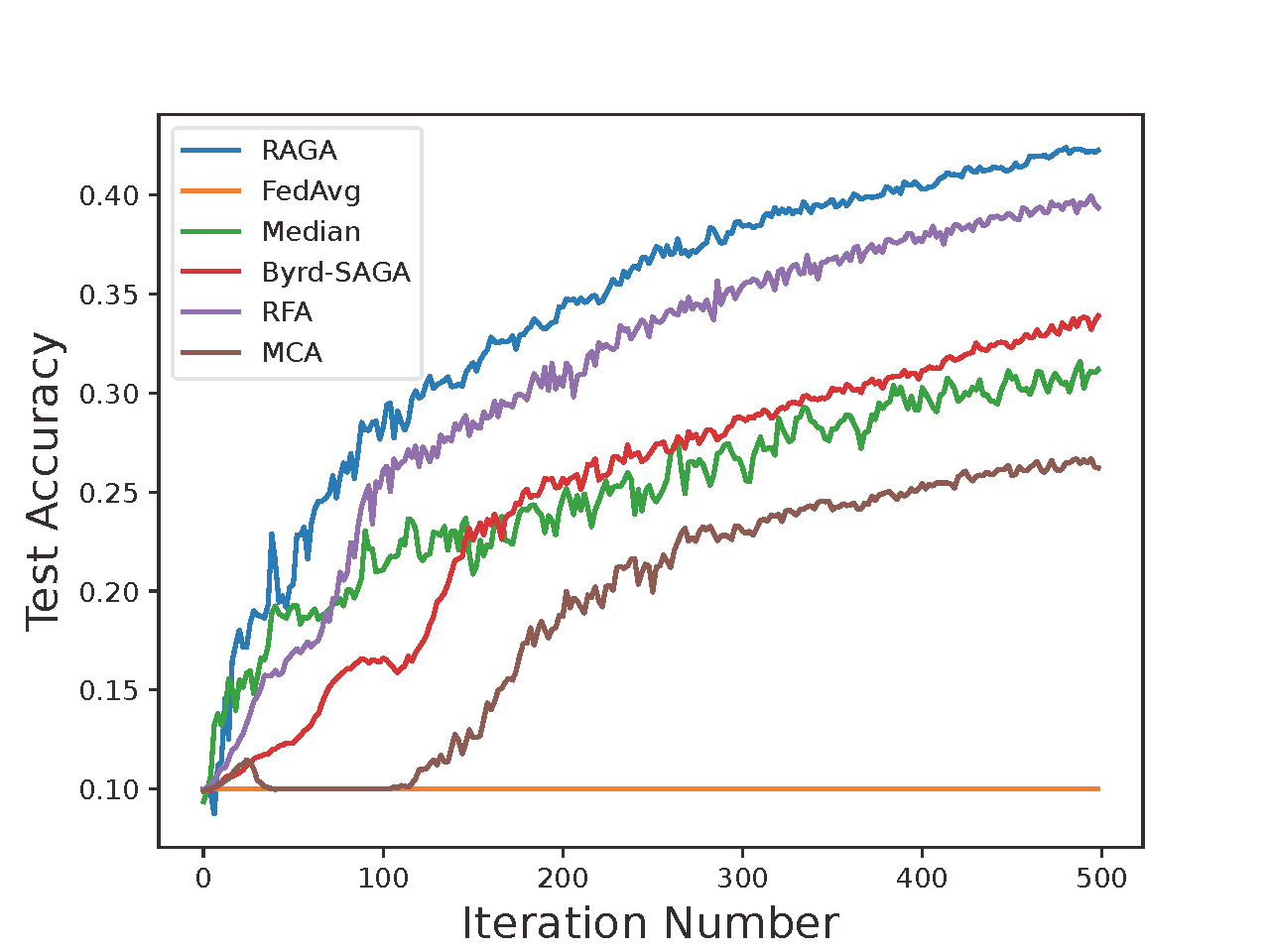}
    }
    \subfigure[LeNet and $\bar{C}_{\alpha} = 0.4$.]{
    \includegraphics[width = 0.2\textwidth]{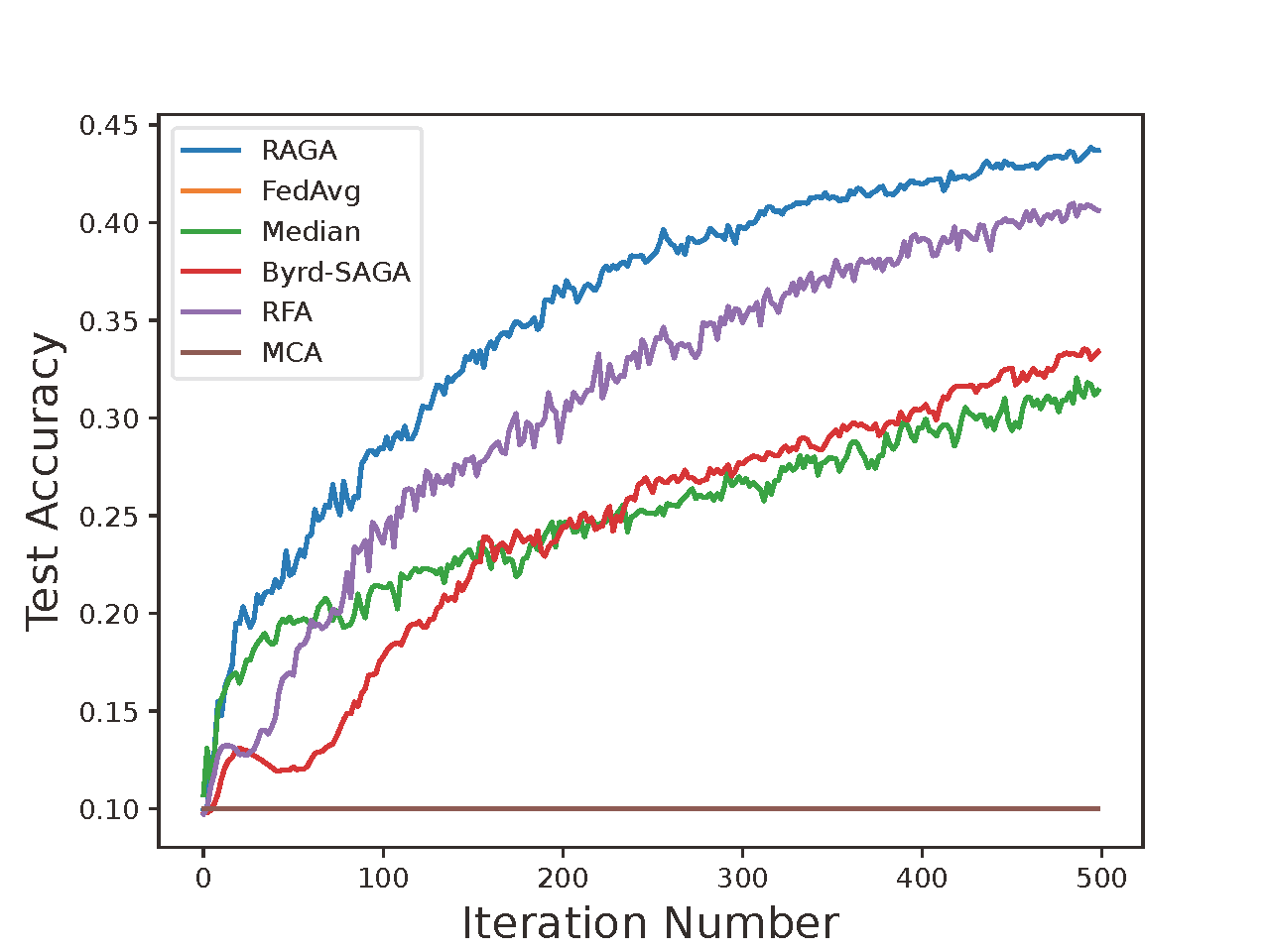}
    }
    \subfigure[MLP and $\bar{C}_{\alpha} = 0.2$.]{
    \includegraphics[width = 0.2\textwidth]{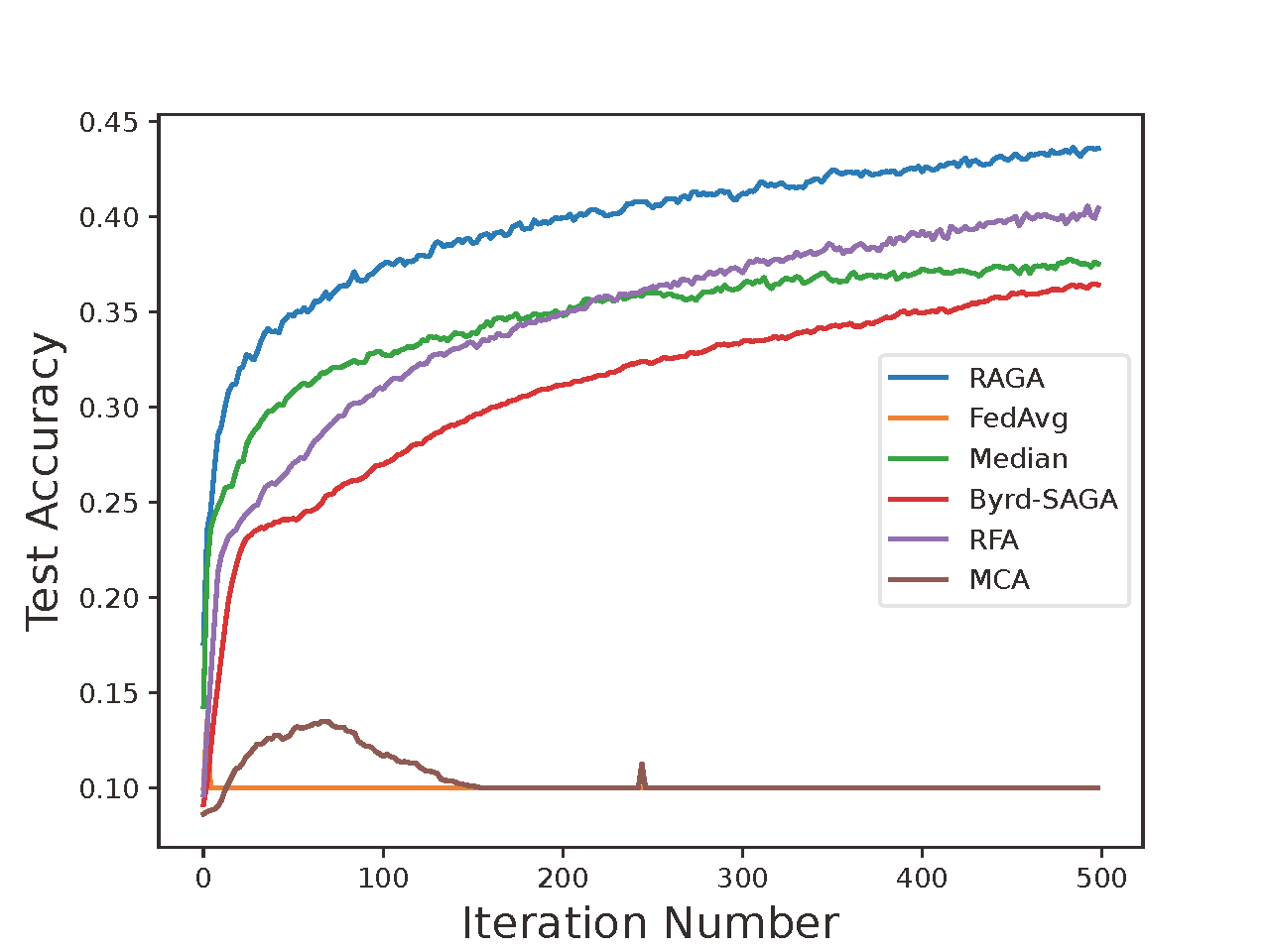}
    }
    \subfigure[MLP and $\bar{C}_{\alpha} = 0.4$.]{
    \includegraphics[width = 0.2\textwidth]{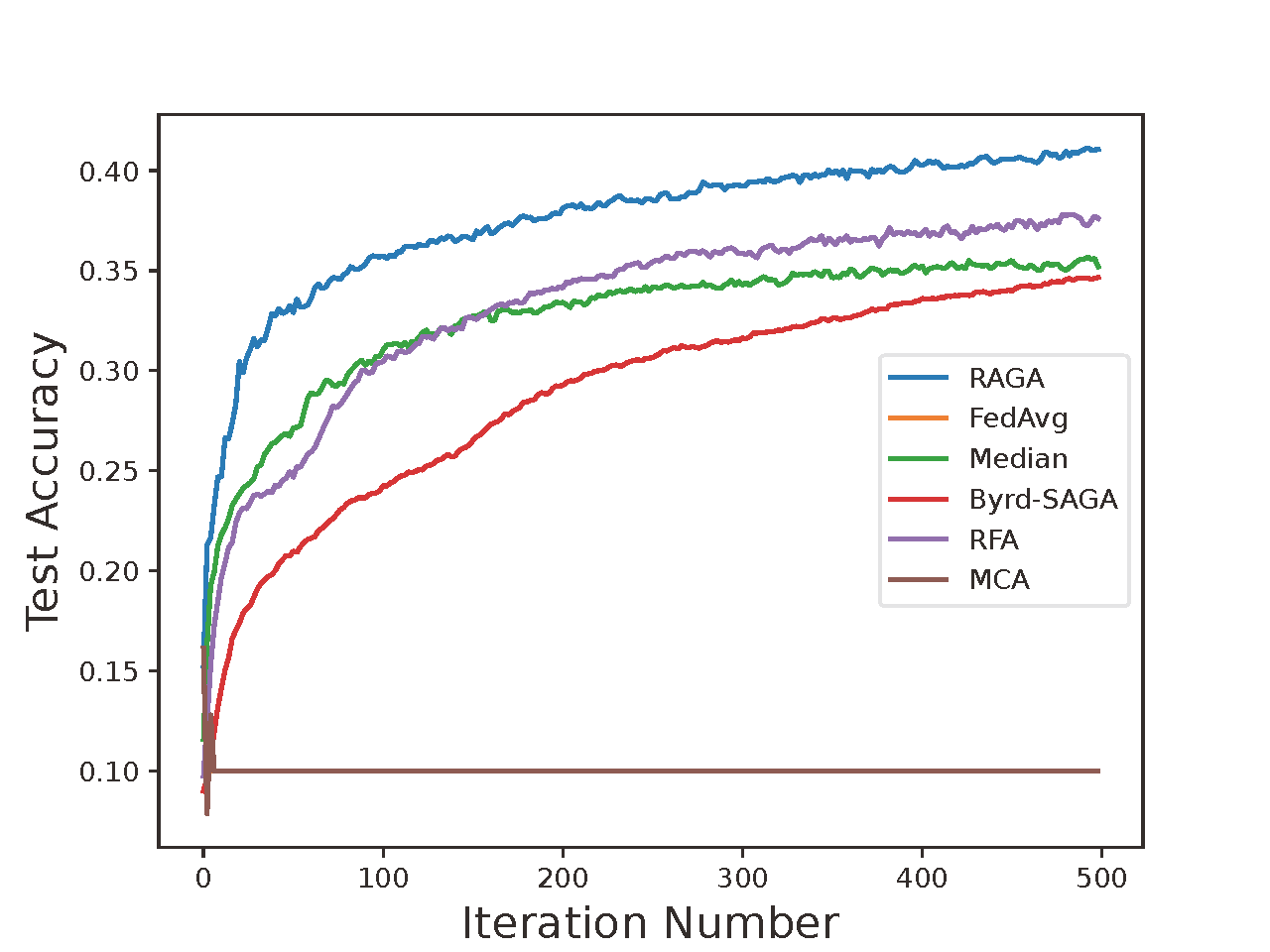}
    }
    \caption{Test accuracy of RAGA and baselines on $\phi = 0.6$, Sign-flip attack, and CIFAR10 dataset.}\label{fig:cta6}
\end{figure*}

\textbf{Byzantine attacks:} The intensity level of Byzantine attack $\bar{C}_{\alpha} = 1 - C_{\alpha}$ is set to 0, 0.1, 0.2, 0.3, 0.4 and we select three types of Byzantine attack, which are introduced as follows:
\begin{itemize}
    \item \textbf{Gaussian attack:} All Byzantine attacks are selected as the Gaussian attack, which obeys $\mathcal{N} (0, 90)$.
    \item \textbf{Sign-flip attack:} All Byzantine clients upload $-3 * \sum_{m \in \mathcal{M} \setminus \mathcal{B}} \bm{z}_m^t$ to the central server on iteration number $t$.
    \item \textbf{LIE attack \cite{baruch2019little}}: LIE attack adds small amounts of noise to each dimension of the benign gradients. The noise is controlled by a coefficient $c$, which enables the attack to evade detection by robust aggregation methods while negatively impacting the global model. Specifically, the attacker calculates the mean $\rho$ and standard deviation $\nu$ of the parameters submitted by honest users, calculates the coefficient $c$ based on the total number of honest and malicious clients, and finally computes the malicious update as $\rho$ + $c \nu$. We set $c$ to 0.7.
\end{itemize}

\textbf{Hyperparameters:}
We assume $M=50$ and the batchsize of all the experiments is 32. The error tolerance $\epsilon$ for numerically working out  geometric median and MCA are $\epsilon = 1 \times 10^{-5}$. The concentration parameter $\phi$ is set to 0.6, 0,4 and 0.2. The iteration number $T=500$ for both MNIST and CIFAR10 dataset, while $T=100$ for CIFAR100 dataset. 
In default, the round number of local updates $K^t = 3$ for all algorithms on LeNet and MLP model, while $K^t = 1$ on VGG16 model. 
In terms of the learning rate, we adopt default setup as shown in corresponding papers for baseline methods. {For RAGA, the learning rate $\eta^{t,k}_m = \eta^t = \frac{K^t}{\sqrt{5}\sqrt{t+5}} = \frac{K^t}{5\sqrt{0.2t+1}}$ for LeNet and MLP models, while $\eta^{t,k}_m = \eta^t = \frac{4*10^{-2}K^t}{\sqrt{t+100}} = \frac{4*10^{-3}K^t}{\sqrt{0.01t+1}}$ for VGG16 model. This learning rate configuration is motivated by Remark 1, which is derived from Theorem 1 and tailored for non-convex loss functions, covering the experimental models including LeNet, MLP, and VGG16. It is worth noting that the learning rate used in practice slightly deviates from that suggested in Remark 1 due to implementation considerations. Specifically, we set $\delta = 1/6$ and $c = 5$ for LeNet and MLP, and $\delta = 1/6$ and $c = 100$ for VGG16. Since the exact smoothness constant $L$ is unknown, we adopt heuristic coefficients, e.g., $\frac{K^t}{\sqrt{5}}$ for LeNet/MLP and $4*10^{-2}K^t$ for VGG16, to ensure the learning rate remains sufficiently small for convergence.}

\textbf{Baselines:} The convergence performance of four algorithms: RAGA, FedAvg \cite{zhao2018federated}, Median \cite{xie2018generalized}, Byrd-SAGA \cite{wu2020federated}, RFA \cite{pillutla2022robust}, MCA \cite{luan2024robust} are compared. 
For the latter five algorithms, FedAvg is renowned in traditional FL but does not concern Byzantine attack and can be taken as a performance metric when there is no Byzantine attack, Median takes the coordinate-wise median of the uploaded vectors to update the global model parameter, Byrd-SAGA only executes one local updating in each round of iteration, RFA selects the trimmed mean of the model parameters over multiple local updating as the uploaded vector, and MCA utilizes its the maximum correntropy of the model parameters' distribution characteristics and compute the central value to update the global model parameter.
It is worthy to note that there is another Byzantine-resilient FL algorithm in related works: BROADCAST \cite{zhu2023byzantine}. We do not show the performance of it because BROADCAST is very similar with Byrd-SAGA and only differs from Byrd-SAGA by uploading vector from each user to the central server in a compressive, rather than lossless way like in Byrd-SAGA. 


\subsection{Convergence Performance}

\begin{figure}[h]
    \centering
    \subfigure[LeNet and Gaussian attack.]{
    \includegraphics[width = 0.2\textwidth]{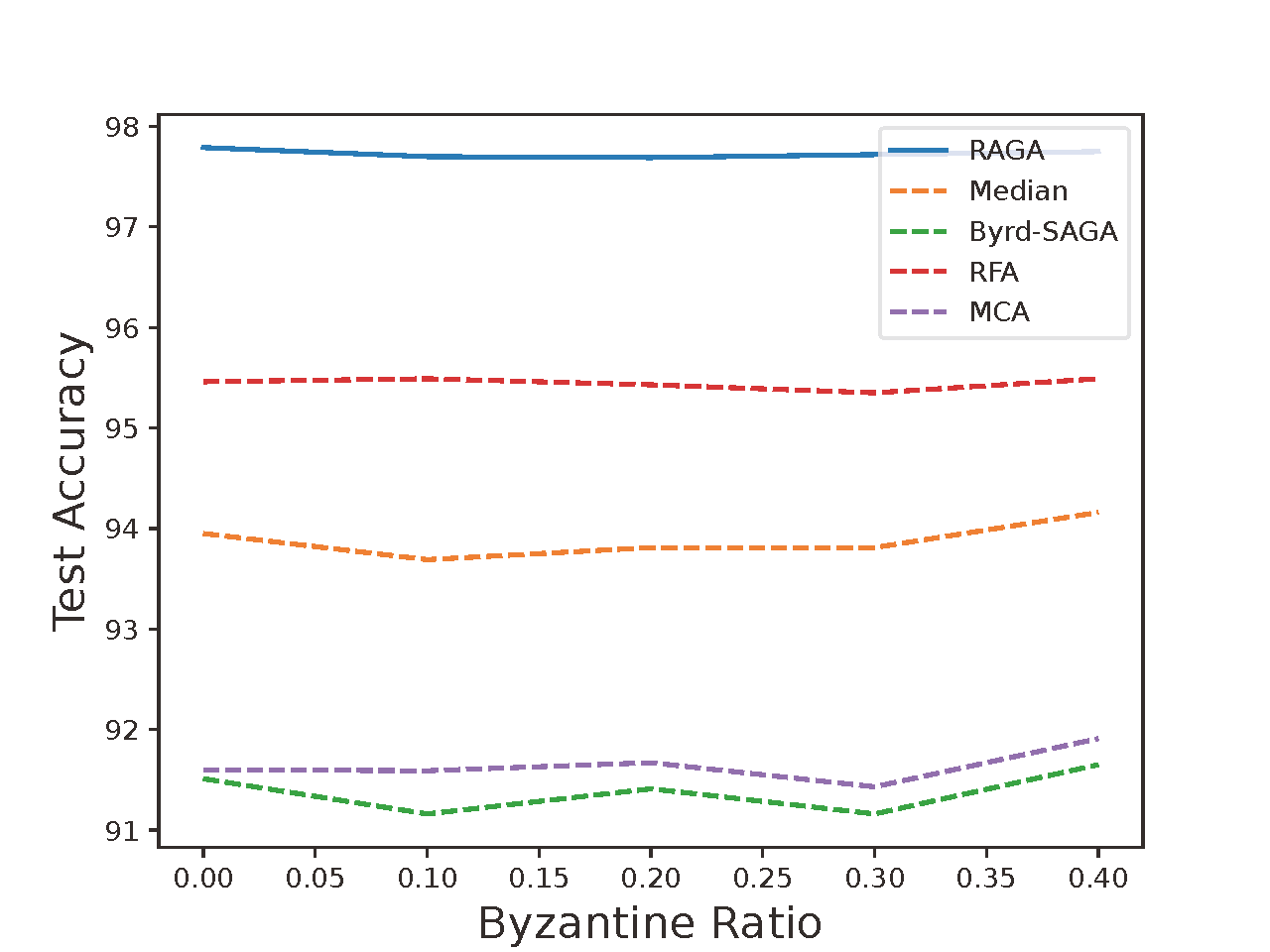}
    }
    \subfigure[LeNet and LIE attack.]{
    \includegraphics[width = 0.2\textwidth]{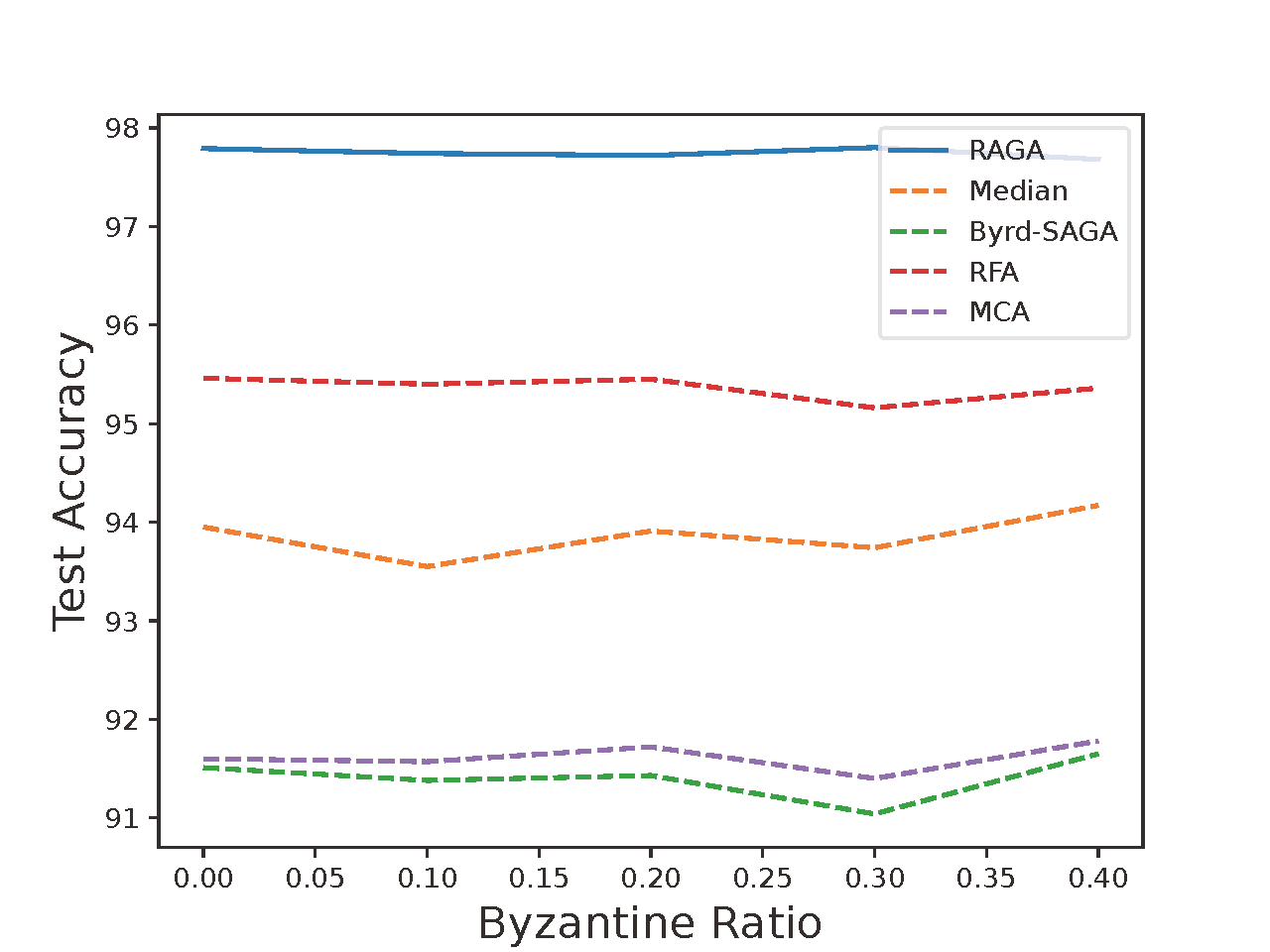}
    }
    \subfigure[MLP and Gaussian attack.]{
    \includegraphics[width = 0.2\textwidth]{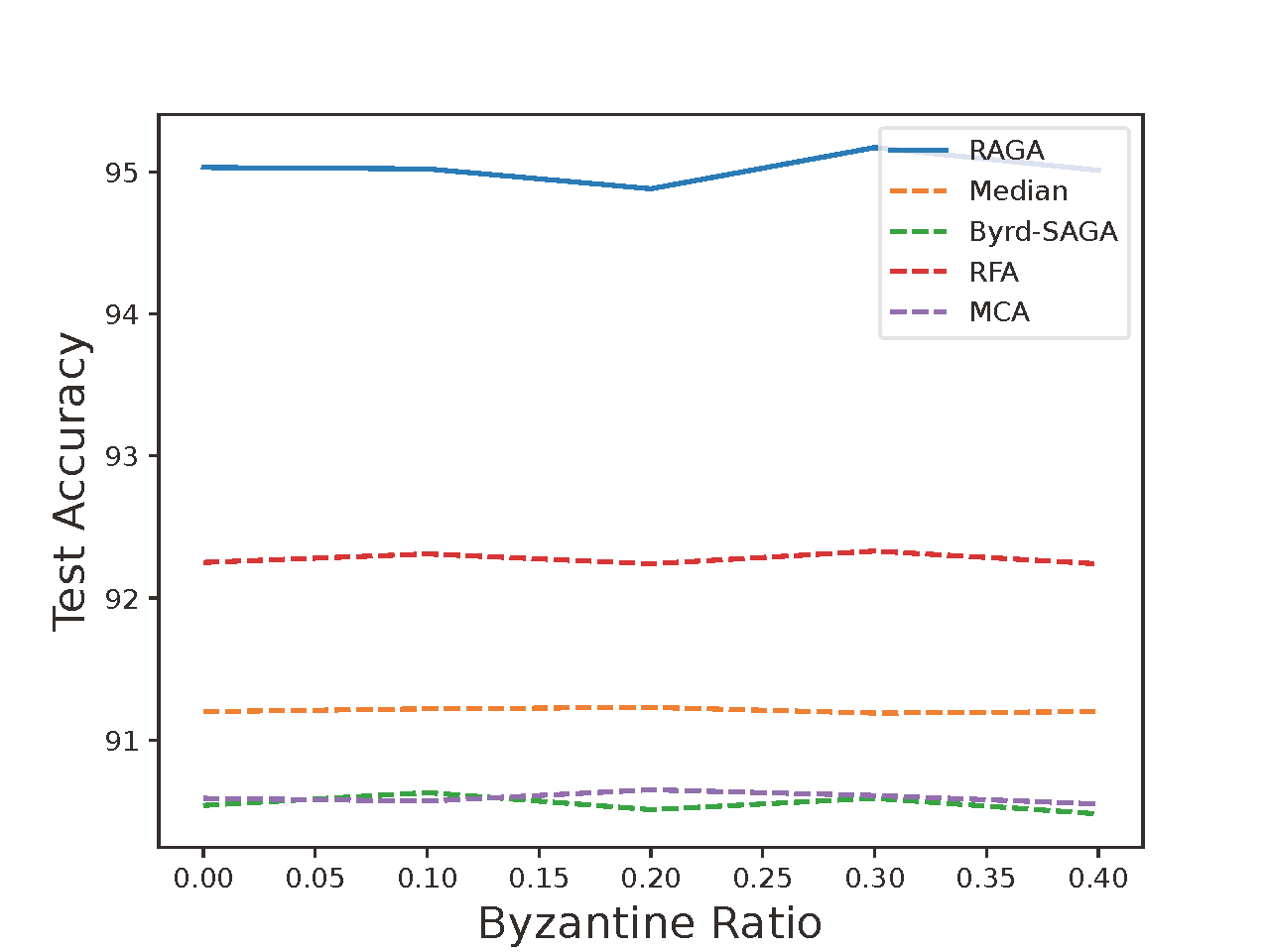}
    }
    \subfigure[MLP and LIE attack.]{
    \includegraphics[width = 0.2\textwidth]{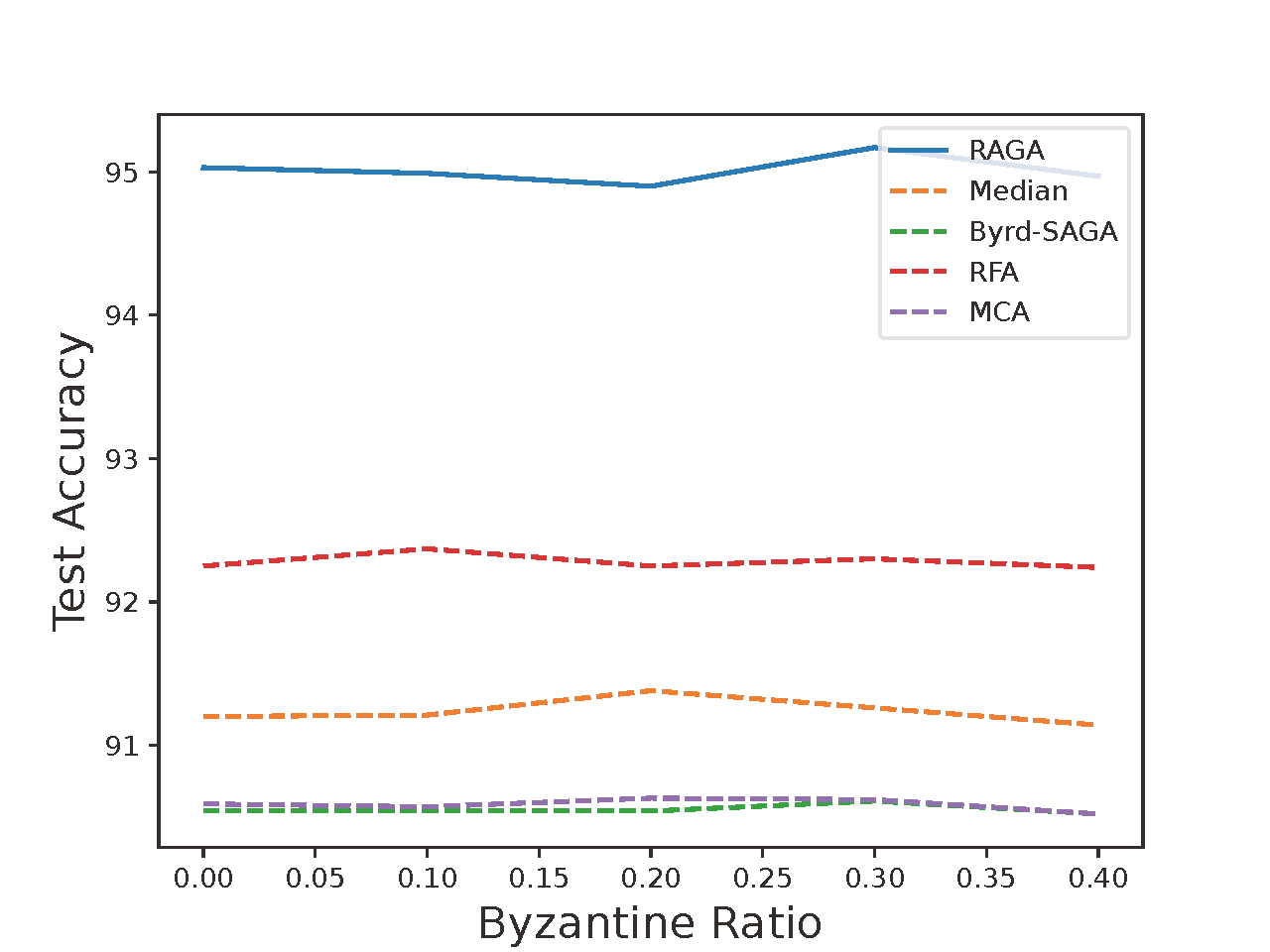}
    }
    \vfill
    \caption{The maximum test accuracy(\%) of 500 iteration number of RAGA and baselines on $\phi = 0.4$ and MNIST dataset. }\label{fig:m4}
\end{figure}

\begin{table*}[t]
\caption{\centering The maximum test accuracy(\%) of 500 iteration number of RAGA, FedAvg, Median, Byrd-SAGA, RFA and MCA with different types, and ratios of Byzantine attack on CIFAR10 dataset and two learning models. } 
\label{tab:accc}
\resizebox{\textwidth}{!}{
\renewcommand{\arraystretch}{1.5}
\begin{tabular}{c|cc|c|cccc|cccc|cccc}
\multirow{3}{*}{Model} & \multicolumn{2}{c|}{Attack Name}                       & \textbf{No Attack} & \multicolumn{4}{c|}{\textbf{Gaussian Attack}}                                             & \multicolumn{4}{c|}{\textbf{Sign-flip Attack}}                                            & \multicolumn{4}{c}{\textbf{LIE Attack}}                                                   \\ \cline{2-16} 
                       & \multicolumn{2}{c|}{$\bar{C}_{\alpha}$}                & \multirow{2}{*}{0} & \multirow{2}{*}{0.1} & \multirow{2}{*}{0.2} & \multirow{2}{*}{0.3} & \multirow{2}{*}{0.4} & \multirow{2}{*}{0.1} & \multirow{2}{*}{0.2} & \multirow{2}{*}{0.3} & \multirow{2}{*}{0.4} & \multirow{2}{*}{0.1} & \multirow{2}{*}{0.2} & \multirow{2}{*}{0.3} & \multirow{2}{*}{0.4} \\ \cline{2-3}
                       & \multicolumn{1}{c|}{$\phi$}               & Algorithms &                    &                      &                      &                      &                      &                      &                      &                      &                      &                      &                      &                      &                      \\ \hline
\multirow{6}{*}{\textbf{LeNet}} & \multicolumn{1}{c|}{\multirow{6}{*}{0.4}} & RAGA       & \textbf{42.92}     & \textbf{43.28}       & \textbf{41.68}       & \textbf{42.81}       & \textbf{41.89}       & \textbf{41.58}       & 41.40                & \textbf{41.03}       & \textbf{42.29}       & \textbf{42.91}       & 40.20                & \textbf{42.12}       & \textbf{41.42}       \\
                       & \multicolumn{1}{c|}{}                     & FedAvg     & 39.45              & 10.91                & 10.00                & 10.00                & 10.00                & 10.00                & 10.00                & 10.00                & 10.00                & 41.37                & 35.38                & 25.40                & 15.89                \\
                       & \multicolumn{1}{c|}{}                     & Median     & 32.17              & 31.70                & 30.83                & 31.59                & 30.29                & 30.09                & 30.54                & 28.00                & 30.19                & 31.50                & 30.83                & 32.25                & 31.25                \\
                       & \multicolumn{1}{c|}{}                     & Byrd-SAGA       & 34.81              & 34.60                & 33.86                & 33.92                & 34.27                & 33.49                & 33.66                & 29.37                & 34.21                & 35.20                & 34.19                & 33.89                & 34.64                \\
                       & \multicolumn{1}{c|}{}                     & RFA        & 41.85              & 41.64                & 41.65                & 41.42                & 41.27                & 40.92                & \textbf{41.61}       & 37.20                & 41.65                & 41.78                & \textbf{41.32}       & 41.21                & 40.88                \\
                       & \multicolumn{1}{c|}{}                     & MCA        & 35.10              & 34.83                & 34.26                & 34.23                & 34.69                & 32.92                & 18.01                & 10.00                & 10.00                & 34.42                & 34.21                & 34.64                & 34.53                \\ \hline
\multirow{6}{*}{\textbf{MLP}}   & \multicolumn{1}{c|}{\multirow{6}{*}{0.4}} & RAGA       & \textbf{44.21}     & \textbf{44.32}       & \textbf{44.06}       & \textbf{44.29}       & \textbf{43.13}       & \textbf{43.63}       & \textbf{43.71}       & \textbf{44.02}       & \textbf{42.88}       & \textbf{44.11}       & \textbf{43.93}       & \textbf{44.65}       & \textbf{43.20}       \\
                       & \multicolumn{1}{c|}{}                     & FedAvg     & 40.13              & 10.86                & 10.00                & 12.03                & 13.84                & 12.32                & 14.05                & 14.46                & 10.00                & 34.38                & 30.80                & 25.30                & 23.83                \\
                       & \multicolumn{1}{c|}{}                     & Median     & 37.66              & 37.34                & 37.48                & 37.62                & 36.51                & 36.77                & 37.06                & 37.12                & 36.23                & 37.56                & 37.42                & 37.76                & 36.54                \\
                       & \multicolumn{1}{c|}{}                     & Byrd-SAGA       & 37.38              & 36.72                & 36.93                & 36.85                & 36.01                & 36.34                & 37.03                & 36.52                & 35.99                & 36.78                & 37.03                & 36.89                & 36.18                \\
                       & \multicolumn{1}{c|}{}                     & RFA        & 40.53              & 40.41                & 40.26                & 40.94                & 39.85                & 37.91                & 40.28                & 39.61                & 40.08                & 40.69                & 40.51                & 40.72                & 39.97                \\
                       & \multicolumn{1}{c|}{}                     & MCA        & 37.18              & 36.87                & 36.94                & 36.99                & 36.36                & 36.47                & 14.68                & 15.40                & 10.00                & 36.63                & 37.18                & 37.00                & 35.97               
\end{tabular}
}
\end{table*}

\begin{figure}[h]
    \centering
    \subfigure[LeNet and Gaussian attack.]{
    \includegraphics[width = 0.2\textwidth]{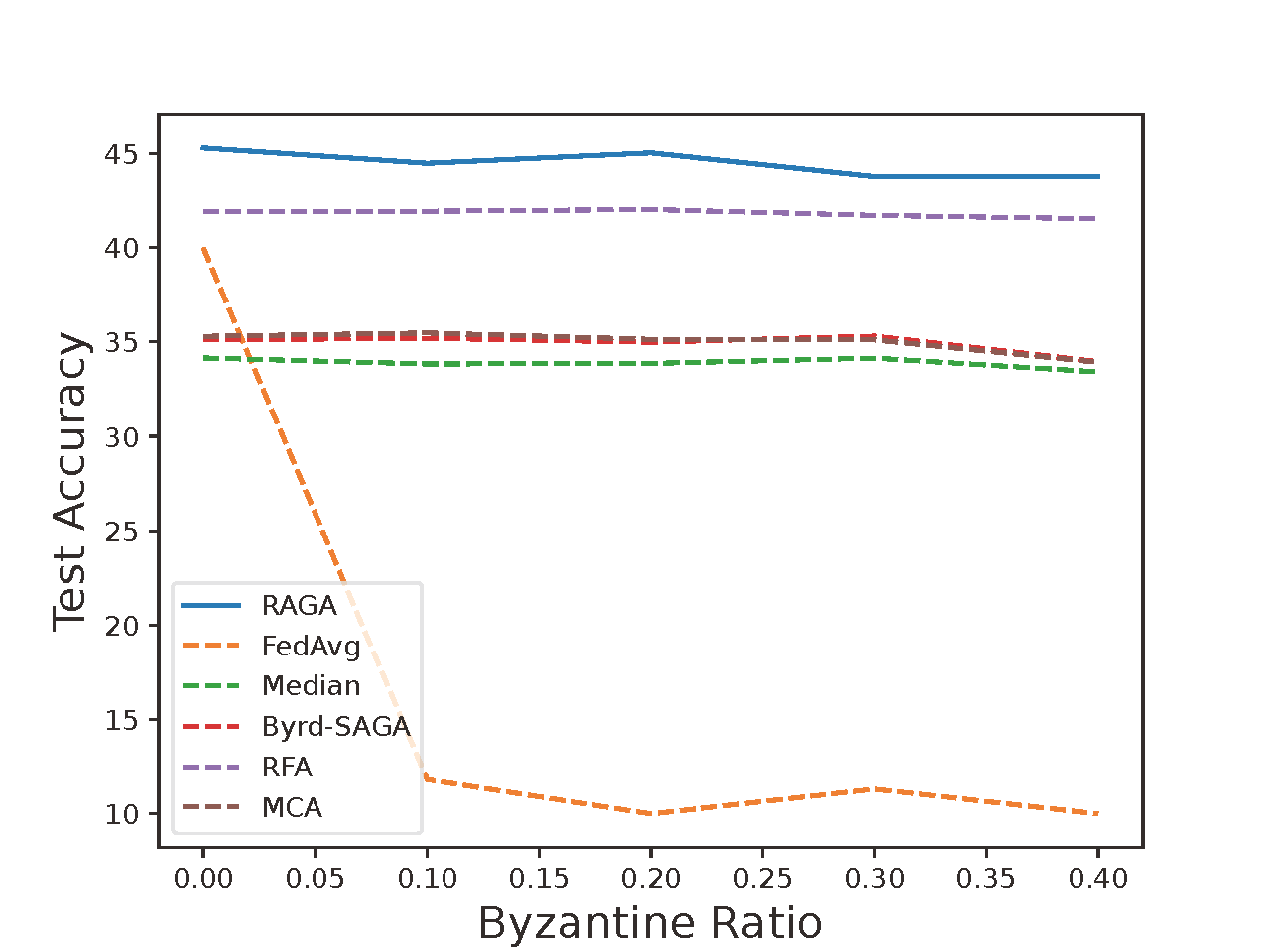}
    }
    \subfigure[LeNet and LIE attack.]{
    \includegraphics[width = 0.2\textwidth]{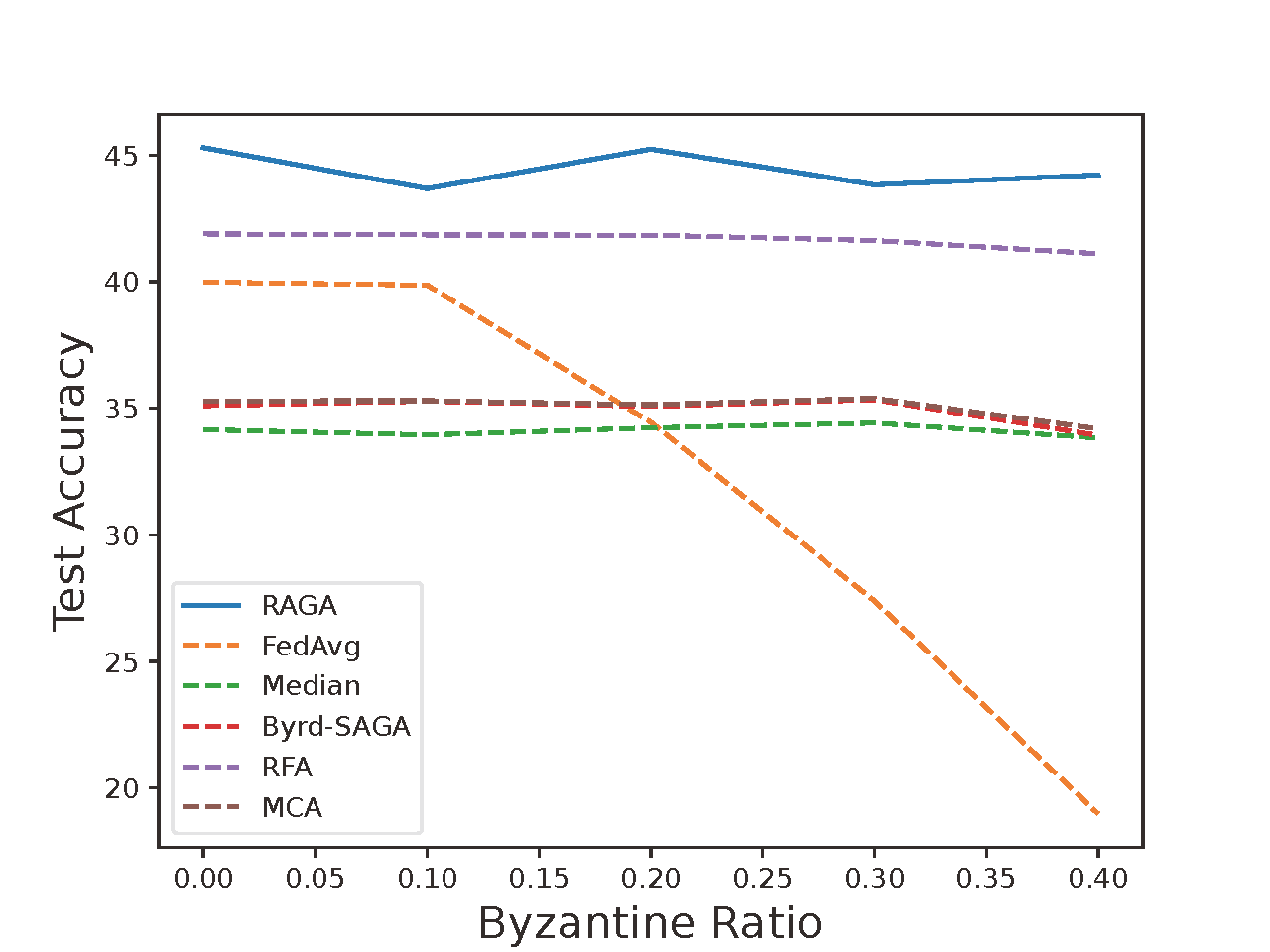}
    }
    \subfigure[MLP and Gaussian attack.]{
    \includegraphics[width = 0.2\textwidth]{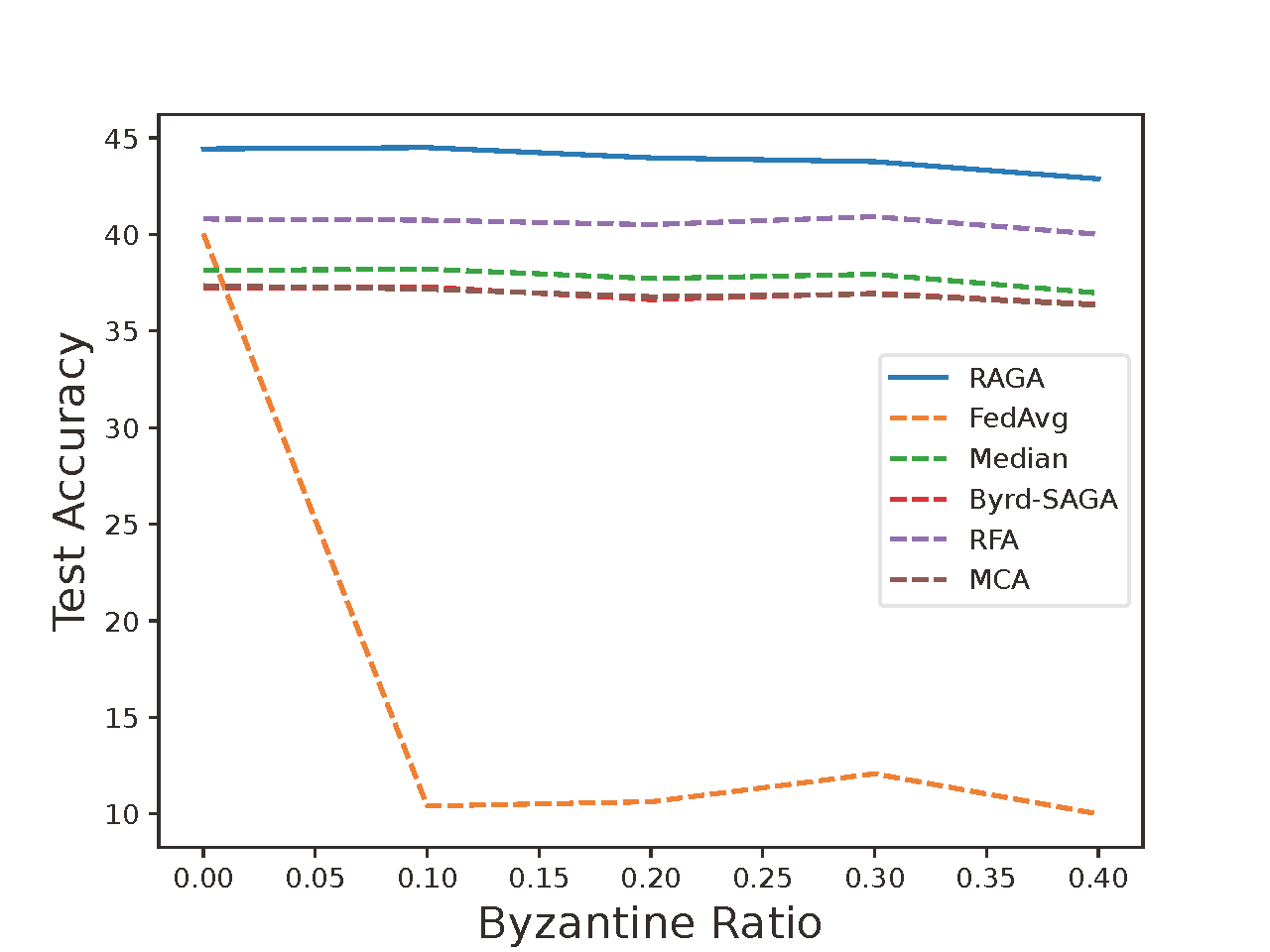}
    }
    \subfigure[MLP and LIE attack.]{
    \includegraphics[width = 0.2\textwidth]{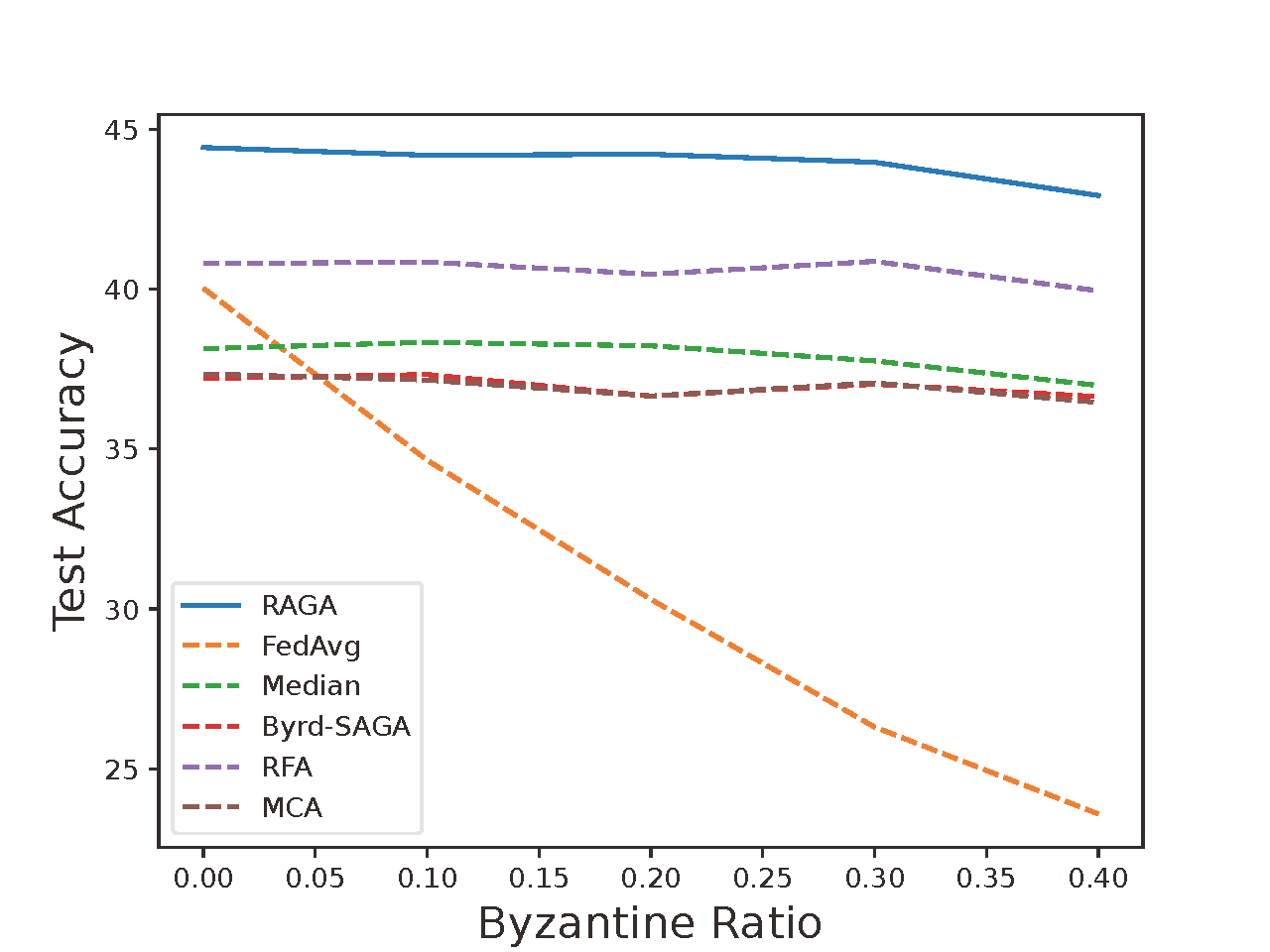}
    }
    \vfill
    \caption{The maximum test accuracy(\%) of 500 iteration number of RAGA and baselines on $\phi = 0.6$ and CIFAR10 dataset. }\label{fig:c6}
\end{figure}

\begin{figure*}[h]
    \centering
    \subfigure[Gaussian attack.]{
    \includegraphics[width = 0.2\textwidth]{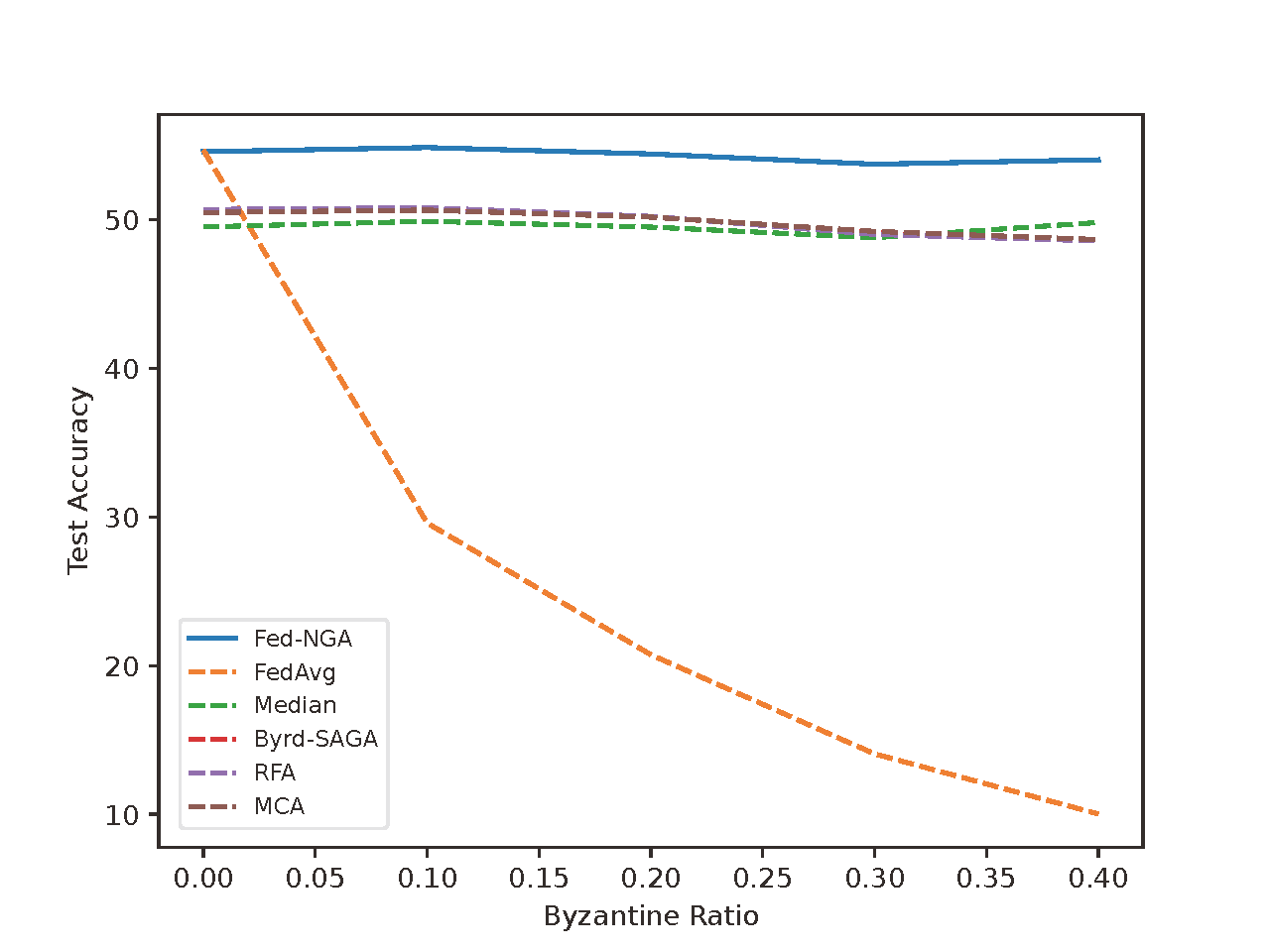}
    }
    \subfigure[Sign-flip attack.]{
    \includegraphics[width = 0.2\textwidth]{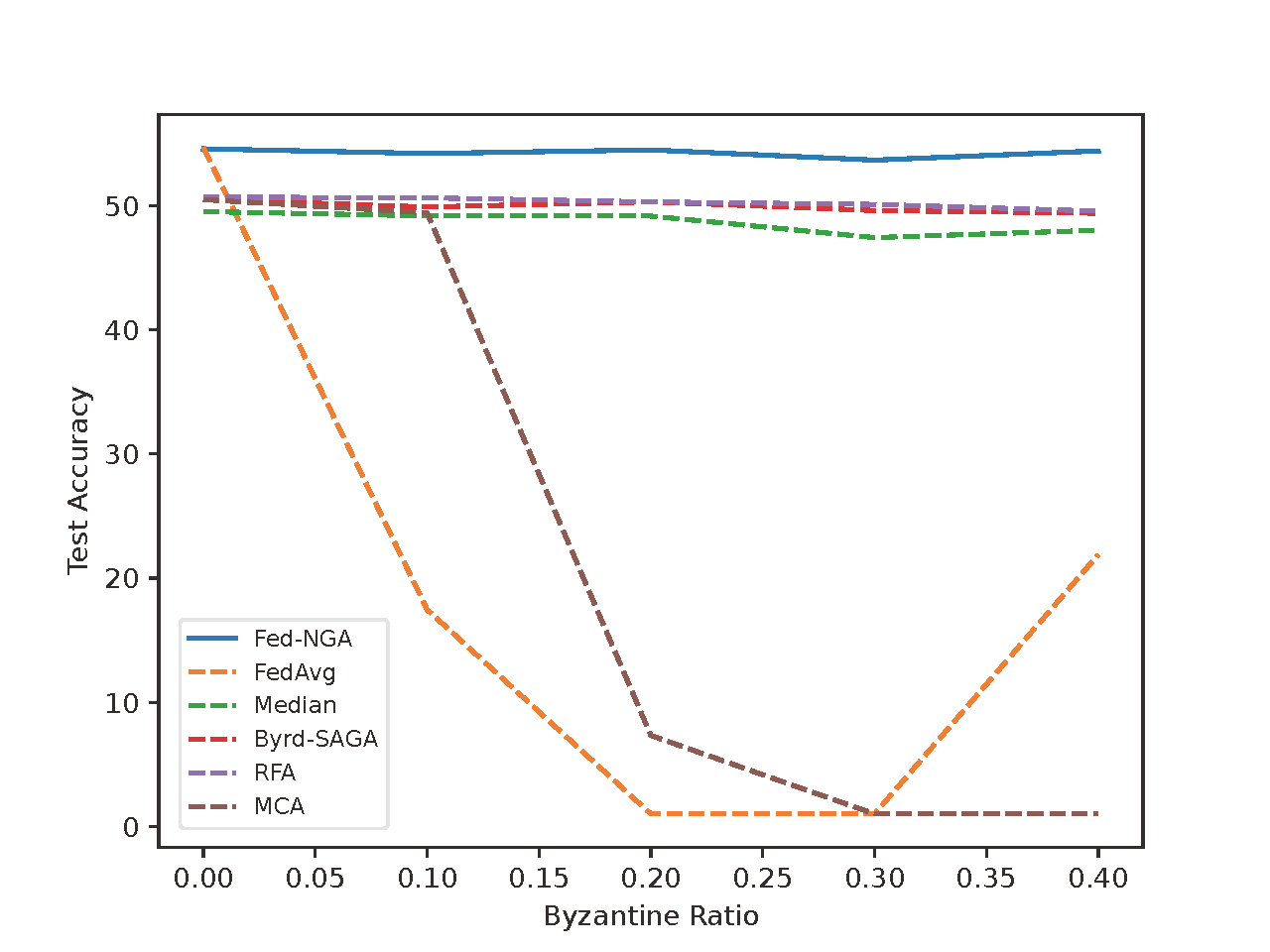}
    }
    \subfigure[LIE attack.]{
    \includegraphics[width = 0.2\textwidth]{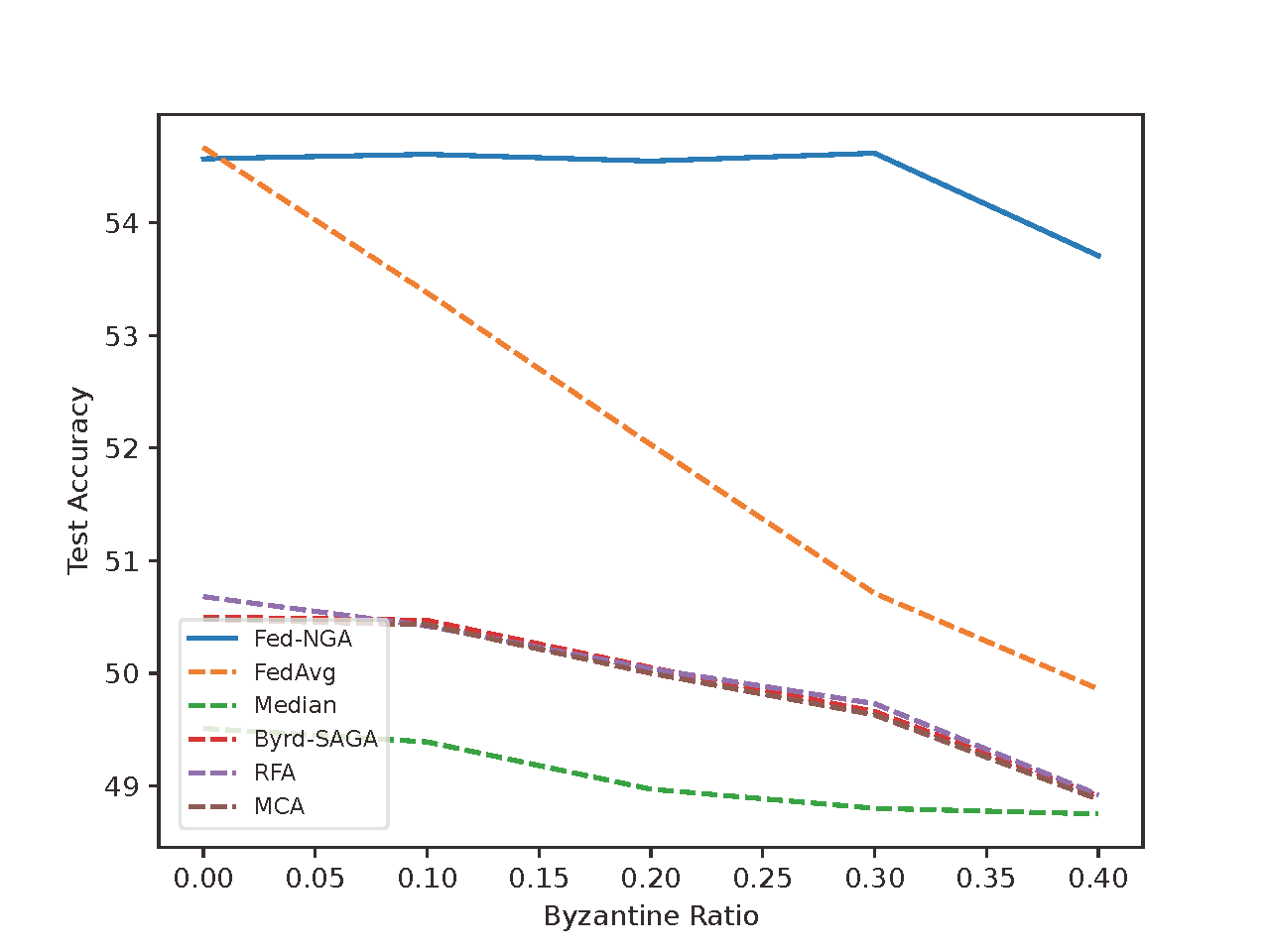}
    }
    \caption{The maximum test accuracy (\%) of 100 iteration number of RAGA and baselines on $\phi = 0.6$ and CIFAR100 dataset.}\label{fig:c1006}
\end{figure*}

\begin{figure}[h]
    \centering
    \subfigure[MNIST and Gaussian attack.]{
    \includegraphics[width = 0.2\textwidth]{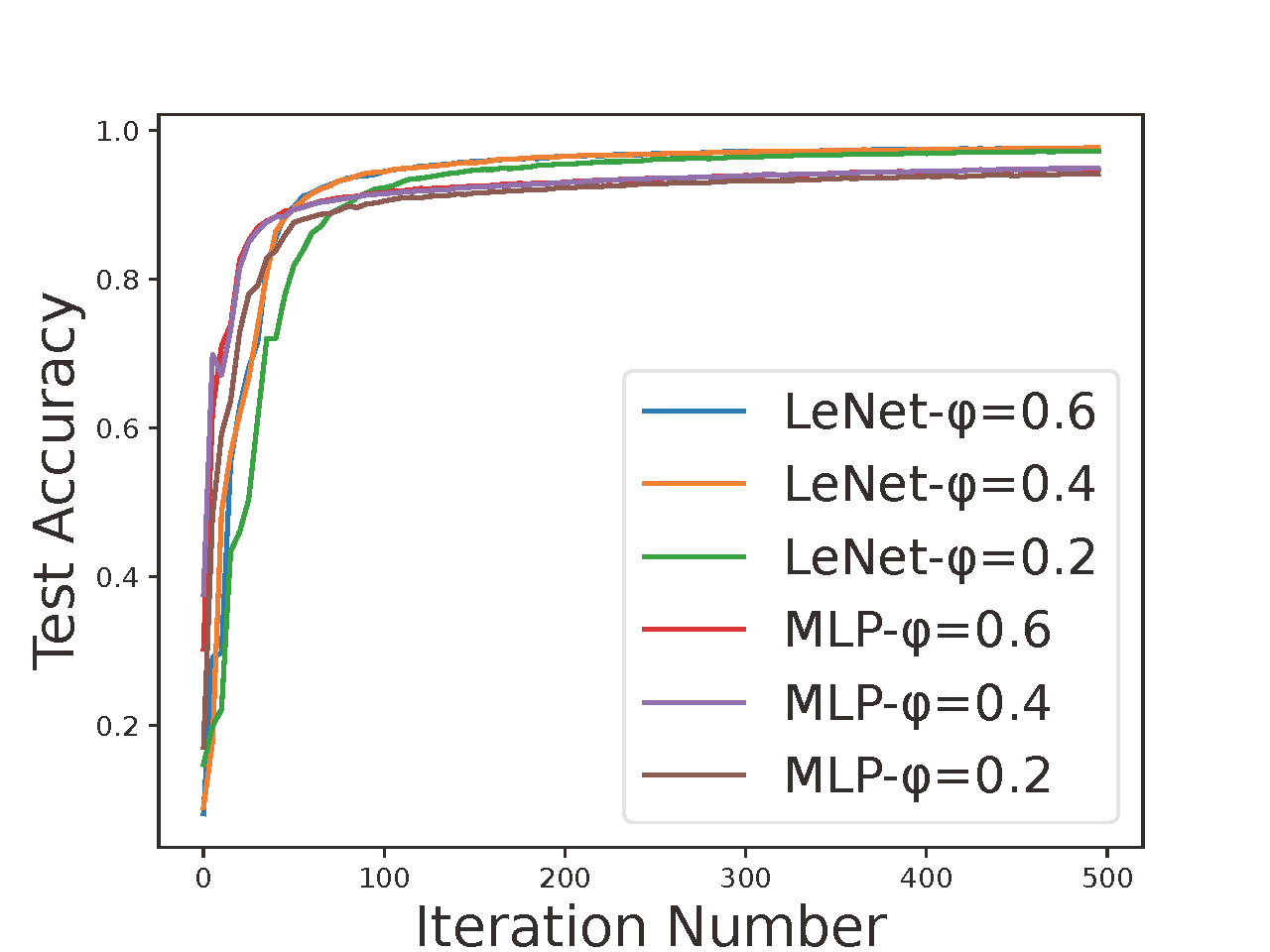}
    }
    \subfigure[MNIST and Sign-flip attack.]{
    \includegraphics[width = 0.2\textwidth]{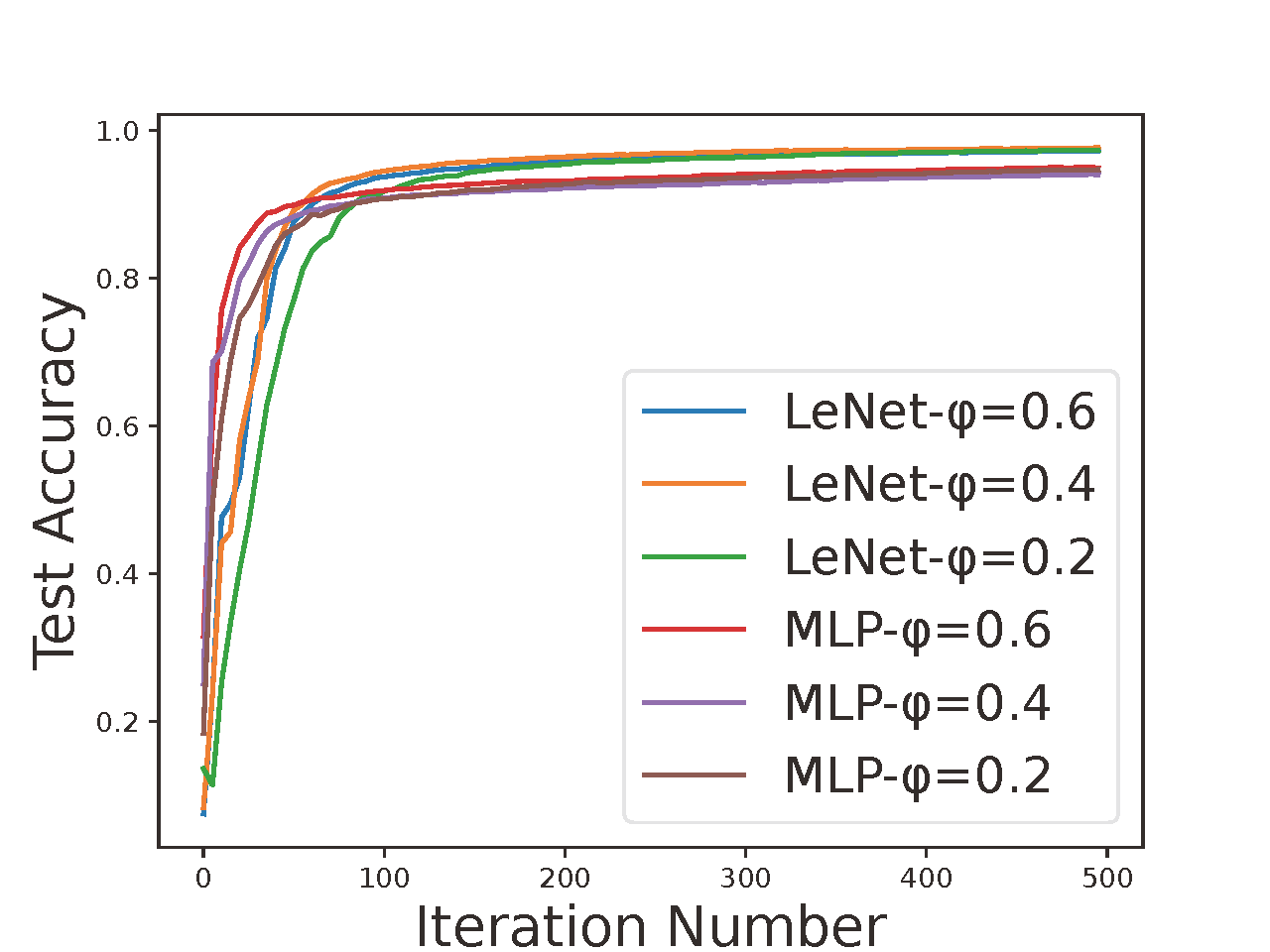}
    }
    \subfigure[MNIST and LIE attack.]{
    \includegraphics[width = 0.2\textwidth]{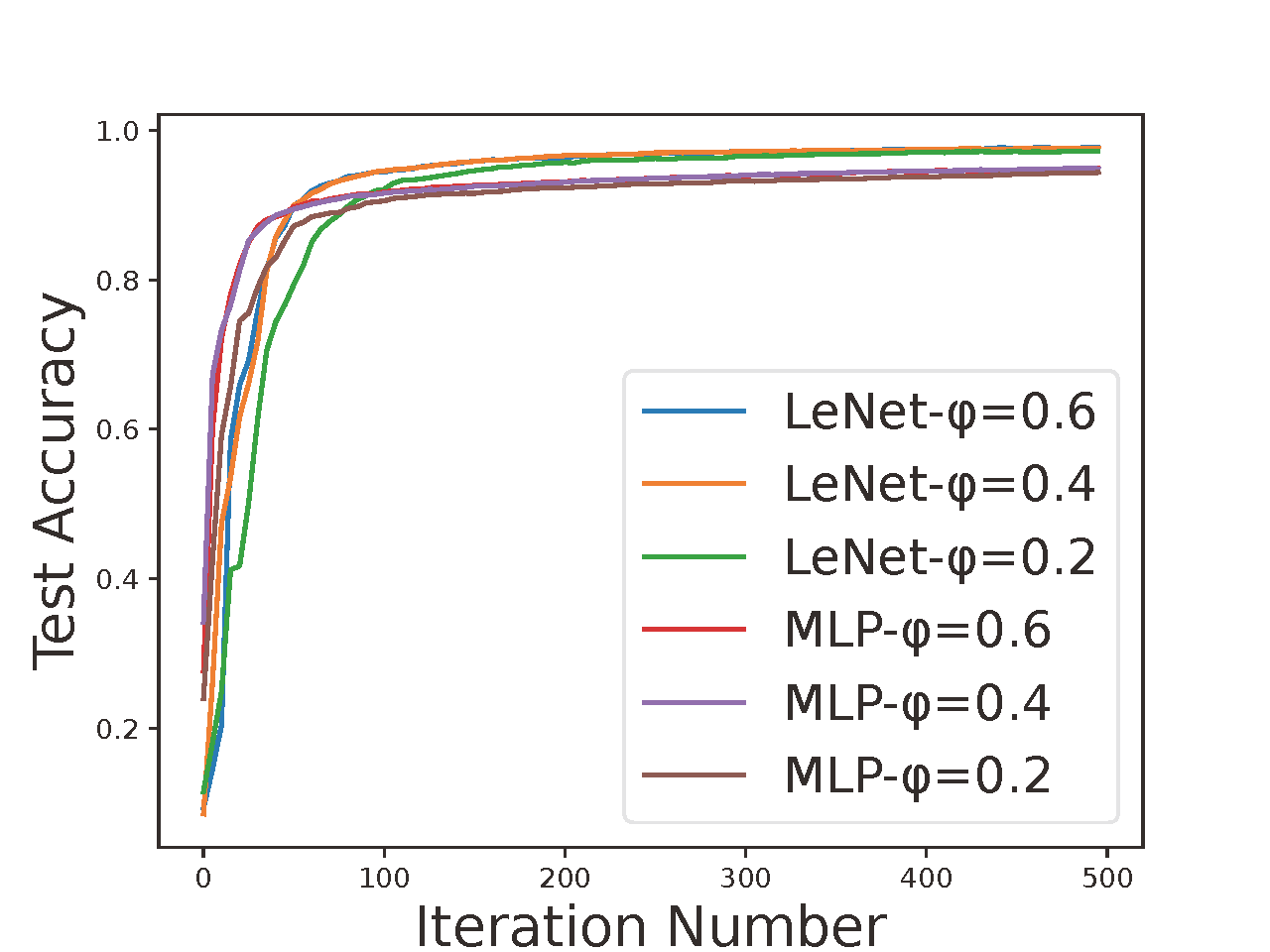}
    }
    \subfigure[CIFAR10 and Gaussian attack.]{
    \includegraphics[width = 0.2\textwidth]{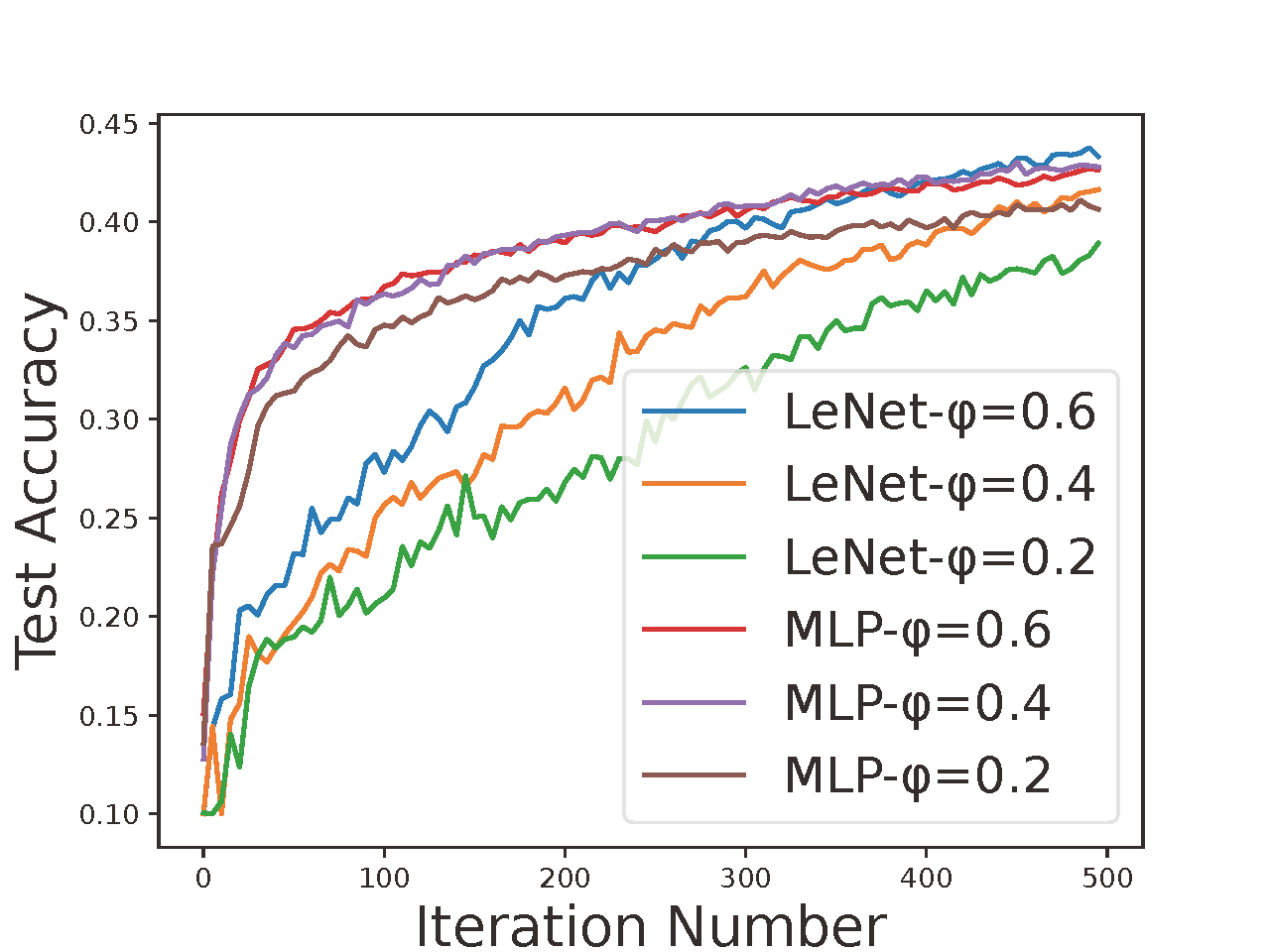}
    }
    \subfigure[CIFAR10 and Sign-flip attack.]{
    \includegraphics[width = 0.2\textwidth]{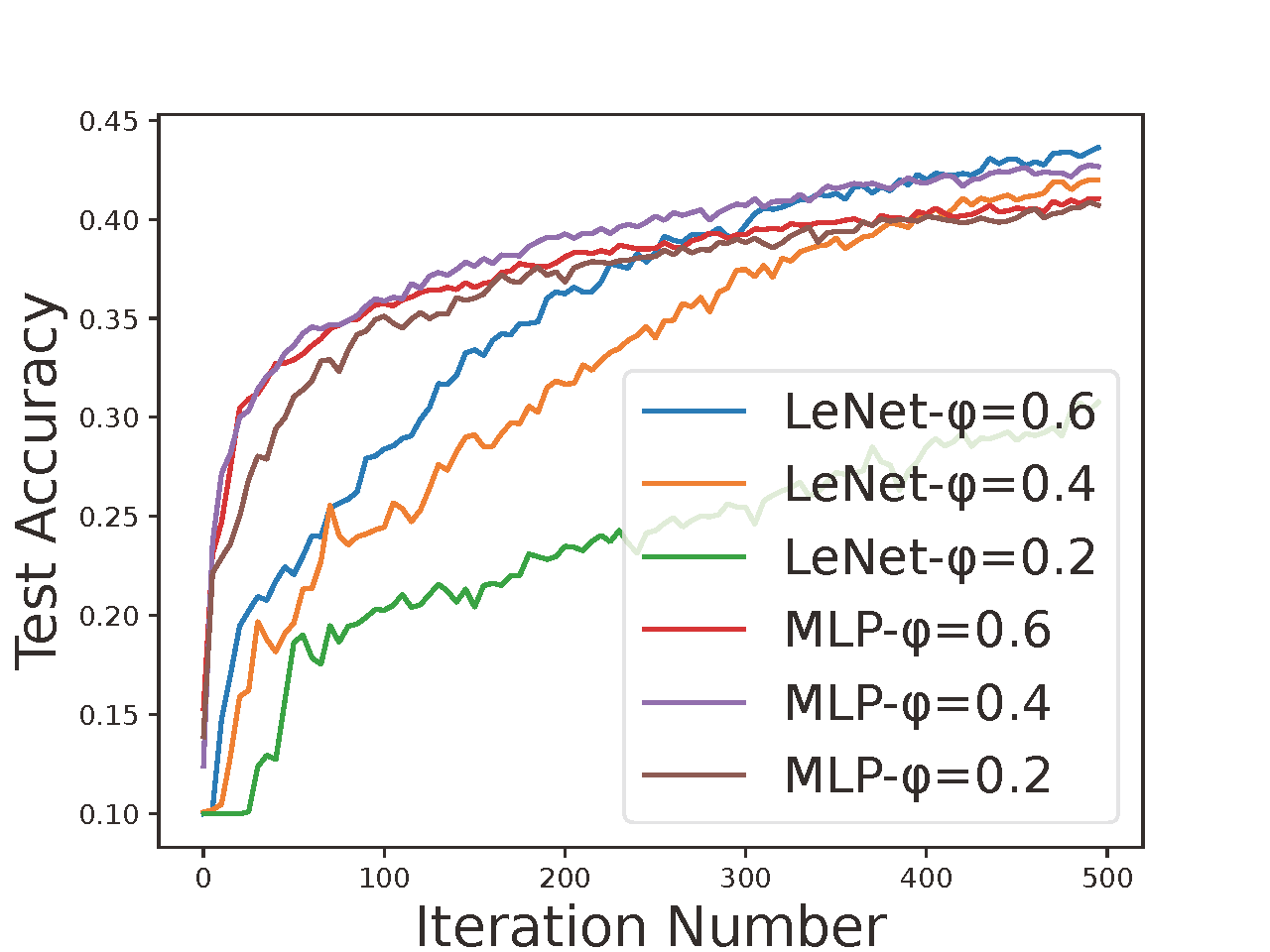}
    }
    \subfigure[CIFAR10 and LIE attack.]{
    \includegraphics[width = 0.2\textwidth]{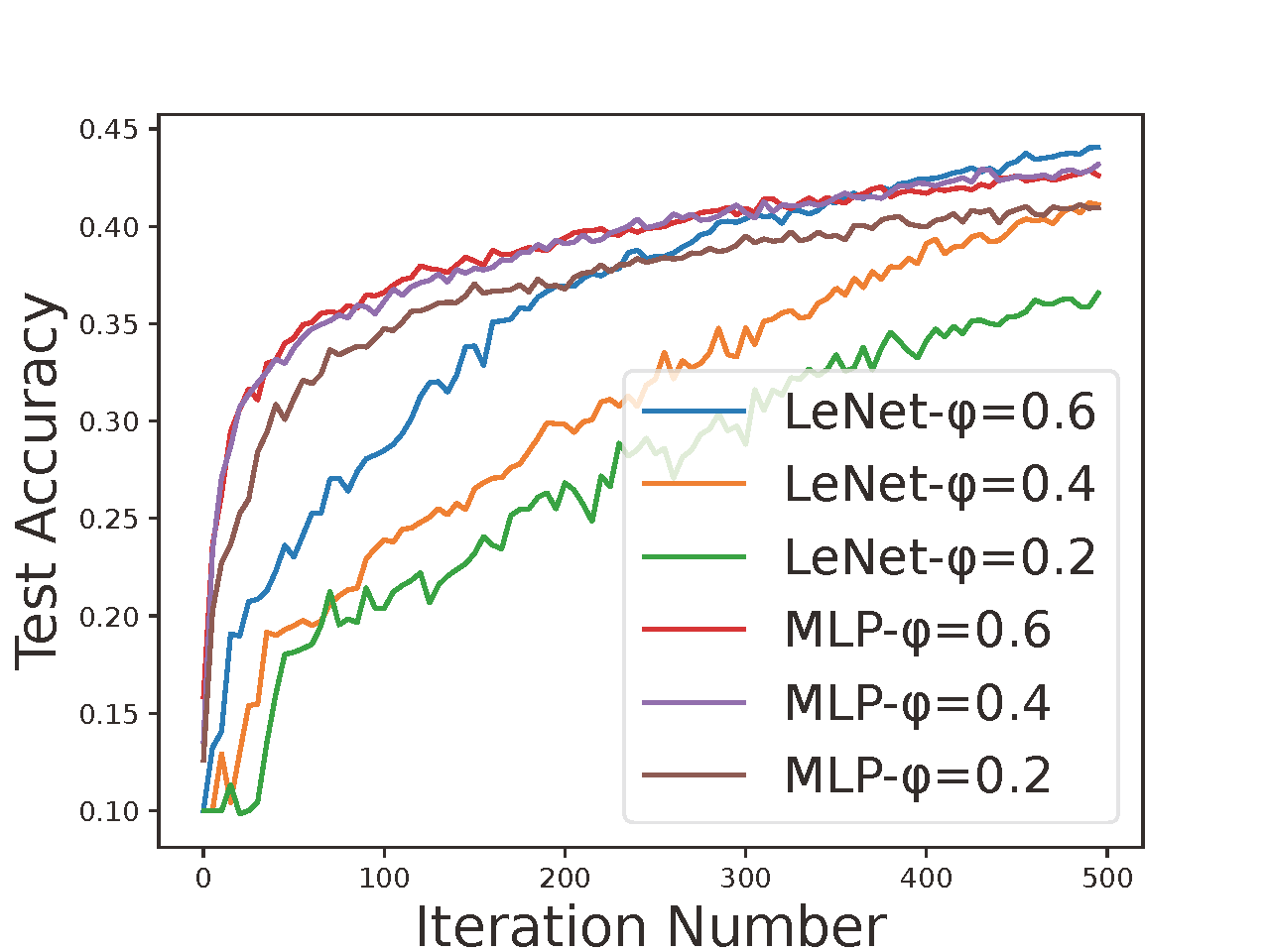}
    }
    \vfill
    \caption{Test accuracy of RAGA with different data heterogeneity parameters on $\bar{C}_{\alpha} = 0.4$.}\label{fig:data4}
\end{figure}

\begin{figure*}[h]
    \centering
    \subfigure[Sign-flip attack with $\bar{C}_{\alpha} = 0.1$.]{
    \includegraphics[width = 0.2\textwidth]{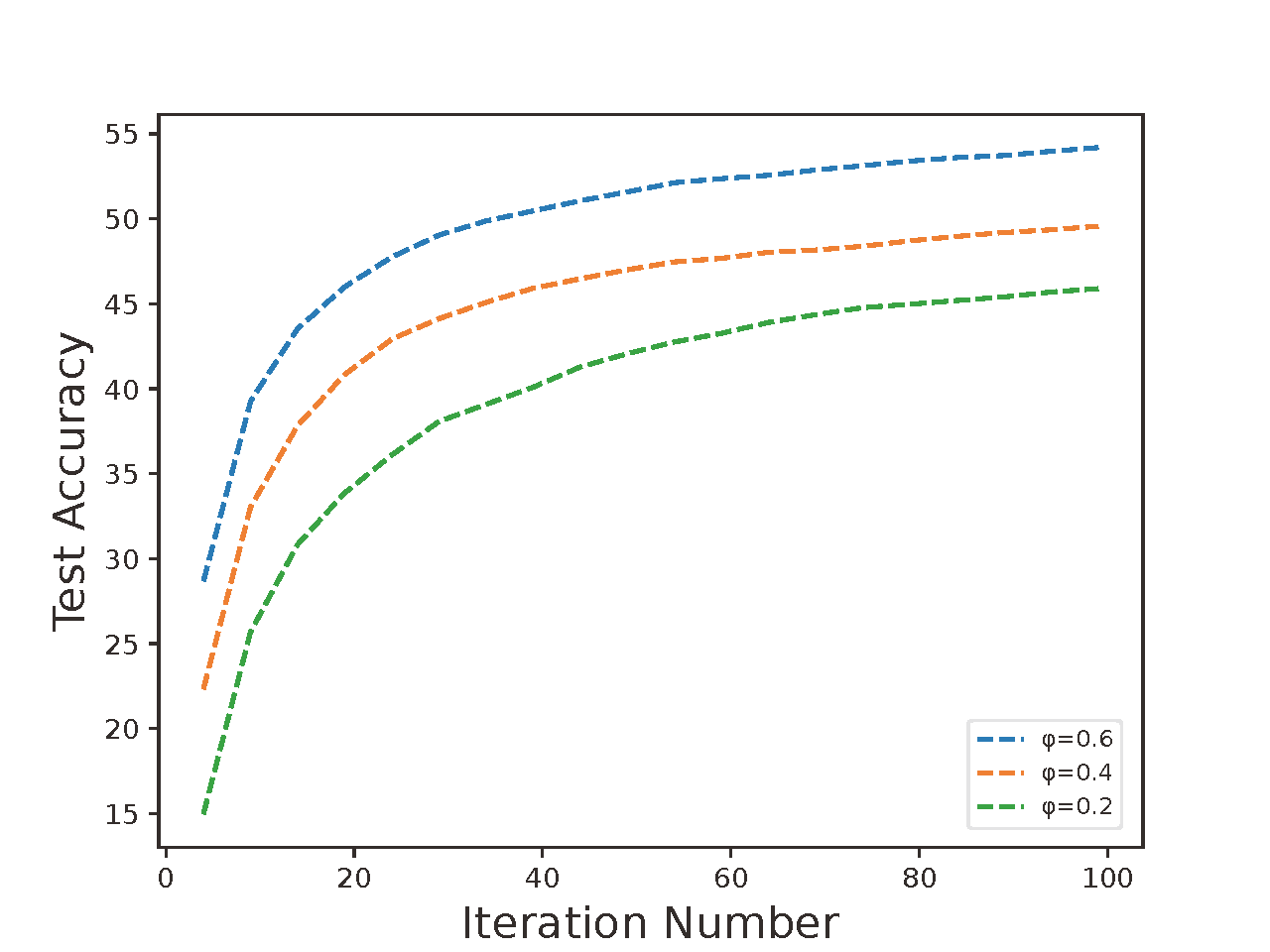}
    }
    \subfigure[LIE attack with $\bar{C}_{\alpha} = 0.1$.]{
    \includegraphics[width = 0.2\textwidth]{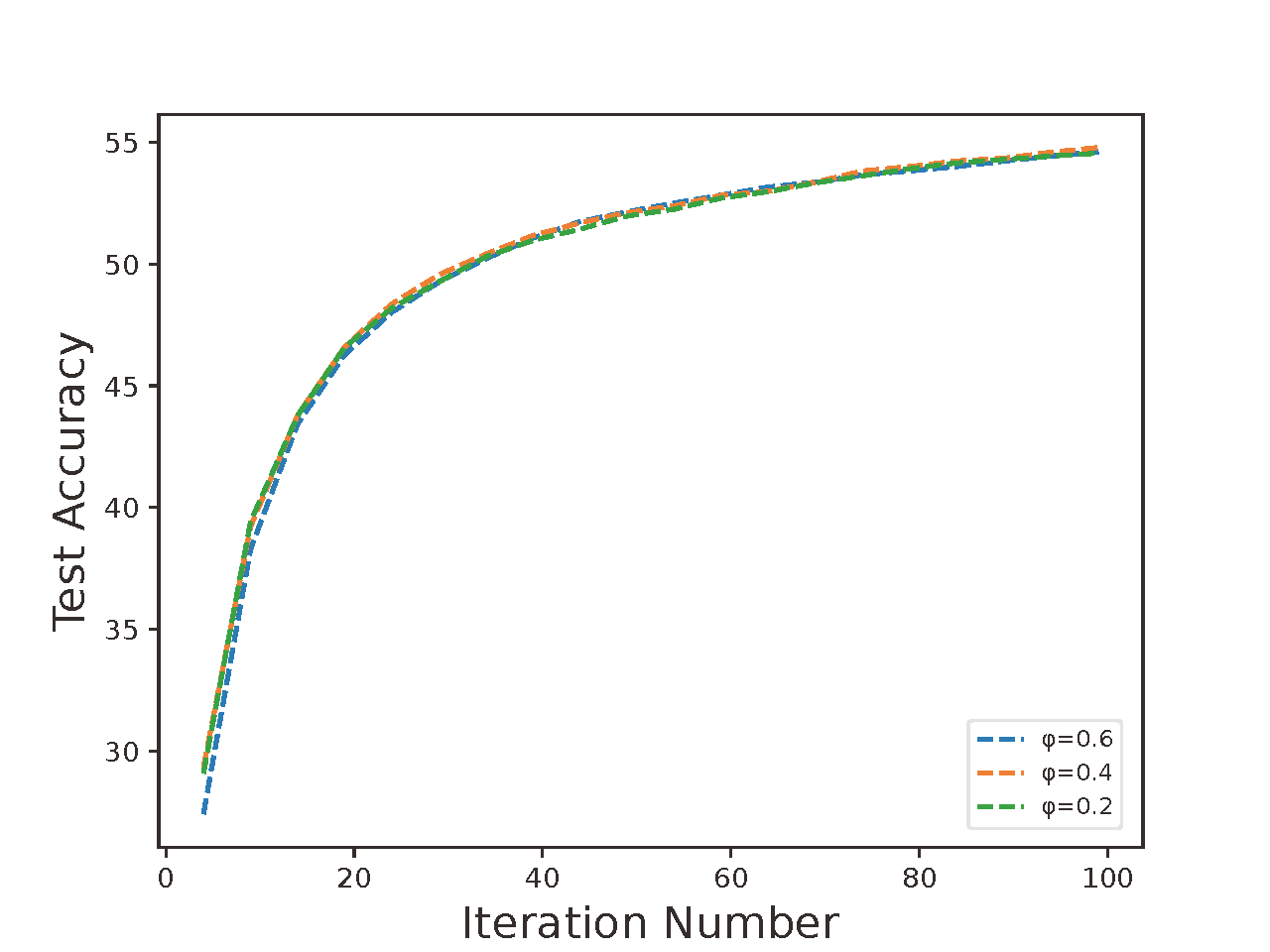}
    }
    \subfigure[Gaussian attack with $\bar{C}_{\alpha} = 0.4$.]{
    \includegraphics[width = 0.2\textwidth]{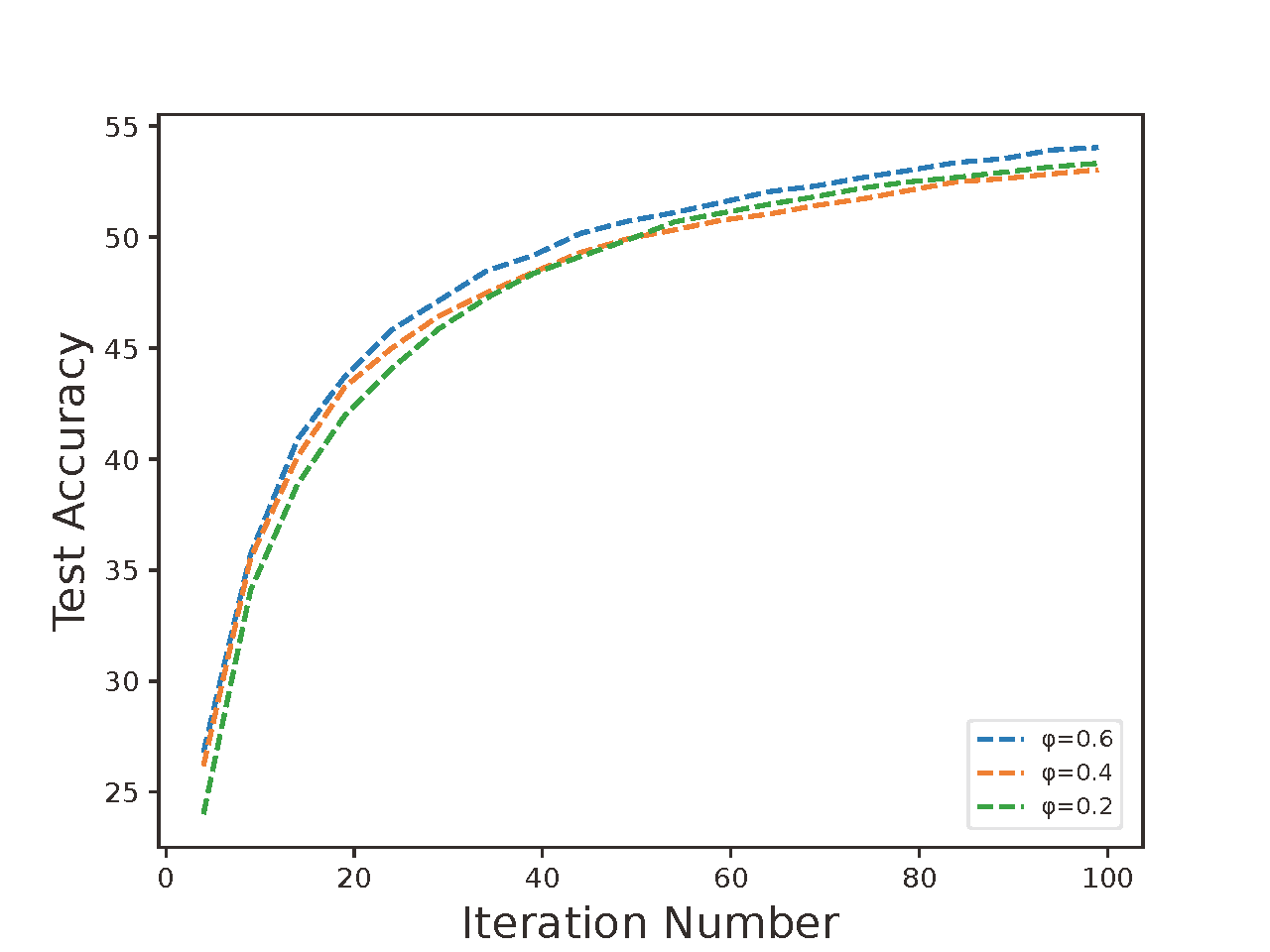}
    }
    \subfigure[LIE attack with $\bar{C}_{\alpha} = 0.4$.]{
    \includegraphics[width = 0.2\textwidth]{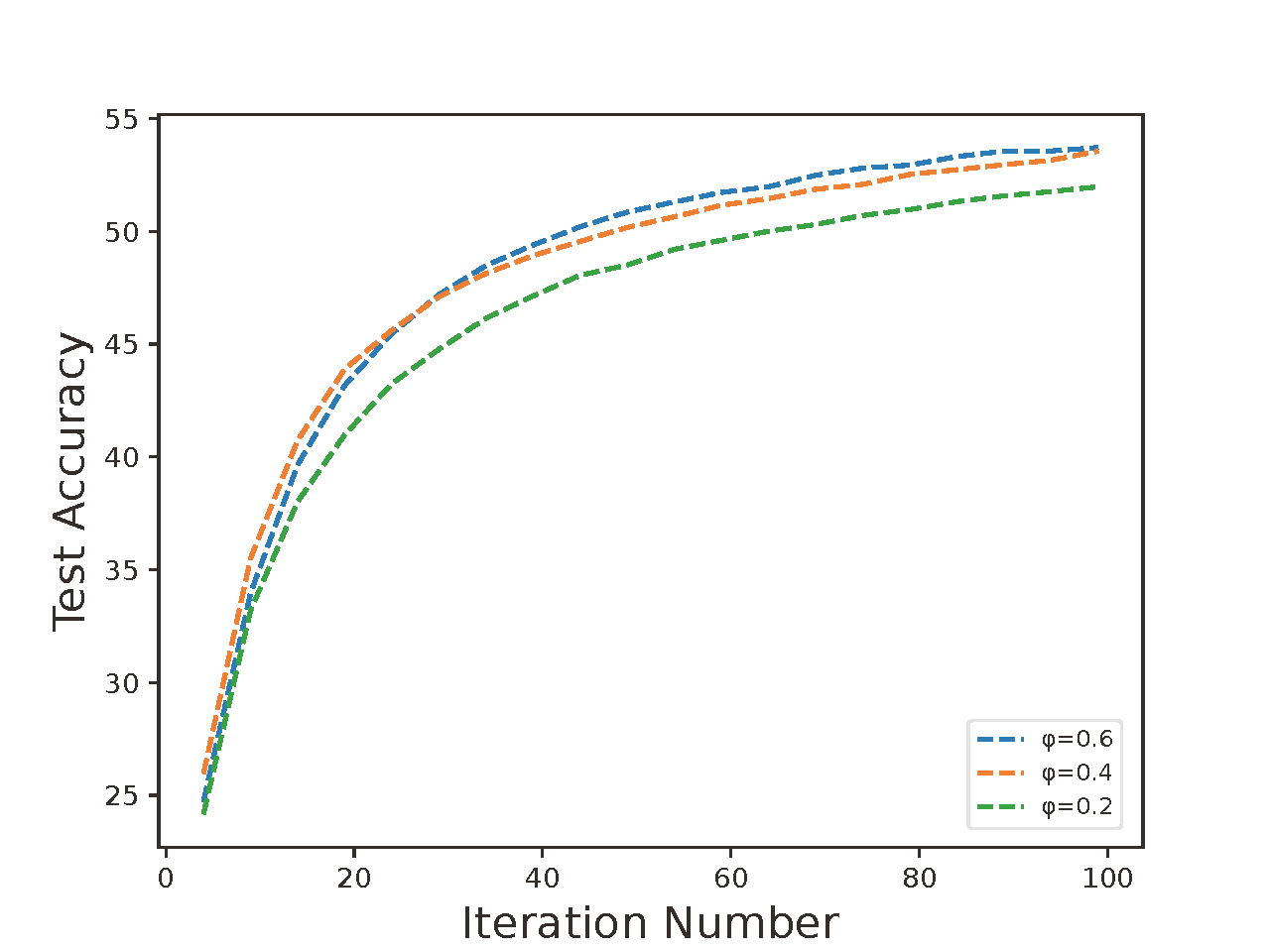}
    }
    \caption{Test accuracy of RAGA with different data heterogeneity parameters on CIFAR100 dataset.}\label{fig:data100}
\end{figure*}

\begin{figure}[h]
    \centering
    \subfigure[MNIST, and LeNet.]{
    \includegraphics[width = 0.2\textwidth]{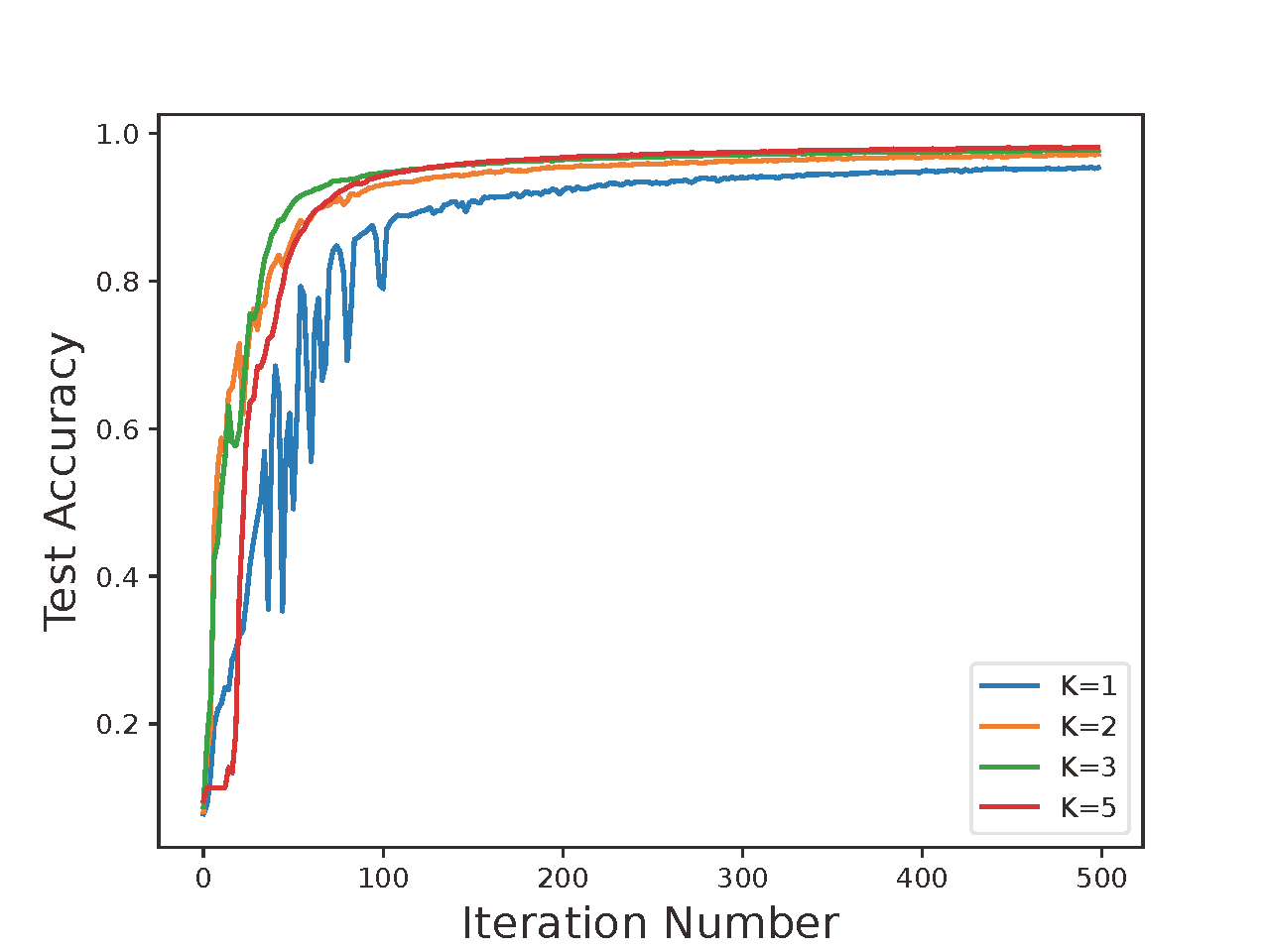}
    }
    \subfigure[MNIST, and MLP.]{
    \includegraphics[width = 0.2\textwidth]{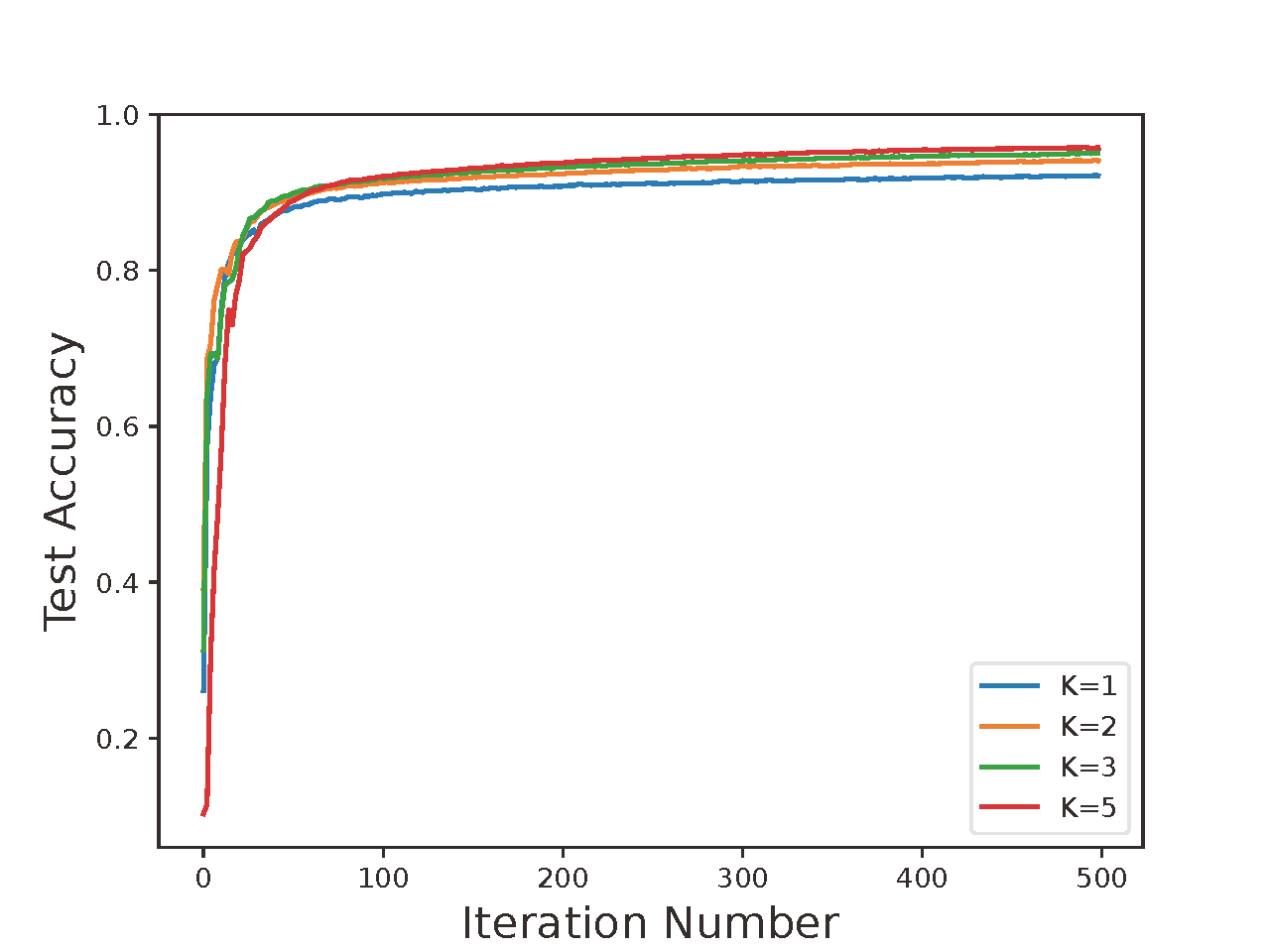}
    }
    \subfigure[CIFAR10, and LeNet.]{
    \includegraphics[width = 0.2\textwidth]{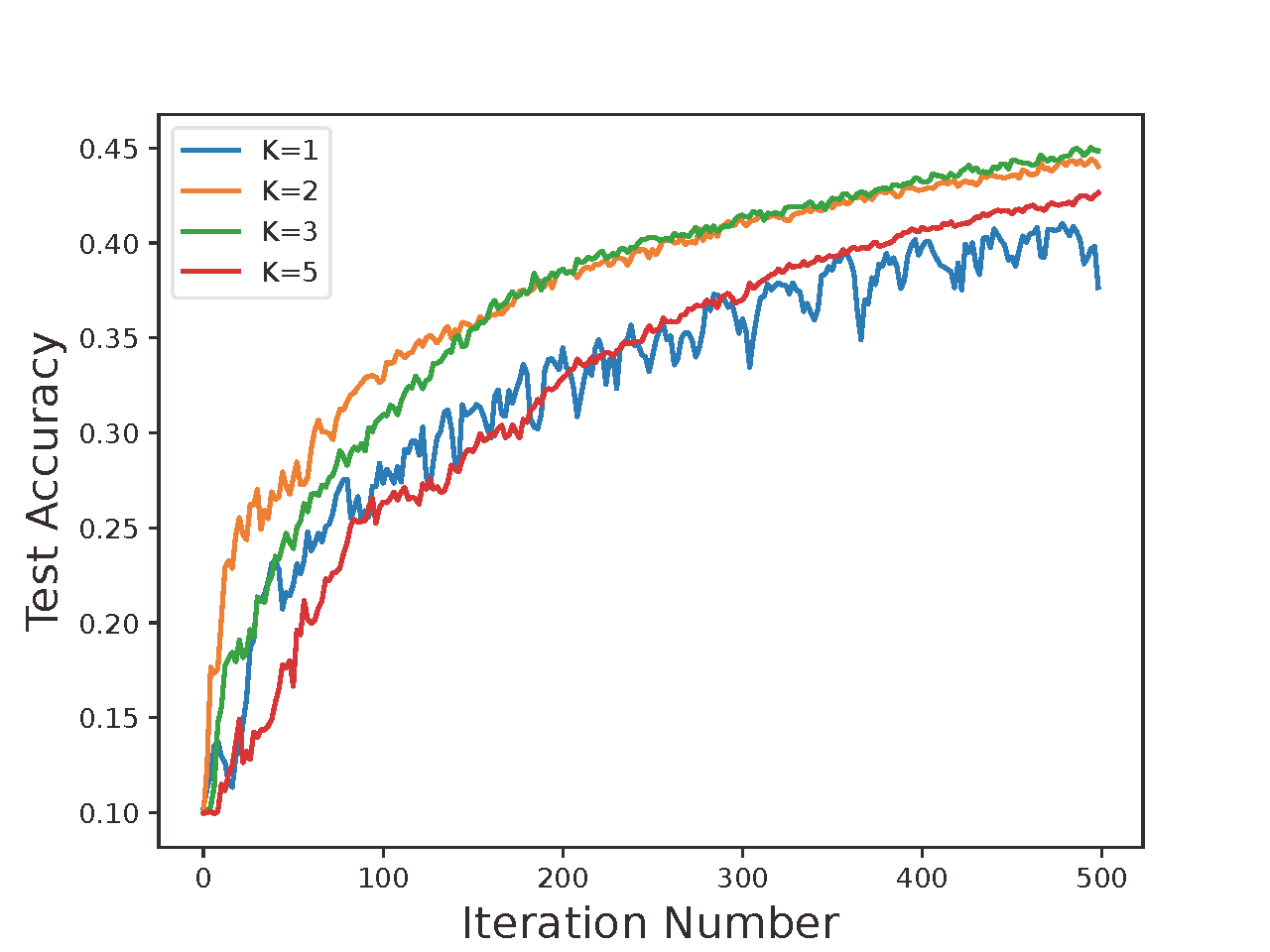}
    }
    \subfigure[CIFAR10, and MLP.]{
    \includegraphics[width = 0.2\textwidth]{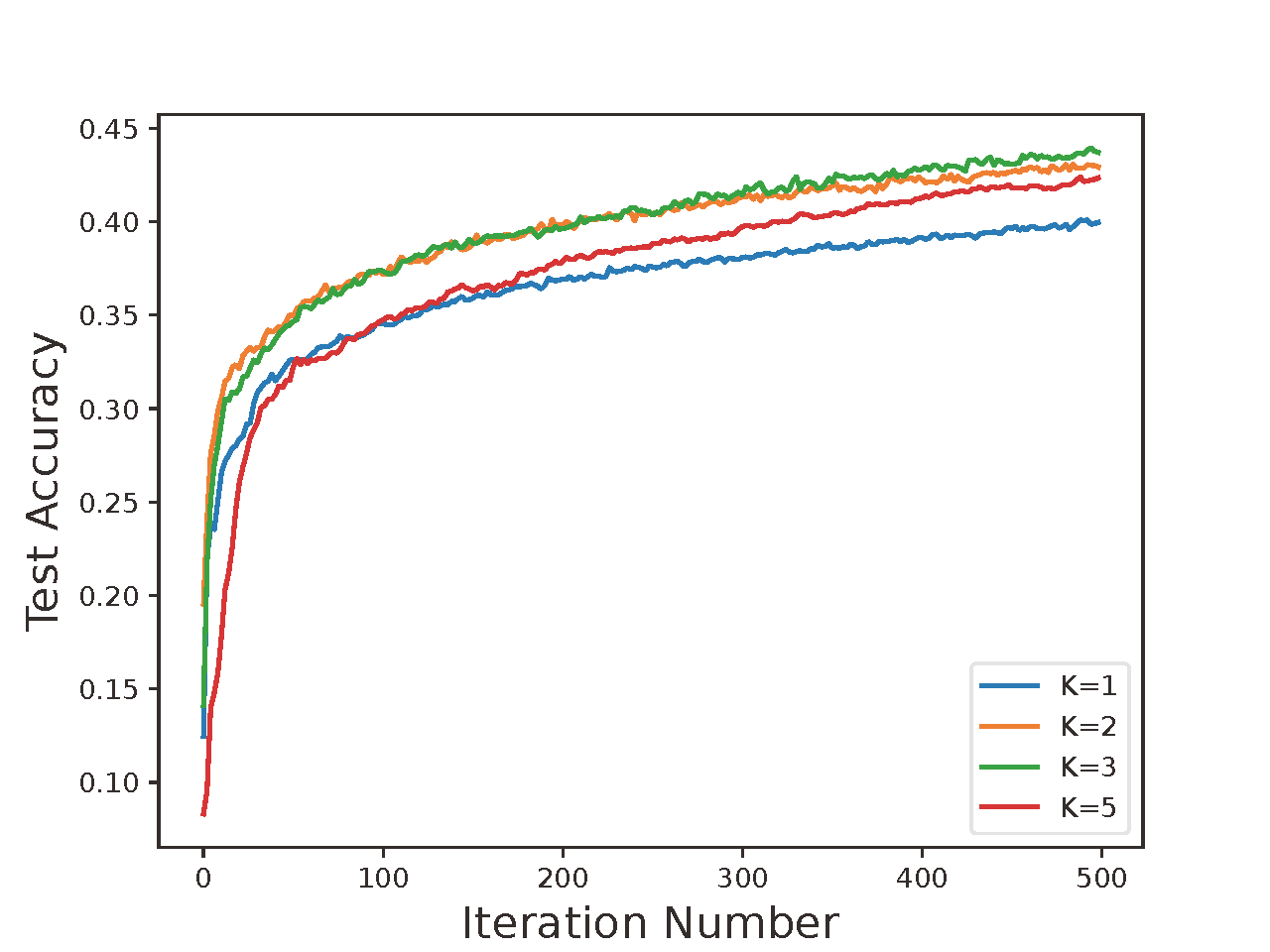}
    }
    \vfill
    \caption{Test accuracy of RAGA with different local epoch on Gaussian attack on $\bar{C}_{\alpha} = 0.2$ with $\phi = 0.6$.}\label{fig:de26}
\end{figure}

\begin{figure}[h]
    \centering
    \subfigure[MNIST, and LeNet.]{
    \includegraphics[width = 0.2\textwidth]{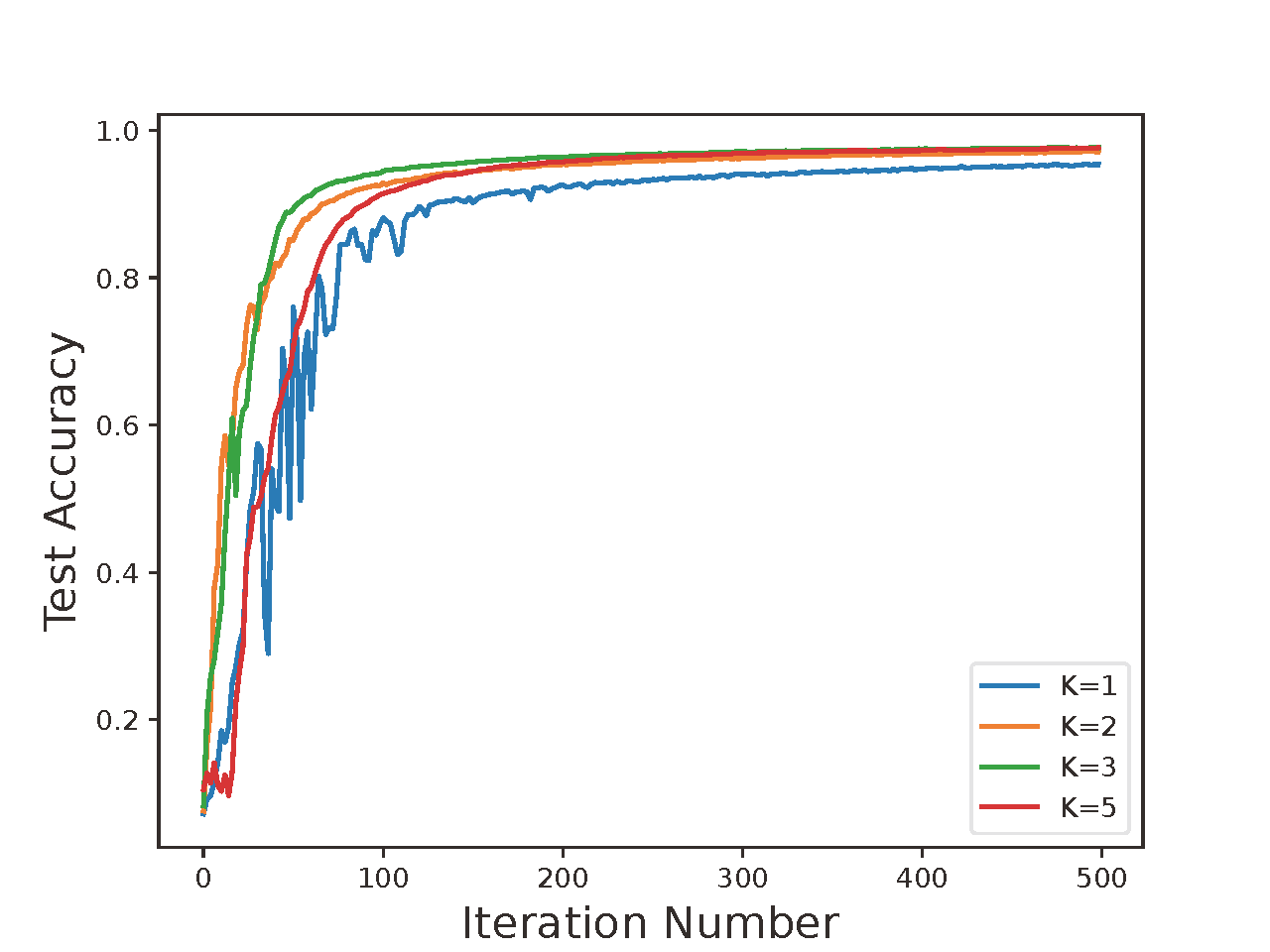}
    }
    \subfigure[MNIST, and MLP.]{
    \includegraphics[width = 0.2\textwidth]{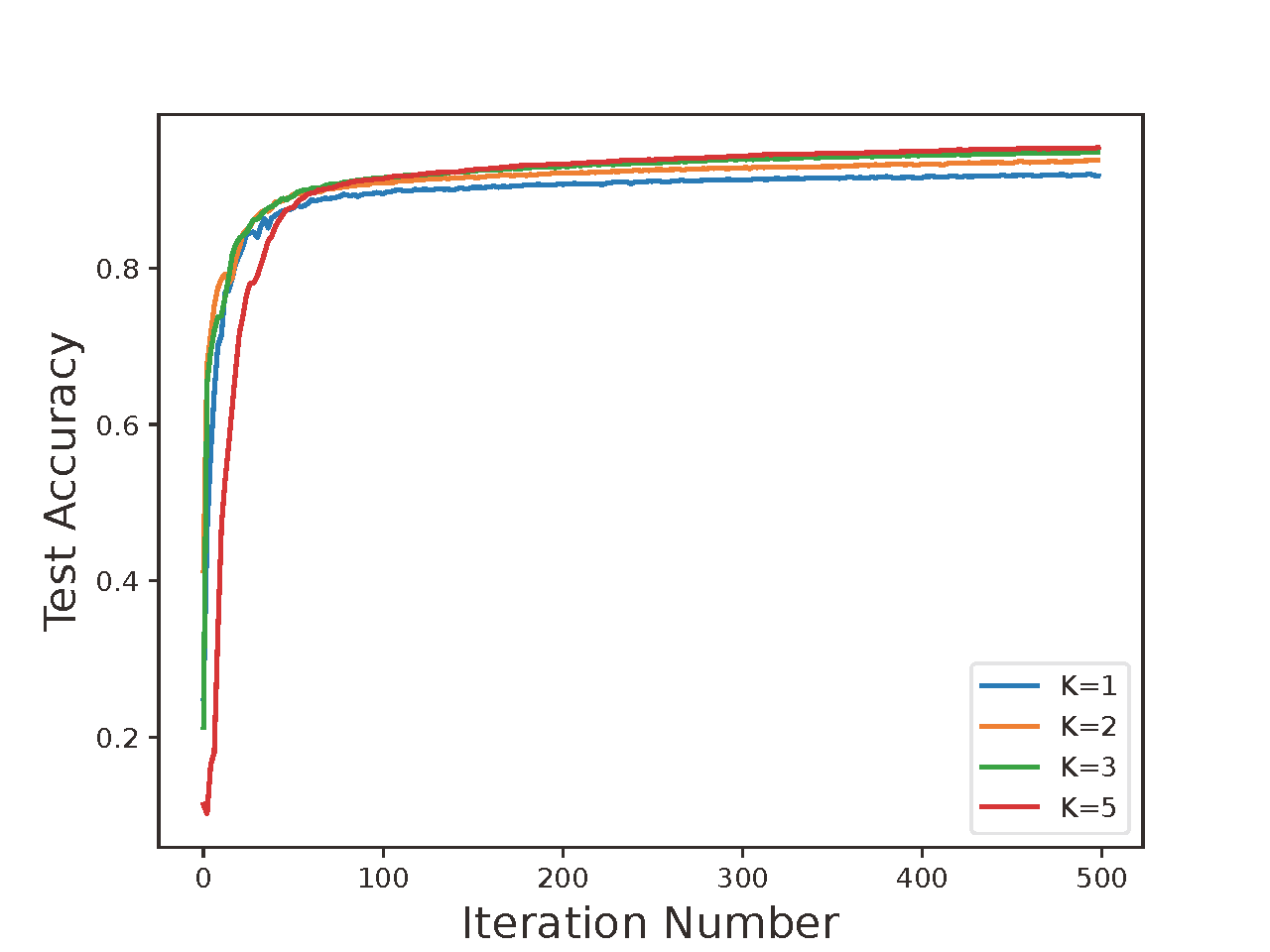}
    }
    \subfigure[CIFAR10, and LeNet.]{
    \includegraphics[width = 0.2\textwidth]{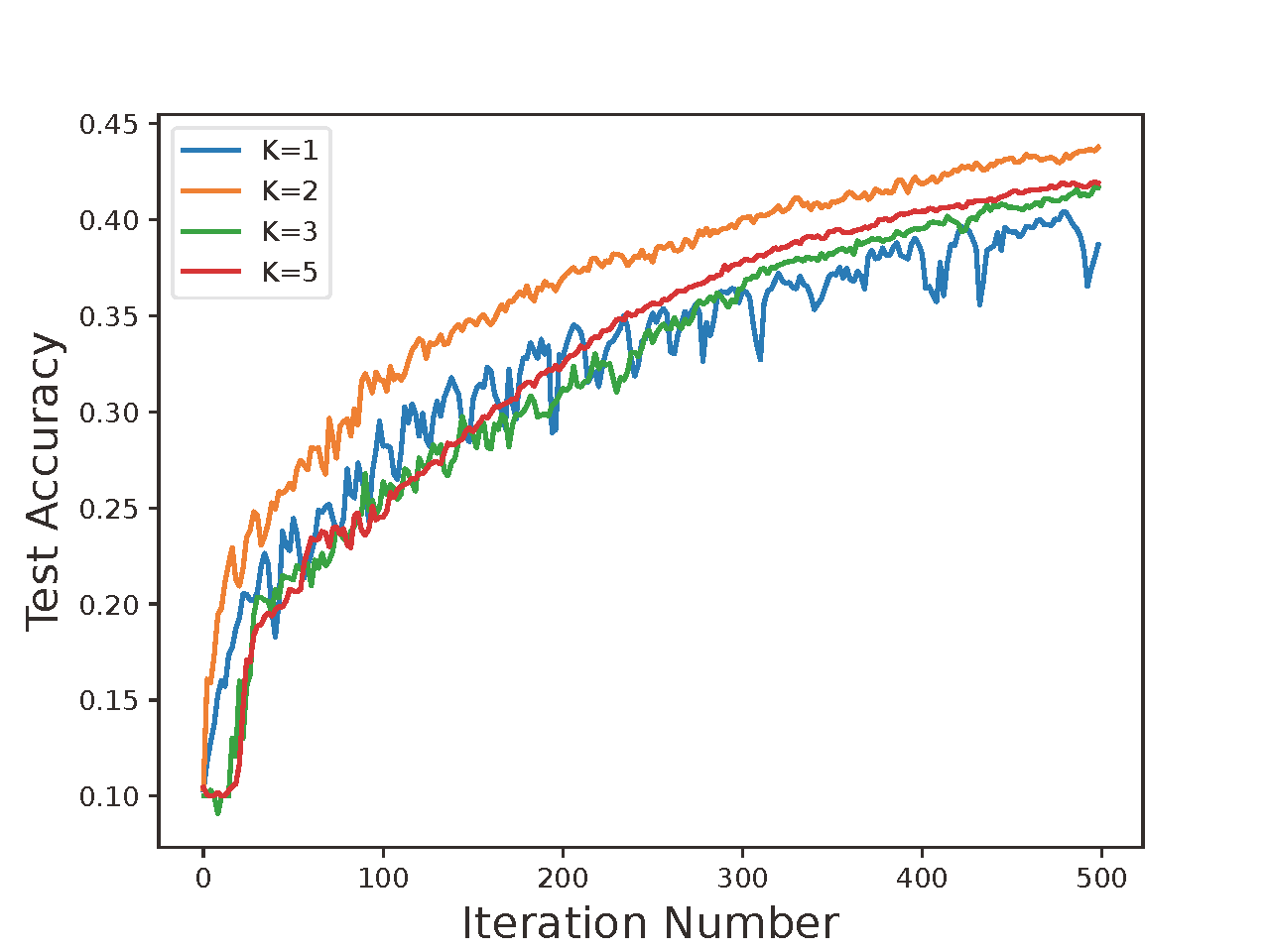}
    }
    \subfigure[CIFAR10, and MLP.]{
    \includegraphics[width = 0.2\textwidth]{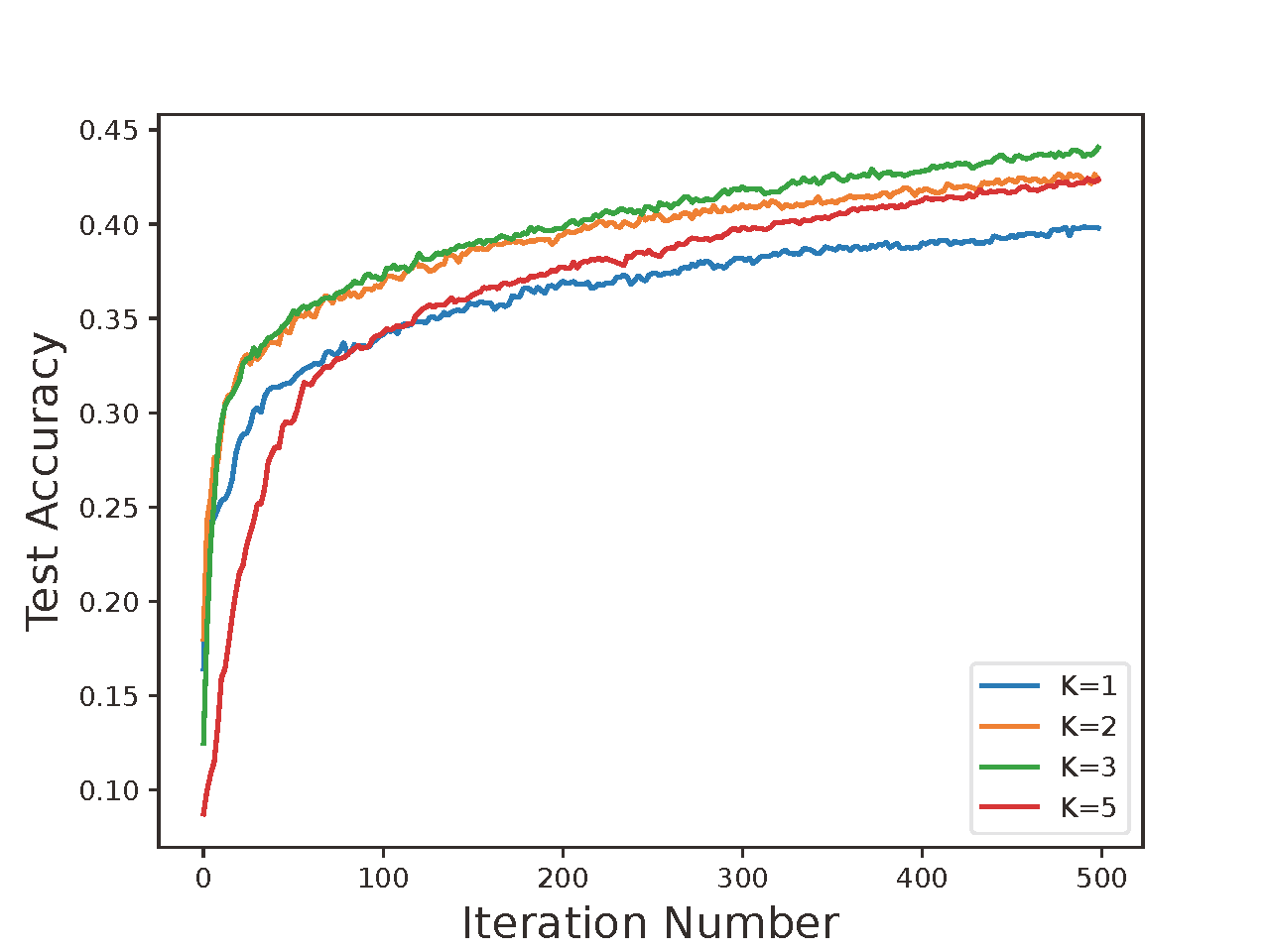}
    }
    \vfill
    \caption{Test accuracy of RAGA with different local epoch on Gaussian attack on $\bar{C}_{\alpha} = 0.2$ with $\phi = 0.2$.}\label{fig:de22}
\end{figure}

We begin with Table \ref{tab:accml}, \ref{tab:accmm}, \ref{tab:accc} and \ref{tab:acc100}, which presents the maximum test accuracy of RAGA and baselines under varying levels of Byzantine attack intensity $\bar{C}_{\alpha}$, across different data heterogeneity concentration parameters $\phi$, and for two learning models. Fig \ref{fig:mta6} and \ref{fig:cta6} illustrates the convergence of RAGA and the baselines under Sign-flip attack and data heterogeneity concentration parameters $\phi=0.6$. 
Fig \ref{fig:m4}, \ref{fig:c6} and \ref{fig:c1006}
present the maximum test accuracy of RAGA and the baseline methods under various Byzantine ratios, excluding the results reported in Table \ref{tab:accml}, \ref{tab:accmm} and \ref{tab:accc}. 
Fig \ref{fig:data4}, \ref{fig:data100} present the test accuracy of RAGA with different data heterogeneity parameters under various Byzantine ratios. 
And Fig \ref{fig:de26} and \ref{fig:de22} illustrates convergence of RAGA under different local epoch $K$ on Gaussian attack with $\bar{C}_{\alpha} = 0.2$. Following this, we will discuss the performance comparison between RAGA and the baselines in different scenarios.

\textbf{No Byzantine attack:} For the easy dataset (MNIST), Table \ref{tab:accml}, \ref{tab:accmm} and Fig. \ref{fig:m4} show that RAGA outperforms all other methods across a range of data heterogeneity concentration parameters, $\phi$, and two different learning models. Specifically, compared to the baselines, RAGA increases test accuracy by 2.19\% and 2.85\% for the LeNet and MLP models, respectively, with $\phi = 0.6$, and by 2.49\% and 2.52\% for the LeNet and MLP models with $\phi = 0.2$. For the CIFAR10 dataset, Table \ref{tab:accc} and Fig. \ref{fig:c6} demonstrate that RAGA also achieves the highest performance across different data heterogeneity concentrations, particularly for $\phi = 0.4$ with the LeNet and MLP models. In comparison to the baselines, RAGA increases test accuracy by 1.07\% and 3.68\% for the LeNet and MLP models, respectively, at $\phi = 0.4$, and consistently shows the highest accuracy at $\phi = 0.6$. {As presented in Table \ref{tab:acc100}, when applied to the CIFAR100 dataset, RAGA demonstrates a notable improvement in test accuracy, achieving a 3.89\% increase compared to the robust baselines. In contrast, the performance gap between RAGA and FedAvg is minimal, narrowing to just 0.1\%.} 

\begin{table*}[t]
\caption{\centering The maximum test accuracy(\%) of 100 iteration number of RAGA, FedAvg, Median, Byrd-SAGA, RFA and MCA with different types, and ratios of Byzantine attack on CIFAR100 dataset and VGG16 model.}
\label{tab:acc100}
\resizebox{\textwidth}{!}{
\renewcommand{\arraystretch}{1.5}
\begin{tabular}{c|cc|c|cccc|cccc|cccc}
\multirow{3}{*}{Model}          & \multicolumn{2}{c|}{Attack Name}                       & \textbf{No Attack} & \multicolumn{4}{c|}{\textbf{Gaussian Attack}}                                             & \multicolumn{4}{c|}{\textbf{Sign-flip Attack}}                                            & \multicolumn{4}{c}{\textbf{LIE Attack}}                                                   \\ \cline{2-16} 
                                & \multicolumn{2}{c|}{$\bar{C}_{\alpha}$}                & \multirow{2}{*}{0} & \multirow{2}{*}{0.1} & \multirow{2}{*}{0.2} & \multirow{2}{*}{0.3} & \multirow{2}{*}{0.4} & \multirow{2}{*}{0.1} & \multirow{2}{*}{0.2} & \multirow{2}{*}{0.3} & \multirow{2}{*}{0.4} & \multirow{2}{*}{0.1} & \multirow{2}{*}{0.2} & \multirow{2}{*}{0.3} & \multirow{2}{*}{0.4} \\ \cline{2-3}
                                & \multicolumn{1}{c|}{$\phi$}               & Algorithms &                    &                      &                      &                      &                      &                      &                      &                      &                      &                      &                      &                      &                      \\ \hline
\multirow{6}{*}{\textbf{VGG16}} & \multicolumn{1}{c|}{\multirow{6}{*}{0.6}} & RAGA       & 54.57              & \textbf{54.85}       & \textbf{54.42}       & \textbf{53.73}       & \textbf{54.03}       & \textbf{54.19}       & \textbf{54.48}       & \textbf{53.66}       & \textbf{54.40}       & \textbf{54.61}       & \textbf{54.55}       & \textbf{54.62}       & \textbf{53.71}       \\
                                & \multicolumn{1}{c|}{}                     & FedAvg     & \textbf{54.67}     & 29.59                & 20.73                & 14.06                & 10.02                & 17.45                & 1.00                 & 1.00                 & 21.88                & 53.38                & 52.03                & 50.71                & 49.86                \\
                                & \multicolumn{1}{c|}{}                     & Median     & 49.51              & 49.89                & 49.50                & 48.79                & 49.81                & 49.17                & 49.15                & 47.42                & 48.01                & 49.39                & 48.97                & 48.80                & 48.75                \\
                                & \multicolumn{1}{c|}{}                     & SAGA       & 50.50              & 50.78                & 50.19                & 49.11                & 48.64                & 49.88                & 50.30                & 49.60                & 49.38                & 50.47                & 50.05                & 49.66                & 48.91                \\
                                & \multicolumn{1}{c|}{}                     & RFA        & 50.68              & 50.80                & 50.22                & 49.01                & 48.57                & 50.62                & 50.29                & 50.09                & 49.55                & 50.42                & 50.04                & 49.73                & 48.92                \\
                                & \multicolumn{1}{c|}{}                     & MCA        & 50.48              & 50.64                & 50.17                & 49.21                & 48.66                & 49.41                & 7.32                 & 1.00                 & 1.00                 & 50.43                & 50.00                & 49.63                & 48.88               
\end{tabular}
}
\end{table*}

\textbf{Gaussian attack:} First, we evaluate the performance of RAGA and the baseline methods on the MNIST dataset. As shown in Table \ref{tab:accml} and \ref{tab:accmm}, RAGA achieves the highest test accuracy among all FL robust algorithms across different Byzantine attack ratios, two learning models, and two distinct data heterogeneity concentration parameters. For the LeNet model, RAGA shows an improvement in test accuracy by 2.08\% to 2.34\% for the data heterogeneity concentration parameter $\phi=0.6$, and by 2.24\% to 2.69\% for $\phi=0.2$, across four different Byzantine ratios ($\bar{C}_{\alpha}=0.1, 0.2, 0.3, 0.4$). For the MLP model, RAGA improves test accuracy by 2.65\% to 2.97\% for $\phi=0.6$, and by 2.57\% to 2.77\% for $\phi=0.2$, across the same four Byzantine ratios. Additionally, as shown in Fig. \ref{fig:m4}, RAGA outperforms five other FL robust algorithms across various Byzantine attack ratios, regardless of the learning model type, when the data heterogeneity concentration parameter is set to $\phi=0.4$. Next, we analyze the performance of RAGA and the baseline methods on the more complex CIFAR-10 dataset. With the data heterogeneity concentration parameter $\phi=0.4$, RAGA consistently outperforms all baseline methods in terms of test accuracy, as shown in Table \ref{tab:accc}. Specifically, for the LeNet model, RAGA achieves improvements by 0.03\% to 1.64\%, while for the MLP model, improvements by 3.25\% to 3.91\%, across four different Byzantine ratios ($\bar{C}_{\alpha}=0.1, 0.2, 0.3, 0.4$). Furthermore, RAGA consistently performs better than all baseline methods based on Fig. \ref{fig:c6}, irrespective of whether the LeNet or MLP model is used, when the data heterogeneity concentration parameter is set to $\phi=0.6$. {For the CIFAR100 dataset, Table \ref{tab:acc100}  demonstrates that RAGA achieves a maximum accuracy improvement of 4.05\% to 4.52\% across four distinct Byzantine ratios when compared to baseline methods. Additionally, as shown in Fig. \ref{fig:c1006}, RAGA consistently outperforms all baselines on the VGG16 model under Gaussian attacks.}

\textbf{Sign-flip attack:} First, for the relatively simple dataset (MNIST), as shown in Table \ref{tab:accml} and \ref{tab:accmm}, all FL robust methods, with the exception of MCA, effectively mitigate the sign-flip attack across various Byzantine ratios. In contrast, MCA fails to defend against this attack at higher Byzantine ratios. For the LeNet model, Table \ref{tab:accml} and \ref{tab:accmm} show that RAGA outperforms all other methods in terms of test accuracy, improving by 2.33\% to 2.71\% with the data heterogeneity concentration parameter of $\phi=0.6$, and by 3.06\% to 3.75\% with $\phi=0.2$, across four different Byzantine ratios. For the MLP model, RAGA also achieves the highest test accuracy among all baseline methods, improving by 2.68\% to 3.42\% for $\phi=0.6$, and by -0.97\% to 2.68\% for $\phi=0.2$, across the same four Byzantine ratios. Then, for the more complex dataset (CIFAR10), as shown in Table \ref{tab:accc}, similar to the results on the MNIST dataset, all FL robust methods, except MCA, effectively mitigate the sign-flip attack across various Byzantine ratios. With the data heterogeneity concentration parameter of $\phi=0.4$, RAGA performs the best in most cases (except for the LeNet model with $\bar{C}_{\alpha}=0.2$), with test accuracy improvements of -0.21\% to 4.01\% for the LeNet model, and 2.8\% to 5.72\% for the MLP model, across the four Byzantine ratios. Additionally, as shown in Fig. \ref{fig:mta6} and \ref{fig:cta6}, RAGA achieves the fastest convergence speed among all baseline methods with the same local epoch $K^t=3$ and data heterogeneity concentration parameter $\phi=0.6$, regardless of the learning model type or the Byzantine ratio. {According to Table \ref{tab:acc100}, for the CIFAR100 dataset, RAGA consistently boosts test accuracy by 3.57\% to 4.85\% across four distinct Byzantine ratios compared to baselines when $\phi = 0.6$. Fig. \ref{fig:c1006} further illustrates that RAGA outperforms all baselines across the evaluated conditions. Regarding the FedAvg result with $C_{\alpha} = 0.4$ that exhibits a seemingly "maximum" accuracy, this outcome should not be attributed to statistical abnormality. Instead, the extremely high loss value of 856667.4125 indicates that the model likely converged to a distant saddle point in the loss landscape. }

\textbf{LIE attack:} Using the simple MNIST dataset, RAGA achieves the best performance among six baseline methods across four Byzantine ratios. As shown in Table \ref{tab:accml} and \ref{tab:accmm}, for the LeNet model, RAGA improves test accuracy by 2.25\% to 2.45\% when the data heterogeneity concentration parameter is $\phi=0.6$, and by 2.57\% to 3.00\% when $\phi=0.2$, across four different Byzantine ratios. Similarly, for the MLP model, RAGA outperforms all baseline methods, improving test accuracy by 2.7\% to 2.97\% when the data heterogeneity concentration parameter is $\phi=0.6$, and by 2.46\% to 2.89\% when $\phi=0.2$. Furthermore, as shown in Fig. \ref{fig:m4}, with a data heterogeneity concentration parameter of $\phi=0.4$, RAGA achieves the best test accuracy performance among five FL Byzantine-robust algorithms, regardless of the type of learning model. Additionally, using the more complex CIFAR-10 dataset, RAGA outperforms most baseline methods (with the exception of the LeNet model at a Byzantine ratio of $\bar{C}_{\alpha}=0.2$). With the data heterogeneity concentration parameter of $\phi=0.4$, RAGA enhances test accuracy by -1.12\% to 1.13\% on the LeNet model, and by 3.23\% to 3.93\% on the MLP model, across four Byzantine ratios, compared to other FL Byzantine-robust algorithms. Finally, as shown in Fig. \ref{fig:c6}, under the LIE attack, RAGA also outperforms all six baseline methods with the data heterogeneity concentration parameter $\phi=0.6$. {Regarding the CIFAR100 dataset, RAGA demonstrates superior performance against LIE attacks, achieving improvements in test accuracy ranging from 1.23\% to 3.91\% across four distinct Byzantine ratios when compared to baseline methods, as shown in Table \ref{tab:acc100}. Additionally, Fig. \ref{fig:c1006} confirms that RAGA establishes state-of-the-art (SoTA) performance under these experimental settings, outperforming all baseline approaches evaluated.}

\textbf{Data Heterogeity:} We first analyze the convergence performance of RAGA on the MNIST dataset. As shown in Table \ref{tab:accml} and \ref{tab:accmm}, varying data heterogeneity parameters has minimal impact on the test accuracy of RAGA, with the maximum difference being only 0.66\%, as further confirmed by Fig. \ref{fig:data4}. In contrast, for the more complex CIFAR10 dataset, the effect of different data heterogeneity parameters on RAGA's test accuracy is more pronounced, as depicted in Fig. \ref{fig:data4}. Additionally, we observe that the MLP model is more robust to data heterogeneity than the LeNet model, particularly in the presence of the Sign-flip attack. {For the CIFAR100 dataset, as depicted in Fig. \ref{fig:data100}, the experimental results on the VGG16 model demonstrate that the impact of attack type on accuracy surpasses that of data heterogeneity. This distinction is notably evident when comparing the Sign-flip and LIE attacks, where differing attack strategies produce more significant effects on model accuracy than variations in data distribution heterogeneity.} 

\textbf{Local update:} From Fig. \ref{fig:de26} and \ref{fig:de22}, it is clear that a larger local update size $K$ does not necessarily result in better performance, which is consistent with our theoretical analysis, irrespective of the learning model or dataset. With the LeNet model, the convergence curve exhibits significant fluctuations when $K=1$, leading to suboptimal convergence performance. When $K=5$, although the fluctuations are reduced, the increased number of training rounds adversely affects the algorithm's performance. In contrast, for $K=2$ or $K=3$, the choice of $K$ demonstrates distinct advantages across both datasets and learning models. These experimental findings are in agreement with our theoretical proof. 


In summary, the above numerical experiments demonstrate that RAGA not only outperforms its competitors across different ratios and types of Byzantine attacks but also performs comparably to FedAvg, which excels when there is no Byzantine attack. It is also noteworthy that the RAGA algorithm can converge under various types of Byzantine attacks. Lastly, all these results are based on non-IID datasets with three different data heterogeneity concentration parameters.

%% file: conclusion.tex
\section{Conclusion} \label{sec:conclu}

In this paper, we propose a novel FL aggregation strategy, RAGA, designed to be robust against Byzantine attacks and data heterogeneity. Our proposed RAGA leverages the geometric median for aggregation, allowing flexible selection of the round number for local updates.
Rigorous proofs reveal that RAGA can achieve a convergence rate of $\mathcal{O}({1}/{T^{2/3- \delta}})$ for non-convex loss functions, with $\delta \in (0, 2/3)$, and a linear rate for strongly-convex loss functions, provided that the fraction of the dataset from malicious users is less than half. Experimental results on heterogeneous datasets demonstrate that RAGA consistently outperforms baseline methods in convergence performance under varying intensities of Byzantine attacks.

%% file: biography.tex
\begin{IEEEbiography}[{\includegraphics[width=1.1in,height=1.375in,clip,keepaspectratio]{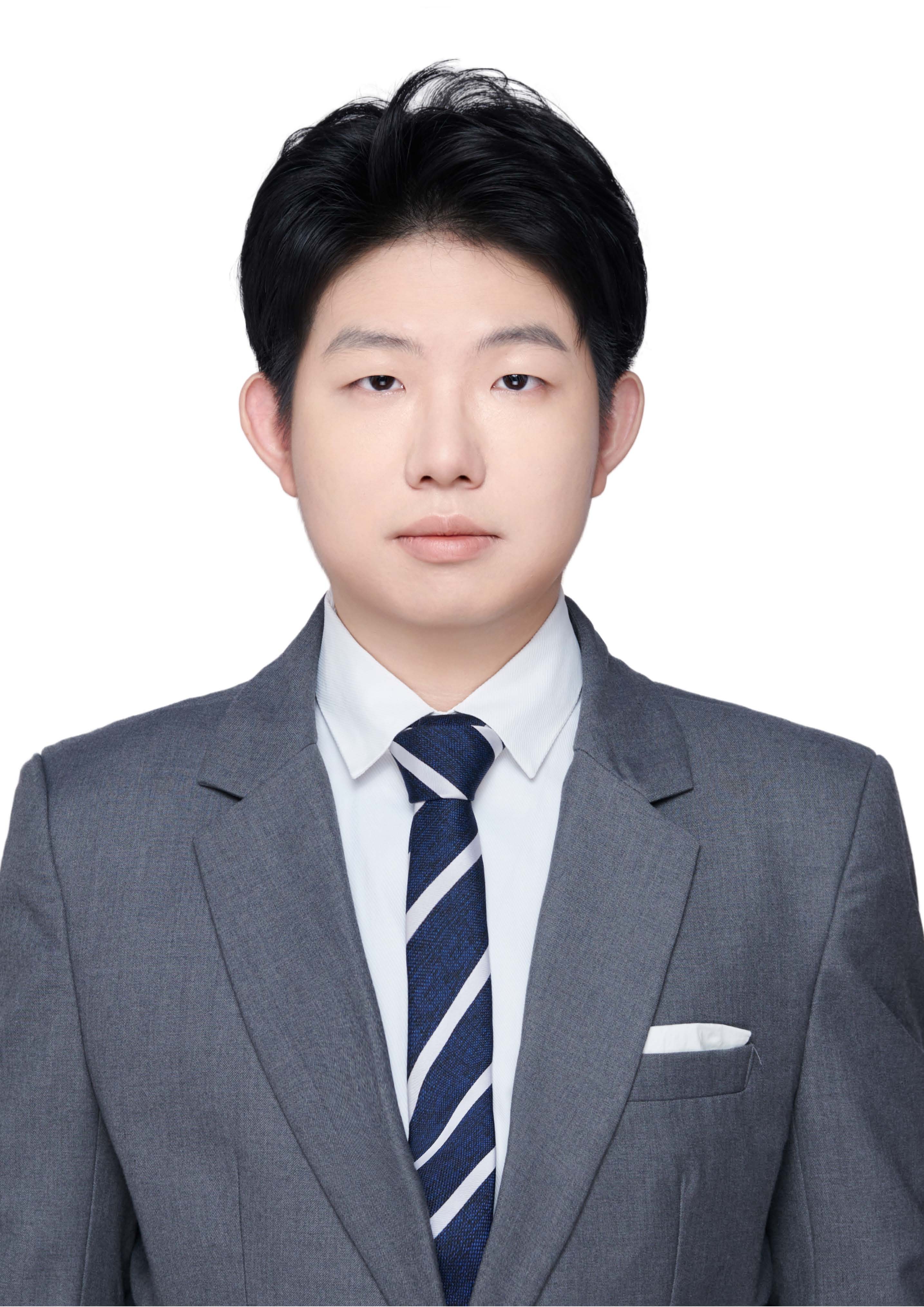}}]
    {Shiyuan Zuo} received the B.E. degree from the Beijing Institute of Technology, China, in 2021. He is currently pursuing a Ph.D. degree at the School of Cyberspace Science and Technology, Beijing Institute of Technology, China. His research interests include federated learning and privacy protection.
\end{IEEEbiography}

\begin{IEEEbiography}[{\includegraphics[width=1.1in,height=1.375in,clip,keepaspectratio]{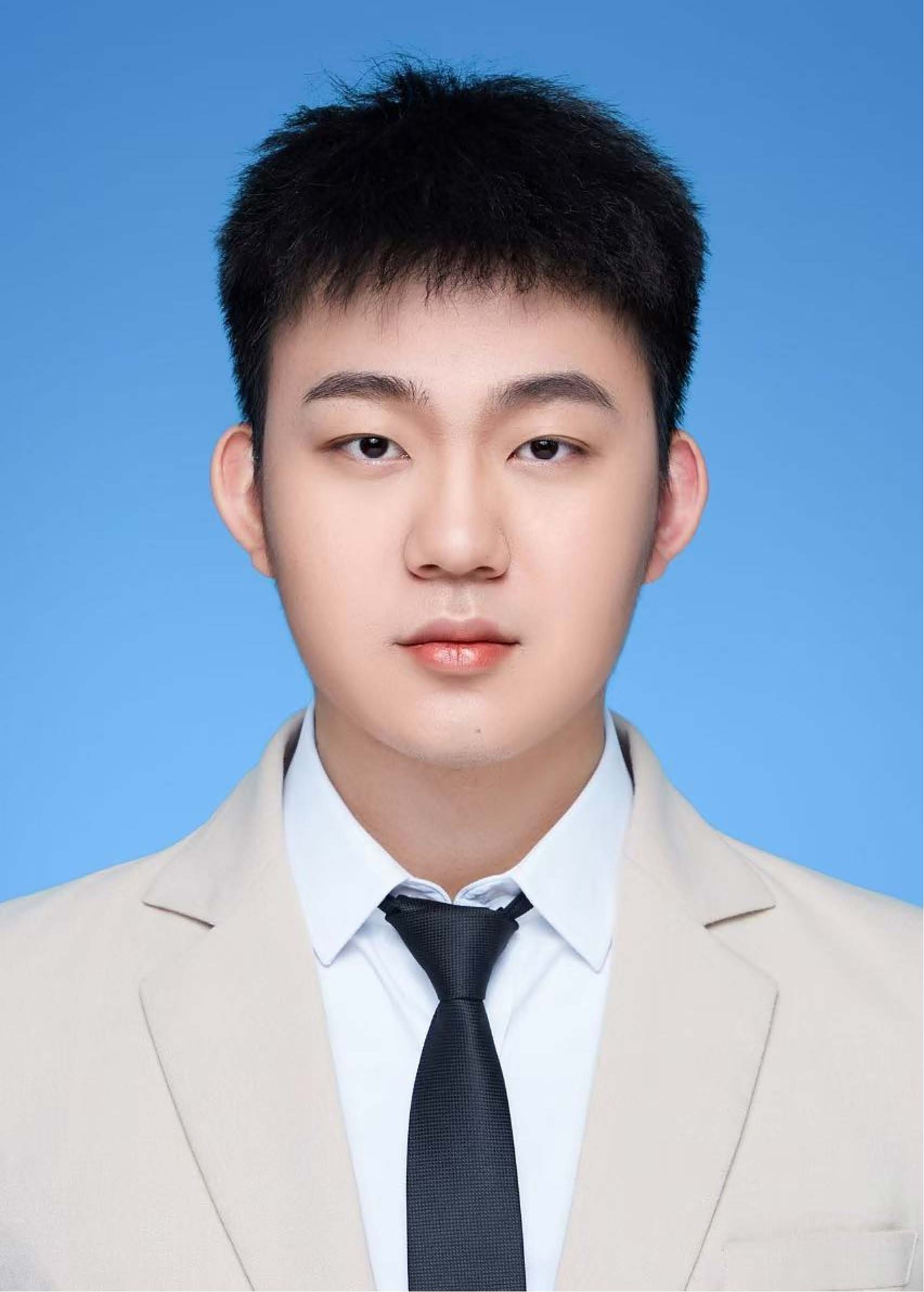}}]
	{Xingrun Yan} received the B.E. degree from Beijing Institute of Technology, China, in 2022. He is currently pursuing the master's degree with the School of Cyberspace Science and Technology, Beijing Institute of Technology, China. His research interests include federated learning and semantic communications.
\end{IEEEbiography}

\begin{IEEEbiography}[{\includegraphics[width=1.1in,height=1.375in,clip,keepaspectratio]{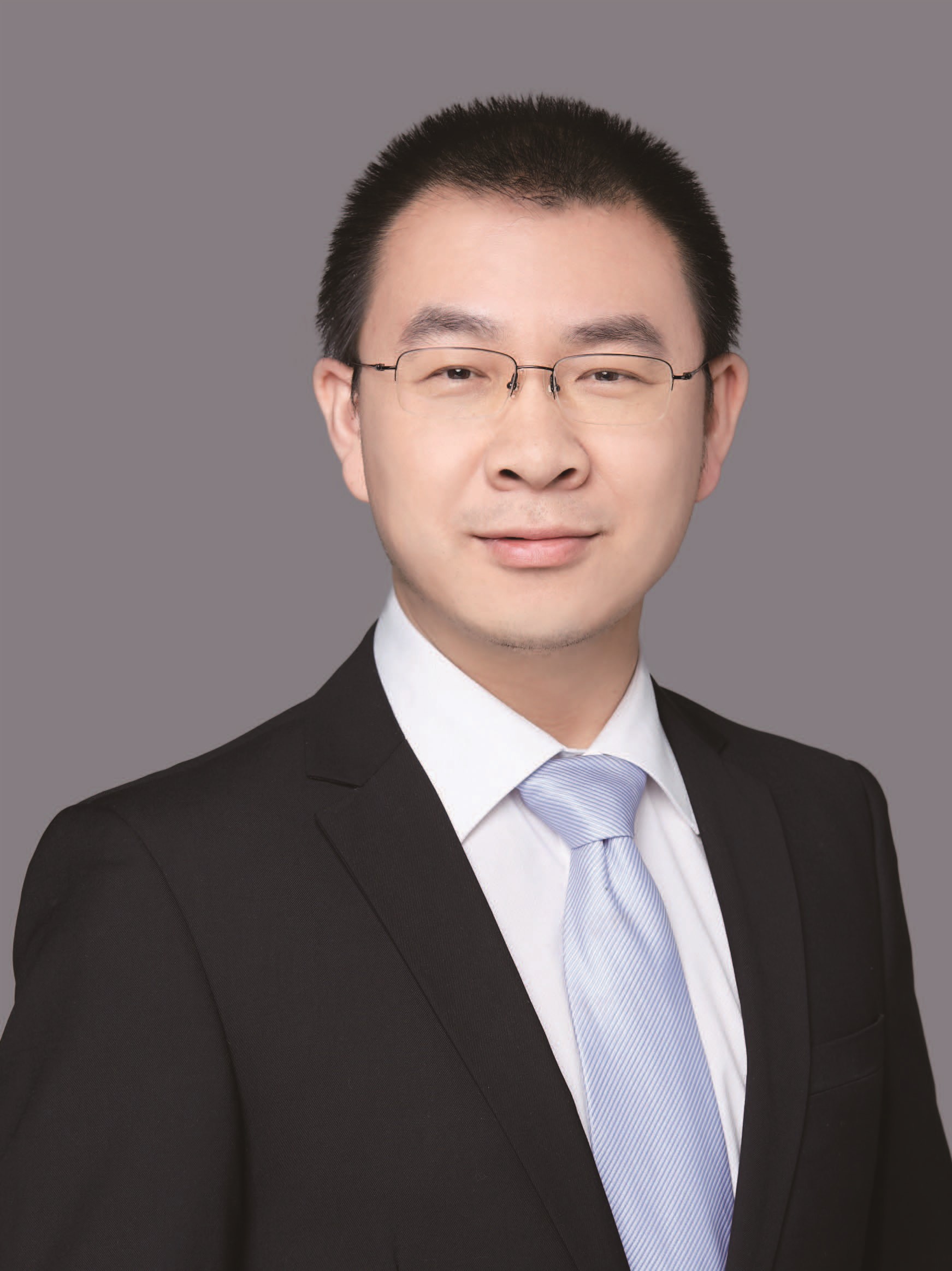}}]
	{Rongfei Fan} (Member, IEEE) received the B.E. degree in communication engineering from Harbin Institute of Technology, Harbin, China, in 2007, and the Ph. D degree in electrical engineering from the University of Alberta, Edmonton, Alberta, Canada, in 2012. Since 2013, he has been a faculty member at the Beijing Institute of Technology, Beijing, China, where he is currently an Associate Professor in the School of Cyberspace Science and Technology. His research interests include federated learning, semantic communications, edge computing, and resource allocation in wireless networks.
\end{IEEEbiography}

\begin{IEEEbiography} [{\includegraphics[width=1.1in,height=1.375in,clip,keepaspectratio]{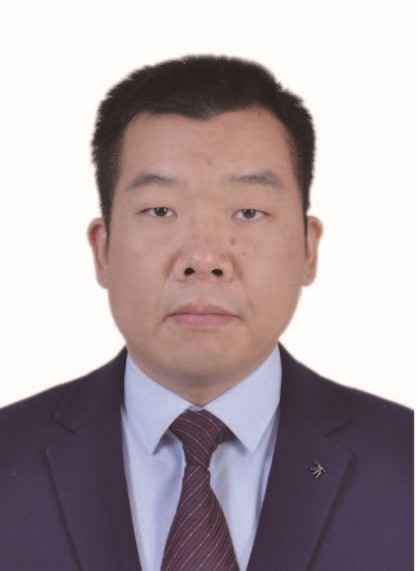}}]
	{Han Hu} (Member, IEEE) is currently a Professor with the School of Information and Electronics, Beijing Institute of Technology, China. He received the B.E. and Ph.D. degrees from the University of Science and Technology of China, China, in 2007 and 2012 respectively. His research interests include multimedia networking, edge intelligence and space-air-ground integrated network. 
    He received several academic awards, including Best Paper Award of IEEE TCSVT 2019, Best Paper Award of IEEE Multimedia Magazine 2015, Best Paper Award of IEEE Globecom 2013, etc. He served as an Associate Editor of IEEE TMM and Ad Hoc Networks, and a TPC Member of Infocom, ACM MM, AAAI, IJCAI, etc.
\end{IEEEbiography}

\begin{IEEEbiography} [{\includegraphics[width=1.1in,height=1.375in,clip,keepaspectratio]{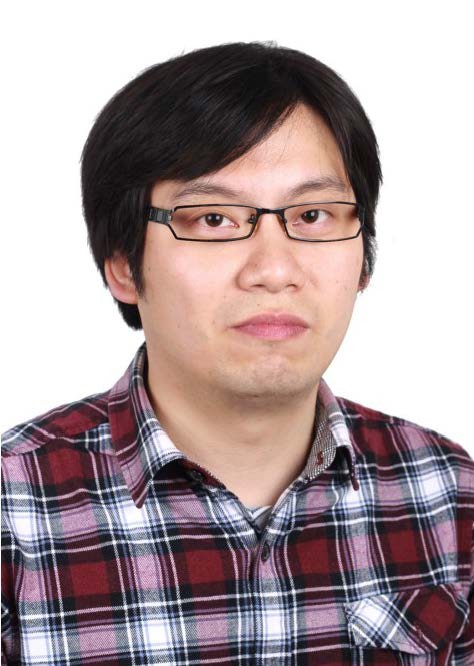}}]
	{Hangguan Shan} (Senior Member, IEEE) received the B.Sc. degree in electrical engineering from Zhejiang University, Hangzhou, China, in 2004, and the Ph.D. degree in electrical engineering from Fudan University, Shanghai, China, in 2009. From 2009 to 2010, he was a Postdoctoral Research Fellow with the University of Waterloo, Waterloo, ON, Canada. Since 2011, he has been with the College of Information Science and Electronic Engineering, Zhejiang University, where he is currently an Associate Professor. He is also with the Zhejiang Provincial Key Laboratory of Information Processing and Communication Networks, Zhejiang University. His current research interests include machine learningenabled resource allocation and quality-of-service provisioning in wireless networks. Dr. Shan has co-received the Best Industry Paper Award from the IEEE WCNC’11 and the Best Paper Award from the IEEE WCSP’23. He has served as a Technical Program Committee Member of various international conferences. He was an Editor of the IEEE TRANSACTIONS ON GREEN COMMUNICATIONS AND NETWORKING. 
\end{IEEEbiography}

\begin{IEEEbiography} [{\includegraphics[width=1.1in,height=1.375in,clip,keepaspectratio]{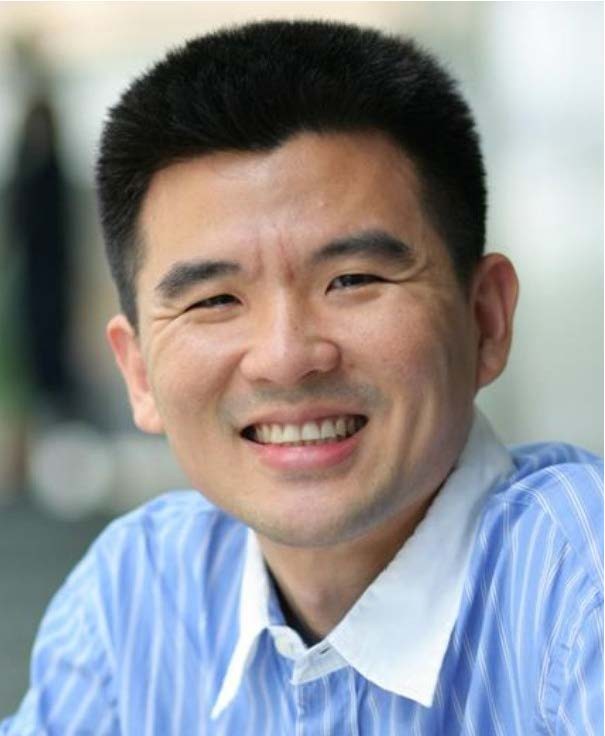}}]
	{Tony Q. S. Quek} (Fellow, IEEE) received the B.E. and M.E. degrees in electrical and electronics engineering from the Tokyo Institute of Technology in 1998 and 2000, respectively, and the Ph.D. degree in electrical engineering and computer science from the Massachusetts Institute of Technology in 2008. Currently, he is the Cheng Tsang Man Chair Professor with Singapore University of Technology and Design (SUTD) and ST Engineering Distinguished Professor. He also serves as the Director of the Future Communications R \& D Programme, the Head of ISTD Pillar, and the Deputy Director of the SUTDZJU IDEA. His current research topics include wireless communications and networking, network intelligence, non-terrestrial networks, open radio access network, and 6G.
    
    Dr. Quek has been actively involved in organizing and chairing sessions, and has served as a member of the Technical Program Committee as well as symposium chairs in a number of international conferences. He is currently serving as an Area Editor for the IEEE TRANSACTIONS ON WIRELESS COMMUNICATIONS.
    
    Dr. Quek was honored with the 2008 Philip Yeo Prize for Outstanding Achievement in Research, the 2012 IEEE William R. Bennett Prize, the 2015 SUTD Outstanding Education Awards – Excellence in Research, the 2016 IEEE Signal Processing Society Young Author Best Paper Award, the 2017 CTTC Early Achievement Award, the 2017 IEEE ComSoc AP Outstanding Paper Award, the 2020 IEEE Communications Society Young Author Best Paper Award, the 2020 IEEE Stephen O. Rice Prize, the 2020 Nokia Visiting Professor, and the 2022 IEEE Signal Processing Society Best Paper Award. He is the AI on RAN Working Group Chair in AI-RAN Alliance. He is a Fellow of IEEE, a Fellow of WWRF, and a Fellow of the Academy of Engineering Singapore.
\end{IEEEbiography}

\begin{IEEEbiography} [{\includegraphics[width=1.1in,height=1.375in,clip,keepaspectratio]{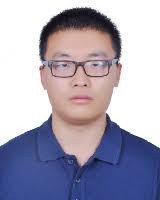}}]
	{Puning Zhao} received the B.S. degree from the University of Science and Technology of China, Hefei, China, in 2017, and the Ph.D. degree from the University of California at Davis, Davis, CA, USA, in 2021. He is currently an associate professor at School of Cyber Science and Technology, Shenzhen Campus of Sun Yat-sen University.  
\end{IEEEbiography}

%% file: appendix.tex
\begin{appendices}

  \section{Assumptions And Theoretical Results} 

  In this section, we provide assumptions and theoretical results of the RAGA under Byzantine attacks. Below, we first present the necessary assumptions.

  \subsection{Assumption}

    For all honest users $m \in \mathcal{M} \setminus \mathcal{B}$, we have the following assumptions. 
    
    \begin{assumption}[Lipschitz Continuity] \label{ass:lips}
      The loss function $f(\bm{w}, \bm{s})$ has $L$-Lipschitz continuity \cite{huang2023achieving, li2022analysis}, i.e., for $\forall \bm{w}_1, \bm{w}_2 \in \mathbb{R}^p$, there is
      \begin{align} \label{e:ass_lips_1}
        f(\bm{w}_1, \bm{s}) - f(\bm{w}_2, \bm{s}) 
        &\leqslant \left\langle \nabla f(\bm{w}_2, \bm{s}), \bm{w}_1 - \bm{w}_2 \right\rangle \nonumber \\ 
        &\qquad \qquad \quad + \frac{L}{2} \left\lVert \bm{w}_1 - \bm{w}_2 \right\rVert^2, 
      \end{align}
    which is equivalent to the following inequality, 
      \begin{align} \label{e:ass_lips_2}
        \left\lVert \nabla f(\bm{w}_1, \bm{s}) - \nabla f(\bm{w}_2, \bm{s}) \right\rVert \leqslant L \left\lVert \bm{w}_1 - \bm{w}_2 \right\rVert .
      \end{align}
    With (\ref{e:ass_lips_1}) or (\ref{e:ass_lips_2}), it can be derived that the global loss function $F(\bm{w})$ and local loss function $F_m(\bm{w})$ both have $L$-Lipschitz continuity. 
    \end{assumption}
    
    \begin{assumption}[{Strongly-Convex}] \label{ass:convex}
      The loss function $f(\bm{w}, \bm{s})$ is $\mu$ strongly-convex \cite{huang2023achieving}, i.e., for $\forall \bm{w}_1, \bm{w}_2 \in \mathbb{R}^p$, there is
      \begin{equation} \label{e:ass_convex}
        \left\lVert \nabla f(\bm{w}_1, \bm{s}) - \nabla f(\bm{w}_2, \bm{s}) \right\rVert \geqslant \mu \left\lVert \bm{w}_1 - \bm{w}_2 \right\rVert.
      \end{equation}
    Similarly, with (\ref{e:ass_convex}), $F(\bm{w})$ and $F_m(\bm{w})$ can be derived to be both $\mu$ strongly-convex.
    \end{assumption}
    
    \begin{assumption}[Local Unbiased Gradient] \label{ass:unbias}
      Suppose a subset of dataset $\mathcal{S}_m$ is selected randomly, which is denote as $\xi_m$, define
      \begin{equation}
        F_m(\bm{w};\xi_m)  \triangleq \frac{1}{\left\lvert \xi_m \right\rvert} \sum_{\bm{s} \in \xi_m} f(\bm{w}, \bm{s}), 
      \end{equation}
      the unbiased gradient \cite{huang2023achieving, li2022analysis} implies that
      \begin{equation}
        \mathbb{E}\left\{ \nabla F_m(\bm{w};\xi_m) \right\} = \nabla F_m(\bm{w}).
      \end{equation}
    \end{assumption}
    \begin{assumption}[Bounded Inner Variance] \label{ass:var}
      For each honest user $m$ and $\bm{w} \in \mathbb{R}^p$, the variance of stochastic local gradient $\nabla F_m(\bm{w};\xi_m)$ is upper-bounded \cite{huang2023achieving, li2022analysis} by
      \begin{align}
        \text{Var} (\nabla F_m(\bm{w};\xi_m)) 
        \leqslant \sigma^2.
      \end{align}
    \end{assumption}
    
    \begin{assumption}[Data Heterogeneity] \label{ass:hete}
        For each honest user $m$ and $\bm{w} \in \mathbb{R}^p$, the data heterogeneity can be defined as
        \begin{align}
            \theta_m = \left\lVert \nabla F_m(\bm{w}) - \nabla F(\bm{w}) \right\rVert. 
        \end{align}
        We also assume the data heterogeneity is bounded \cite{huang2023achieving, li2022analysis}, which implies 
        \begin{align}
            \theta_m \leqslant \theta. 
        \end{align}
    \end{assumption}
    
    \begin{assumption}[Bounded Gradient] \label{ass:boundedg}
      For each honest user $m$ and $\bm{w} \in \mathbb{R}^p$, the ideal local gradient $\nabla F_m(\bm{w})$ is upper-bounded \cite{huang2023achieving, li2022analysis} by 
      \begin{align}
        \left\lVert \nabla F_m(\bm{w}) \right\rVert \leqslant G.
      \end{align}
    \end{assumption}
    
    \subsection{Theoretical Results}
    
    In this subsection, we present the analytical results of convergence on our proposed RAGA. All proofs are deferred to Appendices \ref{app:smooth} and \ref{app:conv}. 
    
    \subsubsection{Case \Rmnum{1}: With Lipschitz Continuity Only for Loss Function}
    
    We first inspect the case with  Assumptions \ref{ass:lips}, \ref{ass:unbias}, \ref{ass:var}, \ref{ass:hete} and \ref{ass:boundedg} imposed, which allows the loss function to be non-convex. 
    By defining  
     \begin{align} \label{def:rato}
        \alpha_m = \frac{S_m}{\sum_{j=1}^{M} S_j}, C_{\alpha} = \sum_{m \in \mathcal{M} \setminus \mathcal{B}} \alpha_m, 
      \end{align}
      and 
      \begin{equation} \label{e:p_t_def}
        p^t = \left\{
        \begin{aligned}
          &{0, }  &       &{\eta^t \leqslant \frac{1}{L}}\\
          &{1, }  &       &{\eta^t > \frac{1}{L}}
        \end{aligned} 
        \right. , 
      \end{equation}
    the following results can be expected.
    \begin{theorem} \label{theo:smooth}
      With {$C_{\alpha} > 0.5$}, there is 
      \begin{align} \label{equa:smooth}
        &\quad \sum_{t=1}^{T} \frac{\eta^t}{\sum_{{t'}=1}^{T} \eta^{t'}} \mathbb{E}\left\{ \left\lVert \nabla F(\bm{w}^t) \right\rVert ^2 \right\} \nonumber \\ 
        &\leqslant \frac{2\mathbb{E} \left\{ F(\bm{w}^{1}) - F(\bm{w}^{T}) \right\}}{\sum_{{t'}=1}^{T} \eta^{t'}} + \frac{1}{\sum_{{t'}=1}^{T} \eta^{t'}} \sum_{t=1}^{T} \eta^t \Delta^t \nonumber \\
        &\quad + \sum_{t=1}^{T} \frac{2(\eta^t + p^t(\eta^t)^2 L - p^t\eta^t)\epsilon^2}{(2C_{\alpha}-1)^2\sum_{{t'}=1}^{T} \eta^{t'} } \nonumber \\
        &\quad + \sum_{t=1}^{T} \frac{8(C_{\alpha})^2(\eta^t\sigma^2+(p^t(\eta^t)^2 L - p^t\eta^t)(G^2+\sigma^2))}{(2C_{\alpha}-1)^2\sum_{{t'}=1}^{T} \eta^{t'}}, 
      \end{align}
      where
      \begin{align} \label{equ:delta}
        \Delta^t = 
        &\frac{8C_{\alpha}}{(2C_{\alpha}-1)^2} \sum_{m \in \mathcal{M} \setminus \mathcal{B}} \Bigg( \frac{2\alpha_mL^2(G^2 + \sigma^2)}{K^t} \sum_{k=2}^{K^t} (k - 1) \nonumber \\ 
        &\quad \cdot \sum_{i=1}^{k-1} (\eta_m^{t, i})^2 + 2 \alpha_m \theta^2 \Bigg). 
      \end{align}  
    \end{theorem}
    \begin{IEEEproof}
      Please refer to Appendix \ref{app:smooth}. 
    \end{IEEEproof}
    
    \subsubsection{Case \Rmnum{2}: With Lipschitz Continuity and Strong Convexity for Loss Function}
    
    For the case with {Assumptions \ref{ass:lips}, \ref{ass:convex}, \ref{ass:unbias}, \ref{ass:var}, \ref{ass:hete} and \ref{ass:boundedg}} imposed, which requires the loss function to be not only Lipschitz but also strongly-convex, the following results can be anticipated.
    \begin{theorem} \label{theo:conv}
      With {$0 < \lambda^t <1$} and $C_{\alpha} > 0.5$, the optimality gap $\mathbb{E} \left\{ F(\bm{w}^{T}) - F(\bm{w}^*) \right\}$ is upper bounded in (\ref{equa:gene}) as follows, 
      \begin{align} \label{equa:gene}
        &\quad \mathbb{E} \left\{ F(\bm{w}^{T}) - F(\bm{w}^*) \right\} \nonumber \\
        &\leqslant \frac{L}{2} \mathbb{E} \left\{ \left\lVert \bm{w}^1  - \bm{w}^{*} \right\rVert^2 \right\} \prod_{t = 1}^{T-1} \gamma^t + \frac{L}{2} \sum_{t=1}^{T-1} \frac{(\eta^t)^2}{\lambda^t} \Bigg( \Delta^t +  \nonumber \\ 
        &\quad \frac{8\sigma^2(C_{\alpha})^2}{(2C_{\alpha}-1)^2} + \frac{2\epsilon^2}{\left( 2C_{\alpha}-1 \right)^2} \Bigg) \cdot \prod_{i=t}^{T-1} (\gamma^{i+1})^{q^{i+1}}, 
      \end{align} 
      where 
      \begin{equation}
        q^t = \left\{
        \begin{aligned}
          &{0, }  &       &{t = T}\\
          &{1, }  &       &{t < T}
        \end{aligned}  \nonumber
        \right. , 
      \end{equation}
      and
      \begin{align}
        \gamma^t = \frac{1 -2\eta^t\mu+L^2(\eta^t)^2}{1-\lambda^t}. \nonumber
      \end{align}
    \end{theorem}
    \begin{IEEEproof}
      Please refer to Appendix \ref{app:conv}. 
    \end{IEEEproof}

  \section{Proof of Theorem \ref{theo:smooth}}\label{app:smooth}
  The proof of Theorem \ref{theo:smooth} relies on the holding of Lemma \ref{lem:unbiaszt} and Lemma \ref{lem:zt}, which is given as follows:
  \begin{lemma} \label{lem:unbiaszt}
    With Assumptions \ref{ass:lips}, \ref{ass:unbias}, \ref{ass:var}, \ref{ass:hete}, \ref{ass:boundedg}, and $C_{\alpha} > 0.5$, the term $\mathbb{E}\left\{ \left\lVert \bm{z}^t - \nabla F(\bm{w}^t) \right\rVert ^2 \right\}$ can be upper bounded as
    \begin{align}
      &\quad \mathbb{E}\left\{ \left\lVert \bm{z}^t - \nabla F(\bm{w}^t) \right\rVert ^2 \right\} \leqslant \Delta^t + \frac{8\sigma^2(C_{\alpha})^2}{(2C_{\alpha}-1)^2} + \frac{2\epsilon^2}{\left( 2C_{\alpha}-1 \right)^2}.
    \end{align}
  \end{lemma}
  \begin{IEEEproof}
    Please refer to Appendix \ref{app:unbiaszt}.
  \end{IEEEproof}

  \begin{lemma} \label{lem:zt}
    With Assumptions \ref{ass:unbias}, \ref{ass:var}, \ref{ass:boundedg}, and $C_{\alpha} > 0.5$, the term $\mathbb{E}\left\{ \left\lVert \bm{z}^t \right\rVert ^2 \right\}$ can be bounded as
    \begin{align}
      0 \leqslant \mathbb{E}\left\{ \left\lVert \bm{z}^t  \right\rVert ^2 \right\} \leqslant \frac{8(C_{\alpha})^2(G^2+\sigma^2)}{(2C_{\alpha}-1)^2} + \frac{2\epsilon^2}{\left( 2C_{\alpha}-1 \right)^2}.
    \end{align}
  \end{lemma}
  \begin{IEEEproof}
    Please refer to Appendix \ref{app:zt}.
  \end{IEEEproof}

Under Assumption \ref{ass:lips}, Lemmas \ref{lem:unbiaszt} and \ref{lem:zt}, and recalling the definition of $p^t$ as given in (\ref{e:p_t_def}), we have
  \begin{subequations} \label{e:app_A_F_t}
    \begin{align}
      &\quad \mathbb{E}\left\{ F(\bm{w}^{t+1}) - F(\bm{w}^{t}) \right\} \nonumber \\
      &\leqslant \mathbb{E}\left\{ \left\langle \nabla F(\bm{w}^{t}), \bm{w}^{t+1} - \bm{w}^{t} \right\rangle \right\} + \frac{L}{2} \mathbb{E}\left\{ \left\lVert \bm{w}^{t+1} - \bm{w}^{t} \right\rVert ^2 \right\} \\
      &= -\eta^t \mathbb{E}\left\{ \left\langle \nabla F(\bm{w}^{t}), \bm{z}^t \right\rangle \right\} + \frac{(\eta^t)^2L}{2} \mathbb{E}\left\{ \left\lVert \bm{z}^t \right\rVert ^2 \right\} \\
      &= \frac{\eta^t}{2} \mathbb{E}\left\{ \left\lVert \bm{z}^t - \nabla F(\bm{w}^t) \right\rVert ^2 \right\} - \frac{\eta^t}{2} \mathbb{E}\left\{ \left\lVert \nabla F(\bm{w}^t) \right\rVert ^2 \right\} \nonumber \\
      &\quad + \frac{(\eta^t)^2 L - \eta^t}{2} \mathbb{E}\left\{ \left\lVert \bm{z}^t \right\rVert ^2 \right\} \\
      &= \frac{\eta^t}{2} \mathbb{E}\left\{ \left\lVert \bm{z}^t - \nabla F(\bm{w}^t) \right\rVert ^2 \right\} - \frac{\eta^t}{2} \mathbb{E}\left\{ \left\lVert \nabla F(\bm{w}^t) \right\rVert ^2 \right\} \nonumber \\
      &\quad + \frac{(\eta^t)^2 L - \eta^t}{2} p^t \mathbb{E}\left\{ \left\lVert \bm{z}^t \right\rVert ^2 \right\} \\
      &\leqslant - \frac{\eta^t}{2} \mathbb{E}\left\{ \left\lVert \nabla F(\bm{w}^t) \right\rVert ^2 \right\} + \frac{\eta^t}{2} \Delta^t \nonumber \\
      & \quad + \frac{(\eta^t + p^t(\eta^t)^2 L - p^t\eta^t)\epsilon^2}{(2C_{\alpha}-1)^2} \nonumber \\
      &\quad + \frac{4(C_{\alpha})^2(\eta^t\sigma^2+(p^t(\eta^t)^2 L - p^t\eta^t)(G^2+\sigma^2))}{(2C_{\alpha}-1)^2}. 
    \end{align}
  \end{subequations}

Summarizing the inequality in (\ref{e:app_A_F_t}) for $t =1, 2, ..., T$ and dividing the summarized inequality with $\sum_{t=1}^{T} {\eta^t}/{2}$ for both sides, we obtain
  \begin{align}
    &\quad \sum_{t=1}^{T} \frac{\eta^t}{\sum_{t'=1}^{T} \eta^t} \mathbb{E}\left\{ \left\lVert \nabla F(\bm{w}^t) \right\rVert ^2 \right\} \nonumber \\ 
    &\leqslant \frac{2\mathbb{E} \left\{ F(\bm{w}^{1}) - F(\bm{w}^{T}) \right\}}{\sum_{t'=1}^{T} \eta^t} + \frac{1}{\sum_{t=1}^{T} \eta^t} \sum_{t=1}^{T} \eta^t \Delta^t \nonumber \\
    &\quad + \sum_{t=1}^{T} \frac{2(\eta^t + p^t(\eta^t)^2 L - p^t\eta^t)\epsilon^2}{(2C_{\alpha}-1)^2\sum_{t'=1}^{T} \eta^t } \nonumber \\
    &\quad + \sum_{t=1}^{T} \frac{8(C_{\alpha})^2(\eta^t\sigma^2+(p^t(\eta^t)^2 L - p^t\eta^t)(G^2+\sigma^2))}{(2C_{\alpha}-1)^2\sum_{t'=1}^{T} \eta^t}. 
  \end{align} 
 
This completes the  proof of Theorem \ref{theo:smooth}. 

  \section{Proof of Theorem \ref{theo:conv}} \label{app:conv}
With Assumptions \ref{ass:lips} and the fact that $\nabla F(\bm{w}^*) = \bm{0}$, there is
  \begin{subequations} \label{e:app_convex_F_w}
    \begin{align}
      &\quad \mathbb{E} \left\{ F(\bm{w}^{t+1}) - F(\bm{w}^*) \right\} \nonumber \\
      &\leqslant \mathbb{E}\left\{ \left\langle \nabla F(\bm{w}^{*}), \bm{w}^{t+1} - \bm{w}^{*} \right\rangle \right\} + \frac{L}{2} \mathbb{E}\left\{ \left\lVert \bm{w}^{t+1} - \bm{w}^{*} \right\rVert ^2 \right\} \\
      &= \frac{L}{2} \mathbb{E}\left\{ \left\lVert \bm{w}^{t+1} - \bm{w}^{*} \right\rVert ^2 \right\}.  
    \end{align}
  \end{subequations}

For the term $\mathbb{E}\left\{ \left\lVert \bm{w}^{t+1} - \bm{w}^{*} \right\rVert ^2 \right\}$, with $0 < \lambda^t < 1$, it leads to 
  \begin{subequations} \label{e:convex_theorem_app}
    \begin{align}
      &\quad \mathbb{E}\left\{ \left\lVert \bm{w}^{t+1} - \bm{w}^{*} \right\rVert ^2 \right\} \nonumber \\ 
      &= \mathbb{E} \Bigl\{ \big\lVert \bm{w}^{t+1} - \bm{w}^t + \eta^t \nabla F(\bm{w}^t) \nonumber \\
      &\qquad \qquad \qquad \qquad + \bm{w}^t - \eta^t \nabla F(\bm{w}^t) - \bm{w}^{*} \big\rVert ^2 \Bigr\}  \\ 
      &\leqslant \frac{1}{\lambda^t} \mathbb{E} \left\{ \left\lVert \bm{w}^{t+1} - \bm{w}^t + \eta^t \nabla F(\bm{w}^t) \right\rVert^2 \right\} \nonumber \\ 
      &\quad + \frac{1}{1 - \lambda^t} \mathbb{E} \left\{ \left\lVert \bm{w}^t - \eta^t \nabla F(\bm{w}^t) - \bm{w}^{*} \right\rVert^2 \right\} \label{equ:wcon1} \\ 
      &= \frac{(\eta^t)^2}{\lambda^t} \mathbb{E} \left\{ \left\lVert \bm{z}^t - \nabla F(\bm{w}^t) \right\rVert^2 \right\} + \frac{1}{1-\lambda^t} \mathbb{E} \left\{ \left\lVert \bm{w}^t - \bm{w}^{*} \right\rVert^2 \right\} \nonumber \\
      &\quad - \frac{2\eta^t}{1-\lambda^t} \left\langle \bm{w}^t - \bm
      {w}^* , \nabla F(\bm{w}^t) - \nabla F(\bm{w}^*) \right\rangle \nonumber \\
      &\quad + \frac{(\eta^t)^2}{1-\lambda^t} \mathbb{E} \left\{ \left\lVert \nabla F(\bm{w}^t) - \nabla F(\bm{w}^*) \right\rVert^2 \right\} \label{equ:wcon2} \\ 
      &\leqslant \frac{(\eta^t)^2}{\lambda^t} \mathbb{E} \left\{ \left\lVert \bm{z}^t - \nabla F(\bm{w}^t) \right\rVert^2 \right\} \nonumber \\
      &\quad + \frac{1 -2\mu\eta^t+L^2(\eta^t)^2}{1-\lambda^t} \mathbb{E} \left\{ \left\lVert \bm{w}^t - \bm{w}^{*} \right\rVert^2 \right\}, \label{equ:wcon3}
    \end{align}
  \end{subequations}
  where 
  \begin{itemize}
    \item the inequality (\ref{equ:wcon1}) comes from Cauchy-Schwarz inequality $\left( \frac{a^2}{c}  + \frac{b^2}{1-c}  \right)\left(c + 1-c\right) \geqslant \left(a+b\right)^2, 0< c < 1$;
    \item the equality (\ref{equ:wcon2}) is established because of $\nabla F(\bm{w}^*) = \bm{0}$;
    \item the inequality (\ref{equ:wcon3}) is bounded by Assumptions \ref{ass:lips} and \ref{ass:convex}.
  \end{itemize}

Applying the inequality in (\ref{e:convex_theorem_app}) for $t=1, 2, ..., T$, and recalling that $\gamma^t = \frac{1 -2\mu\eta^t+L^2(\eta^t)^2}{1 - \lambda^t}$ and the definition of $q^t$, we then have
  \begin{align} \label{e:app_convex_w_z}
    &\mathbb{E}\left\{ \left\lVert \bm{w}^{T} - \bm{w}^{*} \right\rVert ^2 \right\} 
    \leqslant \prod_{t = 1}^{T-1} \gamma^t \mathbb{E} \left\{ \left\lVert \bm{w}^1 - \bm{w}^{*} \right\rVert^2 \right\} \nonumber \\
    &\quad + \sum_{t=1}^{T-1} \frac{(\eta^t)^2}{\lambda^t} \prod_{i=t}^{T-1} (\gamma^{i+1})^{q^{i+1}} \mathbb{E} \left\{ \left\lVert \bm{z}^t - \nabla F(\bm{w}^t) \right\rVert^2 \right\}. 
  \end{align}

With the help of Lemma \ref{lem:unbiaszt}, and combining (\ref{e:app_convex_F_w}) and (\ref{e:app_convex_w_z}), there is
  \begin{align} 
    &\quad \mathbb{E} \left\{ F(\bm{w}^{T}) - F(\bm{w}^*) \right\} \nonumber \\
    &\leqslant \frac{L}{2} \mathbb{E} \left\{ \left\lVert \bm{w}^1  - \bm{w}^{*} \right\rVert^2 \right\} \prod_{t = 1}^{T-1} \gamma^t + \frac{L}{2} \sum_{t=1}^{T-1} \frac{(\eta^t)^2}{\lambda^t} \Bigg[ \Delta^t +  \nonumber \\ 
    &\quad \frac{8\sigma^2(C_{\alpha})^2}{(2C_{\alpha}-1)^2} + \frac{2\epsilon^2}{\left( 2C_{\alpha}-1 \right)^2} \Bigg] \cdot \prod_{i=t}^{T-1} (\gamma^{i+1})^{q^{i+1}}.
  \end{align}
  
This completes the proof of Theorem \ref{theo:conv}. 

  \section{Proof of Lemma \ref{lem:unbiaszt}} \label{app:unbiaszt}
  The proof of Lemma \ref{lem:unbiaszt} relies on Lemma \ref{lem:geom}, i.e., 
  \begin{lemma} \label{lem:geom}
    Let $\mathcal{Z} = \left\{\bm{z}_1, \bm{z}_2, ..., \bm{z}_M \right\}$ be a set of uploaded vectors by $M$ users. The $\mathcal{Z}$ contains $B$ Byzantine attack vectors, which compose of a subset $\mathcal{Z}^{'}$. When $C_{\alpha} > 0.5$, there is
    \begin{align}\label{unequ:gemo}
      &\quad \mathbb{E} \left \{ \left\lVert {\rm geomed} \left(\{\bm{z}_m| m\in \mathcal{M}\}, \epsilon \right) \right \lVert^2 \right \} \nonumber \\
      &\leqslant \frac{8C_{\alpha}}{(2C_{\alpha}-1)^2} \sum_{\bm{z}_i \in \mathcal{Z} \setminus \mathcal{Z}^{'}} \alpha_i \mathbb{E} \left\lVert \bm{z}_{i} \right\rVert^2 + \frac{2\epsilon^2}{\left( 2C_{\alpha}-1 \right)^2}.
    \end{align}
  \end{lemma}
  \begin{IEEEproof}
    Please refer to Appendix \ref{app:geom}. 
  \end{IEEEproof}

With Lemma \ref{lem:geom}, there is
  \begin{align} \label{equ:cengeom}
    &\mathbb{E}\left\{ \left\lVert \bm{z}^t - \nabla F(\bm{w}^t) \right\rVert ^2 \right\} \leqslant \frac{2\epsilon^2}{\left( 2C_{\alpha}-1 \right)^2} \nonumber \\
    &\quad + \frac{8C_{\alpha}}{(2C_{\alpha}-1)^2} \sum_{m \in \mathcal{M} \setminus \mathcal{B}} \alpha_m \mathbb{E} \left\lVert \bm{z}_m^t - \nabla F(\bm{w}^t) \right\rVert^2,
  \end{align}
since \\
$\bm{z}^t - \nabla F(\bm{w}^t) = {\rm geomed} \left(\{\bm{z}_m^t - \nabla F(\bm{w}^t) | m\in \mathcal{M}\}, \epsilon \right)$.

For the term $\mathbb{E}\left\{ \left\lVert \bm{z}_m^t - \nabla F(\bm{w}^t) \right\rVert^2 \right\}$, $m \in \mathcal{M} \setminus \mathcal{B}$, 
  we have
  \begin{subequations} \label{e:app_C_1}
    \begin{align}
      &\quad \mathbb{E}\left\{ \left\lVert \bm{z}_m^t - \nabla F(\bm{w}^t) \right\rVert^2 \right\} \nonumber \\
      &= \mathbb{E}\left\{ \left\lVert \frac{1}{K^t} \sum_{k=1}^{K^t} \nabla F_m(\bm{w}_m^{t, k-1};\xi_m^{t, k}) - \nabla F(\bm{w}^t) \right\rVert^2 \right\} \\
      &= \frac{1}{(K^t)^2} \mathbb{E}\left\{ \left\lVert \sum_{k=1}^{K^t} \left( \nabla F_m(\bm{w}_m^{t, k-1};\xi_m^{t, k}) - \nabla F(\bm{w}^t) \right) \right\rVert^2 \right\} \\
      &\leqslant \frac{1}{K^t} \sum_{k=1}^{K^t} \mathbb{E}\left\{ \left\lVert \nabla F_m(\bm{w}_m^{t, k-1};\xi_m^{t, k}) - \nabla F(\bm{w}^t) \right\rVert^2 \right\} \label{equ:zsmo1} \\
      &= \frac{1}{K^t} \sum_{k=1}^{K^t} \mathbb{E}\Big\{ \Big\lVert \nabla F_m(\bm{w}_m^{t, k-1};\xi_m^{t, k}) - \nabla F_m(\bm{w}_m^{t, k-1}) \nonumber \\
      &\qquad \qquad \qquad \quad + \nabla F_m(\bm{w}_m^{t, k-1}) - \nabla F(\bm{w}^t) \big\rVert^2 \Big\} \\ 
      &= \frac{1}{K^t} \sum_{k=1}^{K^t} \mathbb{E}\Big\{ \left\lVert \nabla F_m(\bm{w}_m^{t, k-1};\xi_m^{t, k}) - \nabla F_m(\bm{w}_m^{t, k-1}) \right\rVert^2 \nonumber \\
      &\quad + \left\lVert \nabla F_m(\bm{w}_m^{t, k-1}) - \nabla F(\bm{w}^t) \right\rVert^2 \nonumber \\
      &\quad + 2 \big\langle \nabla F_m(\bm{w}_m^{t, k-1};\xi_m^{t, k}) - \nabla F_m(\bm{w}_m^{t, k-1}), \nonumber \\ 
      &\qquad \qquad \qquad \qquad \quad \nabla F_m(\bm{w}_m^{t, k-1}) - \nabla F(\bm{w}^t) \big\rangle \Big\}, 
    \end{align}
  \end{subequations}
  where the inequality (\ref{equ:zsmo1}) holds because of the Cauchy-Schwarz inequality $(a_1+a_2+a_3+\dots+a_n)^2 \leqslant n(a_1^2+a_2^2+a_3^2+\dots+a_n^2)$. With Assumptions \ref{ass:unbias} and \ref{ass:var}, there are $\mathbb{E} \left\{\nabla F_m(\bm{w}_m^{t, k-1};\xi_m^{t, k}) - \nabla F_m(\bm{w}_m^{t, k-1})\right\} = 0$ and $\mathbb{E} \left\{\left\lVert \nabla F_m(\bm{w}_m^{t, k-1};\xi_m^{t, k}) - \nabla F_m(\bm{w}_m^{t, k-1}) \right\rVert^2\right\} \leqslant \sigma^2$, then (\ref{e:app_C_1}) leads to
  \begin{subequations} \label{equ:mcenbound}
    \begin{align}
      &\quad \mathbb{E}\left\{ \left\lVert \bm{z}_m^t - \nabla F(\bm{w}^t) \right\rVert^2 \right\} \nonumber \\
      &\leqslant \frac{1}{K^t} \sum_{k=1}^{K^t} \mathbb{E}\left\{ \left\lVert \nabla F_m(\bm{w}_m^{t, k-1}) - \nabla F(\bm{w}^t) \right\rVert^2 \right\} + \sigma^2 \\
      &= \frac{1}{K^t} \sum_{k=1}^{K^t} \mathbb{E}\Big\{ \big\lVert \nabla F_m(\bm{w}_m^{t, k-1}) - \nabla F(\bm{w}_m^{t, k-1}) \nonumber \\  
      &\qquad \qquad \qquad + \nabla F(\bm{w}_m^{t, k-1}) - \nabla F(\bm{w}^t) \big\rVert^2 \Big\} + \sigma^2 \\ 
      &\leqslant \frac{2}{K^t} \sum_{k=1}^{K^t} \mathbb{E} \left\{ \left\lVert \nabla F(\bm{w}_m^{t, k-1}) - \nabla F(\bm{w}^t) \right\rVert^2 \right\} \nonumber \\
      &\quad + \frac{2}{K^t} \sum_{k=1}^{K^t} \mathbb{E} \left\{ \left\lVert \nabla F_m(\bm{w}_m^{t, k}) - \nabla F(\bm{w}_m^{t, k}) \right\rVert^2 \right\} + \sigma^2, \label{equ:zsmo4}
    \end{align}
  \end{subequations}
 where the inequality (\ref{equ:zsmo4}) can be derived from $(a+b)^2 \leqslant 2a^2+2b^2$.

To further upper bound the right-hand side of (\ref{equ:mcenbound}), since $(\theta_m^{t, k})^2 = \left\lVert \nabla F_m(\bm{w}_m^{t, k}) - \nabla F(\bm{w}_m^{t, k}) \right\rVert^2 \leqslant \theta ^2$, we only need to inspect the bound of $\sum_{k=1}^{K^t} \mathbb{E} \left\{ \left\lVert \nabla F(\bm{w}_m^{t, k-1}) - \nabla F(\bm{w}^t) \right\rVert^2 \right\}$, which is given as follows 
  \begin{subequations} \label{equ:glomuti}
    \begin{align}
      &\quad \sum_{k=1}^{K^t} \mathbb{E} \left\{ \left\lVert \nabla F(\bm{w}_m^{t, k-1}) - \nabla F(\bm{w}^t) \right\rVert^2 \right\} \nonumber \\
      &\leqslant L^2 \sum_{k=1}^{K^t} \mathbb{E} \left\{ \left\lVert \bm{w}_m^{t, k-1} - \bm{w}^t \right\rVert^2 \right\} \label{equ:zsmo5} \\
      &= L^2 \sum_{k=2}^{K^t} \mathbb{E} \left\{ \left\lVert \sum_{i=1}^{k-1} \eta_m^{t, i} \cdot \nabla F_m(\bm{w}_m^{t, i-1};\xi_m^{t, i}) \right\rVert^2 \right\} \\
      &\leqslant L^2 \sum_{k=2}^{K^t}\left( \sum_{i=1}^{k-1} (\eta_m^{t, i})^2 \cdot \sum_{i=1}^{k-1} \mathbb{E} \left\{ \left\lVert \nabla F_m(\bm{w}_m^{t, i-1};\xi_m^{t, i}) \right\rVert^2 \right\} \right) \label{equ:zsmo6} \\
      &\leqslant L^2 \sum_{k=2}^{K^t}\left( \sum_{i=1}^{k-1} (\eta_m^{t, i})^2 \cdot (k - 1)(G^2 + \sigma^2) \right)  \label{equ:zsmo7} \\
      &= L^2(G^2 + \sigma^2) \sum_{k=2}^{K^t} (k - 1) \sum_{i=1}^{k-1} (\eta_m^{t, i})^2, 
    \end{align}
  \end{subequations}
  where 
  \begin{itemize}
    \item the inequality (\ref{equ:zsmo5}) holds according to Assumption \ref{ass:lips}; 
    \item the inequality (\ref{equ:zsmo6}) comes from Cauchy-Schwarz inequality $(a_1b_1+a_2b_2+a_3b_3+\dots+a_nb_n)^2 \leqslant (a_1^2+a_2^2+a_3^2+\dots+a_n^2)(b_1^2+b_2^2+b_3^2+\dots+b_n^2)$; 
    \item the inequality (\ref{equ:zsmo7}) can be derived from Assumptions \ref{ass:unbias}, \ref{ass:var} and \ref{ass:boundedg}. 
  \end{itemize}

Combining {(\ref{equ:delta})}, (\ref{equ:cengeom}), (\ref{equ:mcenbound}) and (\ref{equ:glomuti}), we obtain the following inequality
  \begin{align}
    \mathbb{E}\left\{ \left\lVert \bm{z}^t - \nabla F(\bm{w}^t) \right\rVert ^2 \right\} 
    \leqslant \Delta^t + \frac{8\sigma^2(C_{\alpha})^2}{(2C_{\alpha}-1)^2} + \frac{2\epsilon^2}{\left( 2C_{\alpha}-1 \right)^2}.
  \end{align}

This completes the proof of Lemma \ref{lem:unbiaszt}. 

  \section{Proof of Lemma \ref{lem:zt}} \label{app:zt}
With Lemma \ref{lem:geom}, there is 
  \begin{align}
    \mathbb{E}\left\{ \left\lVert \bm{z}^t  \right\rVert ^2 \right\} 
    \leqslant \frac{8C_{\alpha}}{(2C_{\alpha}-1)^2} \sum_{m \in \mathcal{M} \setminus \mathcal{B}} \alpha_m \mathbb{E} \left\lVert \bm{z}_m^t \right\rVert^2 + \frac{2\epsilon^2}{\left( 2C_{\alpha}-1 \right)^2}.
  \end{align}
  
For the term $\mathbb{E}\left\{ \left\lVert \bm{z}_m^t \right\rVert^2 \right\}$, $m \in \mathcal{M} \setminus \mathcal{B}$, according to Assumptions \ref{ass:unbias}, \ref{ass:var} and \ref{ass:boundedg}, we have 
  \begin{subequations}
    \begin{align}
      \mathbb{E}\left\{ \left\lVert \bm{z}_m^t \right\rVert^2 \right\} 
      &= \mathbb{E}\left\{ \left\lVert \frac{1}{K^t} \sum_{k=1}^{K^t} \nabla F_m(\bm{w}_m^{t, k-1};\xi_m^{t, k}) \right\rVert^2 \right\} \\ 
      &\leqslant \frac{1}{K^t} \sum_{k=1}^{K^t} \mathbb{E}\left\{ \left\lVert \nabla F_m(\bm{w}_m^{t, k-1};\xi_m^{t, k}) \right\rVert^2 \right\} \label{e:app_D_ineq1} \\
      &\leqslant G^2 + \sigma^2,
    \end{align}
  \end{subequations}
where the inequality (\ref{e:app_D_ineq1}) comes from Cauchy-Schwarz inequality $(a_1+a_2+a_3+\dots+a_n)^2 \leqslant n(a_1^2+a_2^2+a_3^2+\dots+a_n^2)$.
Hence we can get the following inequality
  \begin{align}
    \mathbb{E}\left\{ \left\lVert \bm{z}^t  \right\rVert ^2 \right\} \leqslant \frac{8(C_{\alpha})^2(G^2+\sigma^2)}{(2C_{\alpha}-1)^2} + \frac{2\epsilon^2}{\left( 2C_{\alpha}-1 \right)^2}.
  \end{align}

Since $\mathbb{E}\left\{ \left\lVert \bm{z}^t  \right\rVert ^2 \right\} \geqslant 0$, then there is 
  \begin{align}
    0 \leqslant \mathbb{E}\left\{ \left\lVert \bm{z}^t  \right\rVert ^2 \right\} \leqslant \frac{8(C_{\alpha})^2(G^2+\sigma^2)}{(2C_{\alpha}-1)^2} + \frac{2\epsilon^2}{\left( 2C_{\alpha}-1 \right)^2}.
  \end{align}

This completes the proof of Lemma \ref{lem:zt}. 

\section{Proof of Lemma \ref{lem:geom}} \label{app:geom}

For the ease of presentation, we denote \\
$\bm{z}^* = {\rm geomed} \left(\{\bm{z}_m| m\in \mathcal{M}\} \right)$ and \\
$\bm{z}_{\epsilon }^* ={\rm geomed} \left(\{\bm{z}_m| m\in \mathcal{M}\}, \epsilon\right)$, then there is
  \begin{align}
    \sum_{i=1}^M \frac{S_i}{\sum_{j=1}^{M} S_j} \left\lVert \bm{z}_{\epsilon}^* - \bm{z}_i \right\rVert \leqslant \sum_{i=1}^M \frac{S_i}{\sum_{j=1}^{M} S_j} \left\lVert \bm{z}^* - \bm{z}_i \right\rVert + \epsilon,
  \end{align}
which implies 
\begin{align} \label{e:app_e_geo_ep_geo}
    \sum_{i=1}^M \alpha_i \left\lVert \bm{z}_{\epsilon}^* - \bm{z}_i \right\rVert \leqslant \sum_{i=1}^M \alpha_i \left\lVert \bm{z}^* - \bm{z}_i \right\rVert + \epsilon,
  \end{align}
by recalling the definition of $\alpha_i$ as given in (\ref{def:rato}).

By the definition of geometric median, there is
  \begin{align}
    \sum_{i=1}^M \alpha_i \left\lVert \bm{z}^* - \bm{z}_i \right\rVert = \inf_{\bm{y}} \sum_{i=1}^M \alpha_i \left\lVert \bm{y} - \bm{z}_i \right\rVert 
    \leqslant \sum_{i=1}^M \alpha_i \left\lVert \bm{z}_i \right\rVert. \label{equ:appb1}
  \end{align}

  According to the triangle inequality, we have
  \begin{align}
    \sum_{\bm{z}_i \in \mathcal{Z}^{'}} \alpha_i \left\lVert \bm{z}_{\epsilon}^* - \bm{z}_i \right\rVert \geqslant \sum_{\bm{z}_i \in \mathcal{Z}^{'}} \alpha_i \left( \left\lVert \bm{z}_i \right\rVert - \left\lVert \bm{z}_{\epsilon}^* \right\rVert \right), \label{equ:tri1} \\ 
    \sum_{\bm{z}_i \in \mathcal{Z} \setminus \mathcal{Z}^{'}} \alpha_i \left\lVert \bm{z}_{\epsilon}^* - \bm{z}_i \right\rVert \geqslant \sum_{\bm{z}_i \in \mathcal{Z} \setminus \mathcal{Z}^{'}} \alpha_i \left( \left\lVert \bm{z}_{\epsilon}^* \right\rVert - \left\lVert \bm{z}_i \right\rVert \right). \label{equ:tri2}
  \end{align}
  
By adding (\ref{equ:tri1}) and (\ref{equ:tri2}) together, we obtain
  \begin{align}
    \sum_{\bm{z}_i \in \mathcal{Z}} \alpha_i \left\lVert \bm{z}_{\epsilon}^* - \bm{z}_i \right\rVert 
    &\geqslant \sum_{\bm{z}_i \in \mathcal{Z}} \alpha_i \left\lVert \bm{z}_i \right\rVert + (1 - \sum_{\bm{z}_i \in \mathcal{Z}^{'}} 2\alpha_i) \left\lVert \bm{z}_{\epsilon}^* \right\rVert \nonumber \\
    &\quad - 2\sum_{\bm{z}_i \in \mathcal{Z} \setminus \mathcal{Z}^{'}} \alpha_i \left\lVert \bm{z}_i \right\rVert. \label{equ:appbtri}
  \end{align}

Combining (\ref{e:app_e_geo_ep_geo}), (\ref{equ:appb1}), and (\ref{equ:appbtri}), and with the fact that $\sum_{\bm{z}_i \in \mathcal{Z}^{'}} \alpha_i = 1- C_{\alpha} <1/2$, we have the following inequality
  \begin{align}
    \left\lVert \bm{z}_{\epsilon}^* \right\rVert
    \leqslant \frac{2\sum_{\bm{z}_i \in \mathcal{Z} \setminus \mathcal{Z}^{'}} \alpha_i \left\lVert \bm{z}_{i} \right\rVert + \epsilon}{1 - 2\sum_{\bm{z}_i \in \mathcal{Z}^{'}} \alpha_i}. \label{equ:geomeps}
  \end{align}
  
Squaring both sides of (\ref{equ:geomeps}) and based on the Cauchy-Schwarz inequality, we further have
  \begin{subequations}
    \begin{align}
      &\quad \left\lVert \bm{z}_{\epsilon}^* \right\rVert ^2 \nonumber \\
      &\leqslant \left( \frac{2\sum_{\bm{z}_i \in \mathcal{Z} \setminus \mathcal{Z}^{'}} \alpha_i \left\lVert \bm{z}_{i} \right\rVert + \epsilon}{1 - 2\sum_{\bm{z}_i \in \mathcal{Z}^{'}} \alpha_i} \right)^2 \\ 
      &\leqslant \frac{8\left( \sum_{\bm{z}_i \in \mathcal{Z} \setminus \mathcal{Z}^{'}} \alpha_i \left\lVert \bm{z}_{i} \right\rVert \right)^2 + 2\epsilon^2}{\left( 1 - 2\sum_{\bm{z}_i \in \mathcal{Z}^{'}} \alpha_i \right)^2} \label{equ:z1} \\
      &= \frac{8\left( \sum_{\bm{z}_i \in \mathcal{Z} \setminus \mathcal{Z}^{'}} \alpha_i \right)^2}{\left( 1 - 2\sum_{\bm{z}_i \in \mathcal{Z}^{'}} \alpha_i \right)^2} \left( \sum_{\bm{z}_i \in \mathcal{Z} \setminus \mathcal{Z}^{'}} \frac{\alpha_i}{\sum_{\bm{z}_j \in \mathcal{Z} \setminus \mathcal{Z}^{'}} \alpha_j} \left\lVert \bm{z}_{i} \right\rVert \right)^2 \nonumber \\
      &\quad + \frac{2\epsilon^2}{\left( 1 - 2\sum_{\bm{z}_i \in \mathcal{Z}^{'}} \alpha_i \right)^2} \\
      &\leqslant \frac{8\left( \sum_{\bm{z}_i \in \mathcal{Z} \setminus \mathcal{Z}^{'}} \alpha_i \right) \sum_{\bm{z}_i \in \mathcal{Z} \setminus \mathcal{Z}^{'}} \alpha_i \left\lVert \bm{z}_{i} \right\rVert^2 + 2\epsilon^2}{\left( 1 - 2\sum_{\bm{z}_i \in \mathcal{Z}^{'}} \alpha_i \right)^2}, \label{equ:z2}
    \end{align}
  \end{subequations}
  where
  \begin{itemize}
    \item the inequality (\ref{equ:z1}) comes from $(a+b)^2 \leqslant 2a^2+2b^2$;
    \item the inequality (\ref{equ:z2}) holds because of Jensen's inequality $f\left(\sum_{i=1}^{n} \nu_i x_i \right) \leqslant \sum_{i=1}^{n} \nu_i f(x_i)$ with $\sum_{i=1}^{n} \nu_i =1$, $\nu_i\geq 0$ for $i=1, 2, ..., n$, and $f(x) = x^2$ being a convex function. 
  \end{itemize}

Recalling of the definition of $C_{\alpha}$ in (\ref{def:rato}), we then have
  \begin{align} \label{e:app_e_eq_end}
    \left\lVert \bm{z}_{\epsilon}^* \right\rVert ^2 \leqslant \frac{8C_{\alpha}}{\left(2C_{\alpha}-1\right)^2} \sum_{\bm{z}_i \in \mathcal{Z} \setminus \mathcal{Z}^{'}} \alpha_i \left\lVert \bm{z}_{i} \right\rVert^2 + \frac{2\epsilon^2}{\left( 2C_{\alpha}-1 \right)^2}.
  \end{align} 

Taking the expectation of the both sides of (\ref{e:app_e_eq_end}) can lead to the statement of Lemma \ref{lem:geom}. 

\end{appendices}